\documentclass[10pt]{article}

\usepackage{footmisc}
\usepackage{amsfonts}
\usepackage{amsmath, amssymb}
\usepackage{siunitx}
\usepackage{./style/ws-rotating}     
\usepackage{epstopdf}
\usepackage{./style/graphicx}
\usepackage{./style/bxascmac}

\usepackage{./style/slashbox}
\usepackage{./style/fancybox}
\usepackage{./style/framed}
\usepackage{./style/extarrows}
\usepackage{./style/pifont}

\usepackage{amsmath,amsthm,amssymb}
\usepackage{makeidx,epsfig,lscape}

 \usepackage[colorlinks = false,
            linkcolor = red,
            urlcolor  = magenta,
            citecolor = green]{hyperref}

\DeclareMathAlphabet{\mathpzc}{OT1}{pzc}{m}{it}

\def\bt{\begin{tabular}}
\def\et{\end{tabular}}
\def\and{\mbox{ and }}

\def\1{{\bf 1}}

\begin{document}

$\mbox{ }$

 \vskip 12mm

{ 
{\noindent{\Large\bf
  Some Interesting Features of Memristor CNN}}
\\[6mm]
{\centering
{\large Makoto Itoh\footnote{After retirement from Fukuoka Institute of Technology, he has continued to study the nonlinear dynamics on memristors. }}\\
{\it 1-19-20-203, Arae, Jonan-ku, \\ 
Fukuoka, 814-0101 JAPAN\\
Email: {itoh-makoto@jcom.home.ne.jp}}\\[1mm]
}

 { 
 {\noindent{\\[3mm]\textup{ 
In this paper, we introduce some interesting features of a memristor CNN (Cellular Neural Network). 
We first show that there is the similarity between the dynamics of memristors and neurons.  
That is, some kind of flux-controlled memristors can \emph{not} respond to the sinusoidal voltage source quickly, 
namely, they can \emph{not} switch ``on'' rapidly.   
Furthermore, these memristors have \emph{refractory period} after switch ``on'', which means that it can \emph{not} respond to further sinusoidal inputs until the flux is decreased. 
We next show that the memristor-coupled two-cell CNN can exhibit chaotic behavior.  
In this system, the memristors switch ``off'' and ``on'' at \emph{irregular intervals}, 
and the two cells are connected when either or both of the memristors switches ``on''. 
We then propose the modified CNN model, which can hold a binary output image, even if all cells are disconnected and no signal is supplied to the cell after a certain point of time.  
However, the modified CNN requires power to maintain the output image, that is, it is \emph{volatile}.    
We next propose a new memristor CNN model.  
It can also hold a binary output state (image), even if all cells are disconnected, and no signal is supplied to the cell, by memristor's switching behavior.
Furthermore, even if we turn off the power of the system during the computation, it can resume from the previous \emph{average} output state, since the memristor CNN has functions of both \emph{short-term} (volatile) memory and \emph{long-term} (non-volatile) memory.  
The above suspend and resume feature are useful when we want to save the current state, and continue work later from the previous state. 
Finally, we show that the memristor CNN can exhibit interesting two-dimensional waves, if an inductor is connected to each memristor CNN cell.  
 }}}
\\[4mm]

{\noindent{Keywords: \
 memristor CNN; non-volatile; chaotic behavior; suspend and resume feature; long-term memory; short-term memory; two-dimensional wave; switch; neuron; synapse; excitatory; inhibitory; refractory period.
}

%
%
\section{Introduction}
\label{sec: introduction}
%
%
In this paper, we introduce some interesting features of a memristor CNN (Cellular Neural Network).\footnote{The terminology CNN was originally used for Cellular Neural Network \cite{{Chua1998},{Roska}}.  
Recently, CNN is also used for  Convolutional Neural Network.  In this paper, CNN stands for Cellular Neural Network.}    
We first study the switching behavior of flux-controlled memristors. 
The flux-controlled memristor switches ``off'' and ``on'', depending on the value of the flux.   
Some kind of flux-controlled memristors cannot respond to the sinusoidal voltage source quickly. 
That is, it cannot switch ``on'' rapidly.   
Furthermore, these memristors have \emph{refractory period} after switch ``on'', that is, it can not respond to further sinusoidal inputs until the flux is decreased. 
We also show that the memristor-coupled two-cell can exhibit chaotic behavior.  
In this system, the memristors switch ``off'' and ``on'' at irregular intervals, 
and the two cells are connected when either or both of the memristors switches ``on''. 

We next propose the modified CNN and the memristor CNN. 
The modified CNN can hold a binary output image even if all cells are disconnected and no signal is supplied to the cell after a certain point of time. 
The modified CNN requires power to maintain the output image, that is, it is \emph{volatile}.  
We can realize the above switching behavior by using flux-controlled memristors, 
since the memristor can switch ``off'' and ``on'', depending on the value of the flux.
The memristor CNN can also hold a binary output image, 
even if all cells are disconnected, and even if no signal is supplied to the cell, by memristor's switching behavior. 
However, it also requires power to maintain the output image, since the nonlinear element (nonlinear resistor) of the CNN cell are \emph{volatile}. 
 
It is well-known that the neurons can not respond to inputs quickly and they cannot generate outputs rapidly,   
since charging or discharging the membrane potential energy can take time.  
Furthermore, after firing, the neurons have \emph{refractory period}.  
We show that the image processing (visual computing) of the memoristor CNN can exhibit the similar behavior, 
if the memductance of the flux-controlled memristor has twin-peaks.   

The suspend and resume feature are useful when we want to save the current state, and continue work later from the same state. 
We show that the memristor CNN has this kind of feature.
That is, even if we turn off the power of the memristor CNN during the computation, it can recover the \emph{average} output state \emph{later}, by using the \emph{non-volatile} memristors.     
Furthermore, it can resume from the previous \emph{average} output.    

In our brain's system, a \emph{long-term memory} is a storage system for storing and retrieving information.  
A \emph{short-term memory} is the \emph{short-time} storage system that keeps something in mind before transferring it to a long-term memory.
We also show that the memristor CNN has functions of the short-term and long-term memories.  

Finally, we show that the memristor CNN can exhibit interesting two-dimensional waves, if an inductor is connected to each memristor CNN cell.  
In this case, the dynamics of the CNN cell is given by the $2$nd-order differential equation.  

%
%
%
\section{Basic Notations and Definitions}
\label{sec: basic}
%
%
%
In this section, we introduce some basic notations and definitions which will be used later. 
%
%
%
\subsection{Cellular Neural Networks}
\label{sec: CNN}
%
%
%
Cellular Neural Network (CNN) \cite{{Chua1998},{Roska},{Itoh2003},{Itoh2009}} is a dynamic nonlinear system defined by coupling only identical simple dynamical systems, called cells, located within a prescribed sphere of influence, such as nearest neighbors.  
The dynamics of a standard cellular neural network with a neighborhood of radius $r$ are governed by a system of $n=MN$ differential equations  
\begin{center}
\begin{minipage}[h]{8.5cm}
\begin{itembox}[l]{Dynamics of the CNN}
\begin{equation}
 \begin{array}{c}
  \displaystyle \frac{dx_{ij}}{dt} = - \gamma x_{ij}+ \sum_{k,l \in N_{ij}}
    ( a_{k l} \ y_{kl}+b_{k l} \ u_{kl}) + z_{ij}, \vspace{2mm} \\
    (i, \, j) \in \{ 1, \cdots , M \} \times \{ 1, \cdots , N \}, 
 \end{array}
\label{eqn: cnn1} 
\end{equation}
\end{itembox} 
\end{minipage}
\end{center}
where $x_{ij}$, $y_{kl}$, $u_{kl}$, $z_{ij}$ are called state, output, input, and threshold of cell $C_{ij}$, respectively.   
$N_{ij}$ denotes the $r$-neighborhood of cell $C_{ij}$, and $a_{kl}, \, b_{kl}$, and $z_{ij}$ denote the feedback, control, and threshold template parameters, respectively.   
The output $y_{ij}$ and the state $x_{ij}$ of each cell are usually related via the piecewise-linear saturation function
\begin{equation}
  y_{ij} = f(x_{ij}) \stackrel{\triangle}{=}
  \frac{1}{2} \bigl ( \, |x_{ij}+1|-|x_{ij}-1| \, \bigr ).  
\end{equation}
The matrices $A= [ a_{kl} ]$ and $B= [ b_{kl} ]$ are referred to as the feedback template $A$ and the feedforward (input) template $B$, respectively.  
If we restrict the neighborhood radius of every cell to $1$, then the cell $C_{ij}$ is coupled only to its eight neighbor cells $C_{kl}$, where 
\begin{equation}
(k, l)=(i+1, j+1), (i+1, j), (i+1, j-1), (i, j+1), (i, j-1), (i-1, j+1), (i-1, j), (i-1, j-1).  
\end{equation}
Assume that $z_{ij}$ is the same for the whole network, that is, $z_{ij}=z$, and set $\gamma = 1$ for the sake of simplicity.
Then, the template $\{ A, B, z \}$ is fully specified by $19$ parameters, which are the elements of two $3 \times 3$ matrices $A$ and $B$, namely 
\begin{equation}
 A =
 \begin{array}{|c|c|c|}
  \hline
   a_{-1, -1} & a_{-1, 0} & a_{-1,1} \\
  \hline
   a_{0, -1}  & a_{0, 0}  & a_{0,1}  \\  
  \hline 
   a_{1, -1}  & a_{1, 0}  & a_{1,1}  \\ 
  \hline
  \end{array} \ , \ \ \ 
 B = 
   \begin{array}{|c|c|c|}
  \hline
   b_{-1, -1} & b_{-1, 0} & b_{-1,1} \\
  \hline
   b_{0, -1} & b_{0, 0} & b_{0,1}    \\  
  \hline 
   b_{1, -1} & b_{1, 0} & b_{1,1}    \\ 
  \hline
  \end{array} \ , 
\label{eqn: AB}
\end{equation}
and a real number $z$.  
The output and input for the cell cell $C_{ij}$ are specified by 
\begin{equation}
\begin{array}{c}
 Y =
 \begin{array}{|c|c|c|}
  \hline
   y_{i-1, \, j-1}  & y_{i-1, \, j}  & y_{i-1, \, j+1}  \\
  \hline
   y_{i, \, j-1}    & y_{i, \, j}    & y_{i, \, j+1}    \\  
  \hline 
   y_{i+1, \, j-1}  & y_{i+1, \, j}  & y_{i+1, \, j+1}  \\ 
  \hline
  \end{array} \ , \vspace{3mm} \\ 
 U = 
   \begin{array}{|c|c|c|}
  \hline
   u_{i-1, \, j-1} & u_{i-1, \, j}  & u_{i-1, \, j}   \\
  \hline
   u_{i, \, j-1}   & u_{i, \, j}    & u_{i, \, j+1}   \\  
  \hline 
   u_{i+1, \, j-1}  & u_{i+1, \, j} & u_{i+1, \, j+1} \\ 
  \hline
  \end{array} \ . 
\end{array}
\label{eqn: YU}
\end{equation}
Here, we assumed that the feedback parameters $a_{kl}$ and the control template parameters $b_{kl}$ do not vary with space, that is, they can be defined as \cite{{Chua1998},{Roska}}
\begin{equation}
  a_{k l} \stackrel{\triangle}{=} a_{k-i, \, l-j}, \ \  b_{k l} \stackrel{\triangle}{=} b_{k-i, \, l-j}.   
\end{equation}
For example, 
\begin{equation} 
  \begin{array}{lll}
    \text{if~~} k=i, \, l=j,  &\text{~~then~~} &  \left. 
                                                    \begin{array}{c}
                                                      a_{k l} = a_{k-i, \, l-j} = a_{0,  0} \\ 
                                                      b_{k l} = b_{k-i, \, l-j} = b_{0, \, 0}  
                                                    \end{array} 
                                                  \right \} \vspace{2mm}\\
    \text{if~~} k=i, \, l=j+1, &\text{~~then~~} & \left. 
                                                      \begin{array}{c}
                                                        a_{k l} = a_{k-i, \, l-j} = a_{0, \,1} \\
                                                        b_{k l} = b_{k-i, \, l-j} = b_{0, \,1}  
                                                      \end{array} 
                                                    \right \}
  \end{array}
\end{equation}
Thus, we obtained the four elements:  
\begin{equation}
  a_{0, 0}, \ a_{0, 1}, \ b_{0, 0}, \ b_{0, 1},  
\end{equation}
in Eq. (\ref{eqn: AB}).  
Similarly, we can obtain all other elements of templates $A$ and $B$.  

The CNN template $\{ A, B, z \}$ is usually designed such that the qualitative behavior is not affected by the small perturbation of the template. 
If the CNN template satisfies this property, then the output image remains unchanged, even if we change the template parameters slightly. It also remains unchanged in the presence of sufficiently small noise.    

Many applications of the CNN templates are to convert gray-scale images to binary images.    
In this paper, the luminance value of the pixel would be coded as black $\to +1$, white $\to -1$, gray $\to (-1, \ 1)$.  
Furthermore, in order to simulate Eq. (\ref{eqn: cnn1}) in a computer, the initial condition $x_{ij}(0)$ and the boundary condition must be specified.  
For example, the fixed boundary condition is given by   
\begin{equation}
  v_{k^{*}l^{*}}  =  v_{0},  \  u_{k^{*}l^{*}}  = u_{0},
\end{equation}
where $k^{*}l^{*}$ denotes boundary cells, and $v_{0}$ and $u_{0}$ are constants.  

Finally, let us consider an isolated CNN cell, which does not have the inputs, the outputs from other cells, and the threshold, for later use.     
Its dynamics is given by the first-order differential equation: 
\begin{equation} 
  \begin{array}{lll}
   \displaystyle \frac{dx_{ij}}{dt} 
   &=& - x_{ij} + a_{0, 0} \, y_{ij} = - x_{ij} + a_{0, 0} \, f(x_{ij}) \vspace{2mm}\\
   &=& \displaystyle - x_{ij}+ \frac{a_{0,  0}}{2} \bigl ( \, |x_{ij}+1|-|x_{ij}-1| \, \bigr ),   
  \end{array}
\label{eqn: simplest-cell}
\end{equation}
where $a_{0, 0}$ is the feedback parameter.   
The output $y_{ij}$ and the state $x_{ij}$ of the isolated cell  $C_{ij}$ is related by
\begin{equation}
  y_{ij} = f(x_{ij}) \stackrel{\triangle}{=} \frac{1}{2} \bigl ( \, |x_{ij}+1|-|x_{ij}-1| \, \bigr ).  
\end{equation}
Assume $a_{0, 0} > 1$.  
Then Eq. (\ref{eqn: simplest-cell}) has three equilibrium points, one is unstable and others are stable \cite{Chua1998}.    
We study its detailed behavior in Sec. \ref{sec: isolated}. 

%
%
%
\subsection{Memristors}
\label{sec: memristors}
%
%
%
%
The \emph{memristor} shown in Figures \ref{fig:mem1}-\ref{fig:mem3} is a passive 2-terminal electronic device described by a nonlinear relation
\begin{equation}
  \varphi = g(q), \ \ \text{or} \ \ q = h( \varphi ),
\label{eqn: mem-q-v}
\end{equation}
between the charge $q$ and the flux $\varphi$ \cite{{Chua1971}, {Chua-Kang}}. 
Its terminal voltage $v$ and the terminal current $i$ is given by 

\begin{center}
\begin{minipage}[c]{8.0cm}
\begin{itembox}[l]{$v-i$ characteristic of memristors}
\begin{equation}
  v = M(q)i  , \ \ \text{or} \ \ i = W( \varphi )v,
\label{eqn: mem-i-v}
\end{equation}
where 
\begin{equation}
 v = \frac{d\varphi}{dt}, \ \ \text{and} \ \  i = \frac{dq}{dt}.
\end{equation}
\end{itembox}
\end{minipage}
\end{center}
The two nonlinear functions $M(q)$ and $W( \varphi )$, called the \emph{memristance} and \emph{memductance}, respectively, are defined by
\begin{equation}
  M(q) \stackrel{\triangle}{=} \frac{dg(q)}{dq},
\end{equation}
and
\begin{equation}
  W(\varphi) \stackrel{\triangle}{=} \frac{dh(\varphi)}{d\varphi},
\end{equation}
representing the \emph{slope} of a scalar function $\varphi = \varphi (q)$ and $q = q( \varphi )$, respectively, called the \emph{memristor constitutive relation}.  

A memristor characterized by a differentiable $q-\varphi$ (resp., $\varphi-q$) characteristic curve is \emph{passive} if, and only if, its small-signal memristance $M(q)$ (resp., small-signal memductance $W(\varphi)$) is nonnegative (see \cite{Chua1971}).  
Since the instantaneous power dissipated by the above memristor is given by 
\begin{equation}
  p(t)= M \bigl( q(t) \bigl ) \ i(t)^{2} \ge 0,
\end{equation}
or
\begin{equation}
  p(t)= W \bigl ( \varphi(t) \bigl ) \ v(t)^{2} \ge 0,
\end{equation}
the energy flow into the memristor from time $t_{0}$ to $t$ satisfies 
\begin{equation}
 \int_{t_{0}}^{t} p( \tau ) d \tau \ge 0,
\end{equation}
for all $t \ge t_{0}$.  

In this paper, we assume the followings unless otherwise noted: 
%
%
\begin{center}
\begin{minipage}{12cm}
\begin{shadebox}
\begin{enumerate}
\item The memristor is \emph{ideal}, that is, it is endowed with a \emph{nonvolatility property}.  
\item The \emph{passive} memristor will not lose its flux or charge via the parasitic elements 
when we switch off the power.
\end{enumerate}
\end{shadebox}
\end{minipage}
\end{center}
%
%

\begin{figure}[hpbt]
 \begin{center} 
  \psfig{file=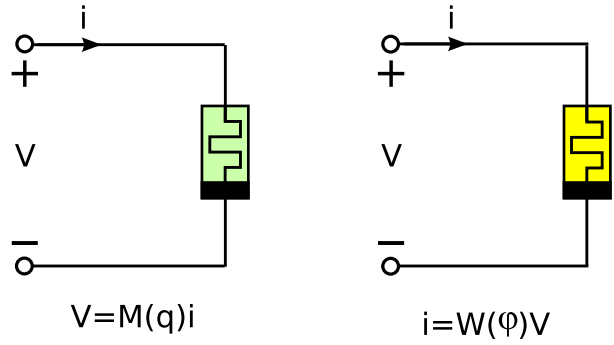,width=8cm}
  \caption{$v-i$ characteristic of memristors: charge-controlled memristor with the terminal voltage $v = M(q)\,i$ ~(left).  
   flux-controlled memristor with the terminal current $i = W( \varphi )\,v$ ~(right).}
 \label{fig:mem1}
 \end{center}
\end{figure}
%
%

\begin{figure}[hpbt]
 \begin{center} 
  \psfig{file=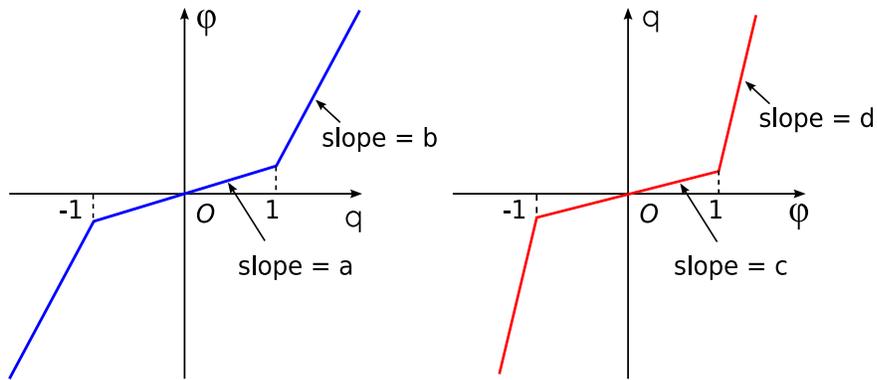,width=11.5cm}
  \caption{The constitutive relation of a monotone-increasing piecewise-linear memristor:  
   charge-controlled memristor described by a nonlinear relation 
   $\varphi = g (q) \stackrel{\triangle}{=} b q + 0.5(a - b)( |q + 1|  - |q - 1|)$ (left).  
   flux-controlled memristor described by a nonlinear relation 
   $q = h (\varphi) \stackrel{\triangle}{=} c \varphi + 0.5(c - d)( |\varphi+ 1|  - |\varphi- 1|)$ (right).}
  \label{fig:mem2} 
 \end{center}
\end{figure}
%
%

\begin{figure}[h]
 \begin{center} 
  \psfig{file=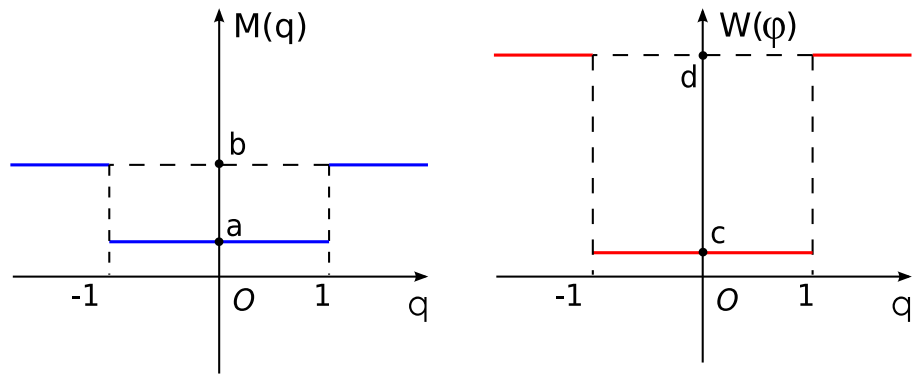,width=12cm}
  \caption{Memristance $M(q)$ of a charge-controlled memristor ~(left).  
   Memductance $W( \varphi )$ of a flux-controlled memristor ~(right).}
 \label{fig:mem3}
 \end{center}
\end{figure}
%
%

\begin{figure}[b]
 \begin{center} 
  \psfig{file=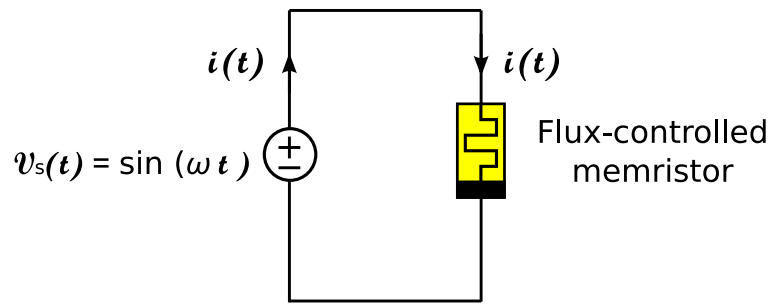,width=10.0cm}
  \caption{Two-element circuit which consists of a flux-controlled memristor and a periodic voltage source $v_{s}(t) = \sin (\omega t)$, where $\omega = 0.2$.}
 \label{fig:mem4}
 \end{center}
\end{figure}
\clearpage

%
%
\section{Neuron-like Response of Memristors} 
\label{sec: example1}
%
Consider the two-element circuit shown in Figure \ref{fig:mem4}.  
It consists of a flux-controlled memristor and a periodic voltage source $v_{s}(t) = \sin (\omega t)$, where $\omega = 0.2$.
The constitutive relation of the flux-controlled memristor is given by 
\begin{equation} 
  q = h (\varphi) = 0.5 \ ( \ |\varphi - a| - |\varphi - b| + b - a  \ ), 
\label{eqn: f-phi}
\end{equation} 
where $a=0.5$, $b=7$ (see Figure \ref{fig:mem5}(a)), and the flux $\varphi (t)$ is defined by 
\begin{equation} 
  \varphi (t) =  \int_{-\infty}^{t}v_{s}(\tau) d\tau.  
\label{eqn: phi}
\end{equation} 
The memductance $W( \varphi )$ is given by   
\begin{equation} 
 \begin{array}{ccc}
   W( \varphi )  &=& \displaystyle  \frac{dh(\varphi)}{d\varphi} 
                  = \mathfrak{s}[\varphi + 0.5] - \mathfrak{s}[\varphi - 7]  \vspace{2mm} \\
   
   &=&  \left \{ 
   \begin{array}{llcc}
     1 & \ for \ & \ & 0.5 < \varphi < 7,  \vspace{2mm} \\
     0 & \ for \ & \ & \varphi \le 0.5  \text{~and~} \varphi \ge 7, 
   \end{array}
    \right.  
 \end{array} 
\label{eqn: w-phi}
\end{equation} 
where $\mathfrak{s}  [\, z \,]$ denotes the \emph{unit step} function, equal to $0$ for $z < 0$ and 1 for $z \ge 0$.
Thus, the memristor switches ``off'' and ``on'', depending on the value of the flux $\varphi$ as shown in Figure \ref{fig:mem5}(b).
Its terminal voltage $v(t)$ and the terminal current $i(t)$ satisfy
\begin{equation}
  i(t) = W( \varphi (t) ) \, v_{s}(t),
\label{eqn: mem-i-v-2}
\end{equation}
where 
\begin{equation}
  \frac{d\varphi (t)}{dt} = v(t) = v_{s}(t) = \sin (\omega t).  
\label{eqn: dphi-dt}
\end{equation}
We show the pinched hysteresis loop of flux-controlled memristor in Figure \ref{fig:mem6}.  
We also show the waveforms of the terminal voltage $v(t)$, the terminal current $i(t)$, the flux $\varphi(t)$, 
and the memductance $W(\varphi (t))$  in Figure \ref{fig:mem7}.  
Observe that the current flows through the memristor if $0.5 < \varphi (t) < 7$ 
and no current flows through the memristor if $\varphi \le 0.5$ and $\varphi \ge 7$ as shown in Figure \ref{fig:mem7}(c), 
since the memristor switches ``off'' and ``on'', depending on the value of the flux $\varphi$ as shown in Figure \ref{fig:mem7}(c).
    
There is the similarity between memristors and neurons.  
That is, the neurons cannot respond to inputs quickly and they cannot generate outputs rapidly.  
Charging or discharging the membrane potential energy can take time.  
Furthermore, after firing, the neurons have \emph{refractory period} (the period during which the neurons can not respond to further stimulation).  
Thus, we conclude as follow: \\ \\
%
%
\begin{center}
\begin{minipage}{12cm}
\begin{shadebox}
The flux-controlled memristor defined by Eq. (\ref{eqn: mem-i-v-2}) cannot respond to the sinusoidal voltage source $v_{s}(t) = \sin (0.2 t)$ quickly. 
That is, it cannot switch ``on'' rapidly.   Furthermore, this memristor has \emph{refractory period} after switch ``on'', that is, it can not respond to further sinusoidal inputs until the flux is decreased.  
\end{shadebox}
\end{minipage}
\end{center}
%
%
%

\begin{figure}[p]
 \centering
  \begin{tabular}{cc}
   \psfig{file=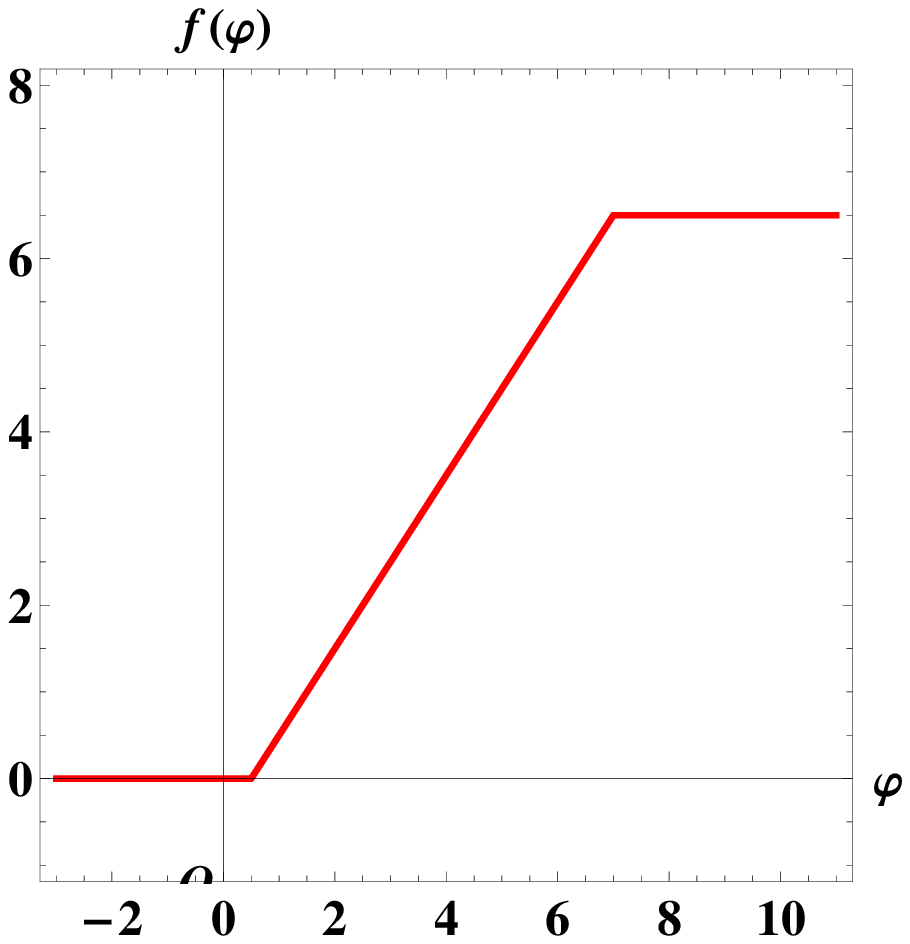, width=6.0cm} & 
   \psfig{file=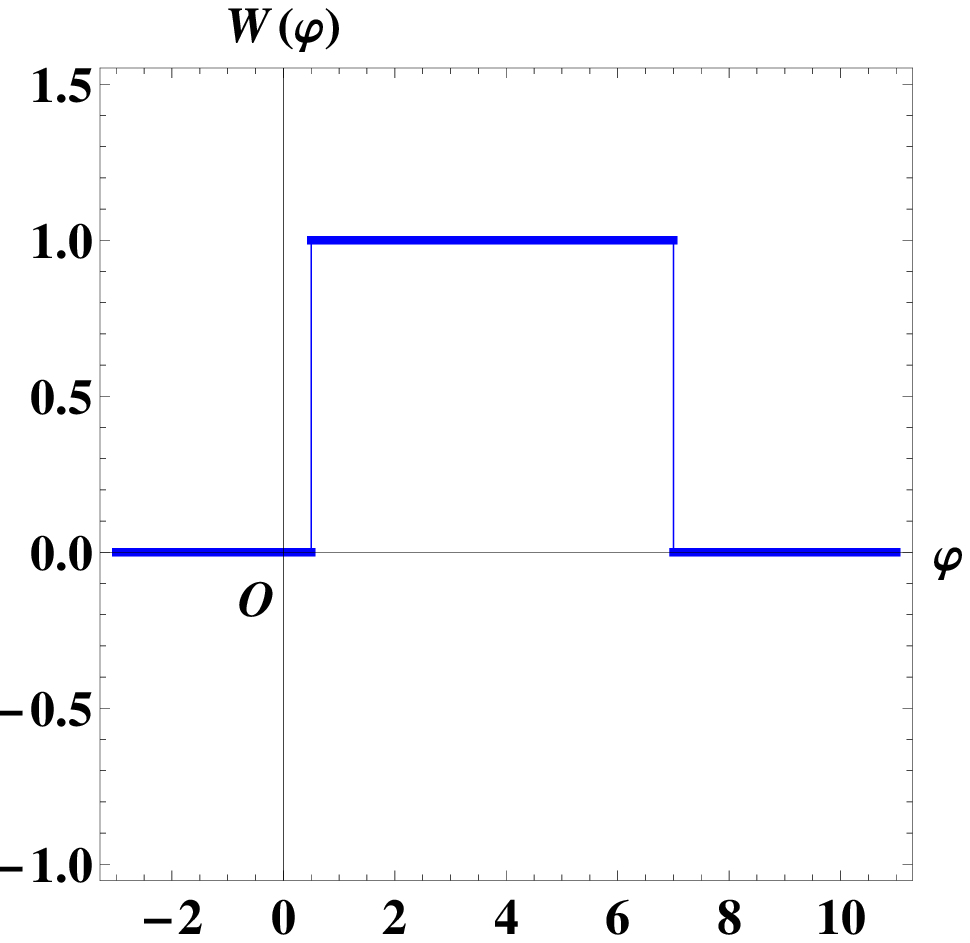, width=6.0cm} \vspace{1mm} \\
   (a) constitutive relation $q = h (\varphi)$ & 
   (b) memductance  $W( \varphi )$ \\
  \end{tabular} 
  \caption{Characteristic of the flux-controlled memristor.  
   The memristor switches off and on depending on the value of the flux $\varphi$ as shown in Figure \ref{fig:mem5}(b). 
   That is, $W( \varphi ) = 0$ implies that the memristor switches off, and $W( \varphi ) = 1$ implies that the memristor switches on.  
   \newline
   (a) The constitutive relation of the memristor, which is given by 
       $q = h(\varphi) \stackrel{\triangle}{=} 0.5 \,( |\varphi - 0.5| - |\varphi - 7| + 6.5)$.  
   \newline
   (b) Memductance $W( \varphi )$ of the  memristor, which is defined by   
       $\displaystyle W( \varphi ) \stackrel{\triangle}{=} \frac{dh(\varphi)}{d\varphi}$.      
       Thus, $W( \varphi ) = 1 $ for $0.5 < \varphi < 7$, 
       and $W( \varphi ) = 0$ for $\varphi \le 0.5$ and $\varphi \ge 7$. }
 \label{fig:mem5}
\end{figure}
%
%

\begin{figure}[p]
 \begin{center} 
  \psfig{file=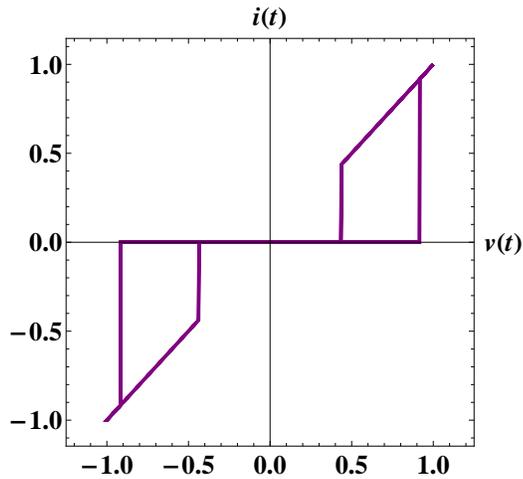,width=7.0cm}
  \caption{Pinched hysteresis loop driven by a periodic voltage source $v_{s}(t) = \sin (\omega t)$, where $\omega = 0.2$.}
 \label{fig:mem6}
 \end{center}
\end{figure}
%
%

\begin{figure}[p]
 \centering
  \begin{tabular}{c}
   \psfig{file=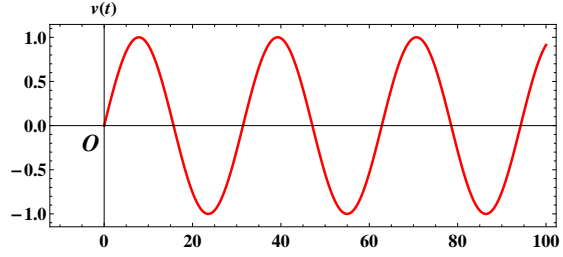, width=8.0cm}  \\
   (a) driving source $v_{s}(t)$  \vspace{1mm} \\
   
   \psfig{file=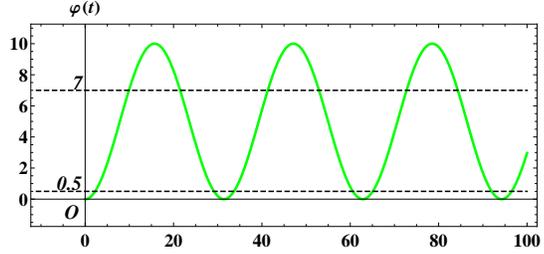, width=8.0cm}  \\
   (b) flux $\varphi (t)$  \vspace{1mm} \\  

   \psfig{file=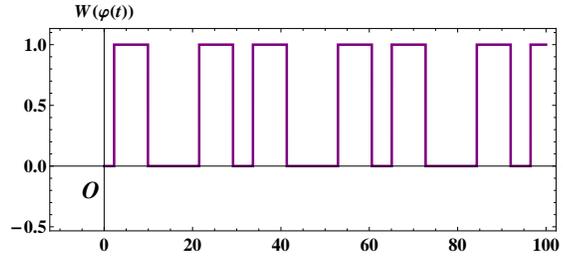, width=8.0cm}  \\
   (c) memductance $W(\varphi (t))$  \vspace{1mm} \\  
   
   \psfig{file=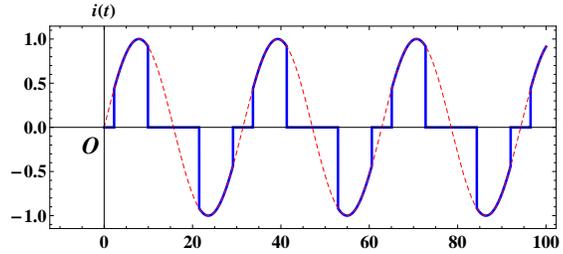, width=8.0cm}  \\
   (d) current $i(t)$ 
  \end{tabular} 
  \caption{Waveforms of the flux-controlled memristor.   
    The memristor switches ``off'' and ``on'', 
    depending on the value of the flux $\varphi$ as shown in Figure \ref{fig:mem7}(c). 
    The current $i(t)$ flows through the memristor 
    if $0.5 < \varphi (t) < 7$,  
    and no current flows through the memristor 
    if $\varphi \le 0.5$ or $\varphi \ge 7$ as shown in Figure \ref{fig:mem7}(d).  
    Initial condition of Eq. (\ref{eqn: dphi-dt}): $\varphi (0) = 0$.
   \newline
   (a) Waveform of the driving source $v_{s}(t)$, 
       which is defined by $v_{s}(t) = \sin (\omega t)$, where $\omega = 0.2$. 
   \newline
   (b) Waveform of the flux $\varphi (t)$ across the memristor, 
       which is defined by $\varphi (t) = \int_{0}^{t} v_{s} (\tau) d \tau $. 
   \newline
   (b) Waveform of the memductance $W(\varphi (t))$, 
       which is defined by $W( \varphi ) = \mathfrak{s}[\varphi + 0.5] - \mathfrak{s}[\varphi - 7] $. 
   \newline
   (d) Waveform of the current $i(t)$ through the memristor, 
       which is defined by $i(t) = W(\varphi)v_{s}(t)$ (blue).   
       Red dashed line denotes the waveform of the driving source $v_{s}(t)$.}
 \label{fig:mem7}
\end{figure}
%
%

\begin{figure}[hpbt]
 \begin{center} 
  \psfig{file=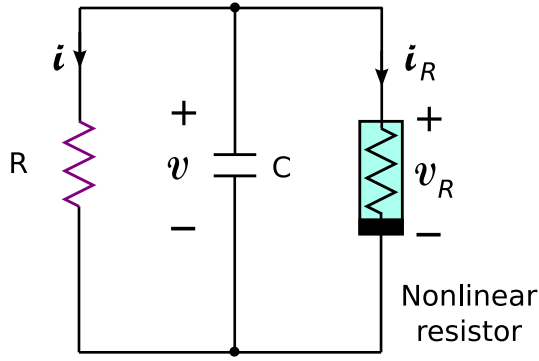,width=7.0cm}
  \caption{Isolated CNN cell which consists of a linear resistor, linear capacitor, and a nonlinear resistor.    
   Parameters: $R=1, \ C=1, \ i_{R} =  f_{R}(v_{R}) \stackrel{\triangle}{=} - 0.5 \, a \, ( |v_{R}+ 1 | -  |v_{R} - 1 | )$, where $a$ is a constant.}
 \label{fig:cell}
 \end{center}
\end{figure}
\newpage

%
%
%
\section{Chaotic Oscillation from Memristor-Coupled CNNs}
\label{sec: oscillation}
%
%
It is well known that an \emph{autonomous} two-cell CNN exhibits a limit cycle, and a second-order \emph{non-autonomous} CNN exhibits a chaotic attractor \cite{{Chua1998}, {Roska}}.   
In this paper, we show that a non-autonomous \emph{memristor-coupled} CNN can also exhibit a chaotic attractor.

%
\subsection{Isolated CNN cell}
\label{sec: isolated}
%
An isolated CNN cell, without the inputs, the outputs from other cells, and the threshold, can be realized by the circuit in Figure \ref{fig:cell}.  
The dynamics of this circuit is given by 
\begin{equation}
  C \frac{dv}{dt} = - \frac{v}{R} -  f_{R}(v_{R}),
\label{eqn: isolated-cell} 
\end{equation}
where $R=1, \ C=1, \ v_{R}=v$.  
The characteristic of the nonlinear resistor is given by 
\begin{equation}
  i_{R}= f_{R}(v_{R}) = - 0.5 \, a \, ( |v_{R}+ 1 | -  |v_{R} - 1 | ),  
\label{eqn: nl-resistor} 
\end{equation}
where $a$ is a constant. If $a>0$, then the nonlinear resistor is active as shown in Figure \ref{fig:nl-1}.    
Substituting Eq. (\ref{eqn: nl-resistor}) into Eq. (\ref{eqn: isolated-cell}), we obtain   
\begin{equation}
  \frac{dv}{dt} = F(v) \stackrel{\triangle}{=} - v + 0.5 \, a \, ( |v + 1 | -  |v  - 1 | ).
\label{eqn: isolated-cell-2} 
\end{equation}
If $a>1$, then Eq. (\ref{eqn: isolated-cell-2}) has three equilibrium points.    
The two equilibrium points are located on both sides of the origin, and they are stable.  
The other one is the origin $O$, which is unstable.   
For example, if $a=2$, then the equilibrium points $P_{1} \, (v= -2)$ and $P_{2} \, (v= 2)$ are stable and the origin is unstable, as shown in Figure \ref{fig:stability}.
Thus, we obtain 
\begin{equation} 
    \begin{cases}
      \text{if~~} v(0)<0  & \displaystyle \lim_{t \to \infty} v(t) \to -2, \\
      \text{if~~} v(0)>0  & \displaystyle \lim_{t \to \infty} v(t) \to 2, \\
      \text{if~~} v(0)=0  & v(t) = 0 \text{~for~all~} t>0. \\
    \end{cases}
\label{eqn: equilibrium-1} 
\end{equation}
The above behavior will be used in Sec. \ref{sec: modified-CNN}.  

Equations (\ref{eqn: isolated-cell}) and (\ref{eqn: isolated-cell-2}) can be recast into the well-known first-order equation of the standard isolated CNN cell 
\begin{center}
\begin{minipage}[t]{8.0cm}
\begin{itembox}[l]{Dynamics of isolated CNN cell}
\begin{equation} 
 \frac{dv}{dt} = - v + a_{0, 0} \, f(v), \vspace{2mm}\\
\end{equation}
\end{itembox}
\end{minipage}
\end{center}
where $a_{0, 0} = a$ and $\displaystyle f(v) = 0.5 \bigl ( \, |v+1|-|v-1| \, \bigr )$ (see Eq. (\ref{eqn: simplest-cell})).

\begin{figure}[h]
 \begin{center} 
  \psfig{file=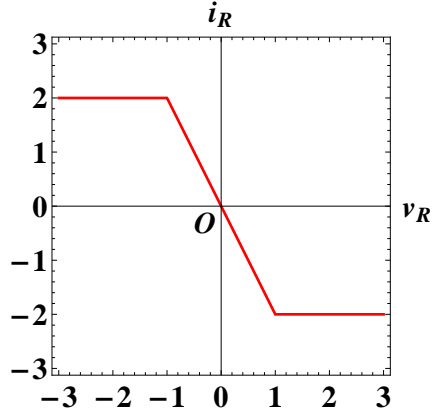,width=6.0cm}
  \caption{$v-i$ characteristic of the nonlinear resistor, which is defined by    
   $i_{R} = f_{R}(v_{R}) \stackrel{\triangle}{=} - 0.5 \, a \, ( |v_{R}+ 1 | -  |v_{R} - 1 | )$, where $a=2$.}
  \label{fig:nl-1}
 \end{center}
\end{figure}
%
%

\begin{figure}[hpbt]
 \begin{center} 
  \psfig{file=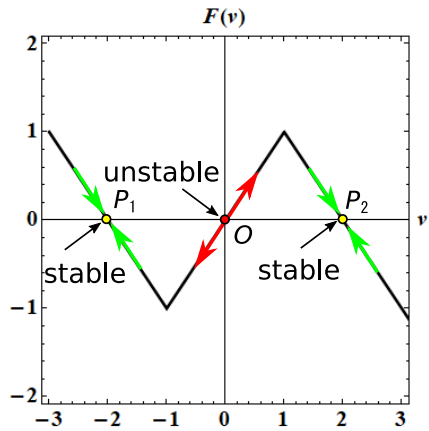,width=6.0cm}
  \caption{Driving-point plot of Eq. (\ref{eqn: isolated-cell-2}) with $a=2$.  
  Equation(\ref{eqn: isolated-cell-2}) has two stable equilibrium points $v = \pm 2$ (yellow), and the origin $v = 0$ (red) is unstable.}
  \label{fig:stability}
 \end{center}
\end{figure}
%
%

%
\subsection{Non-autonomous two-cell CNNs}
\label{sec: standard}
%

Let us consider the \emph{non-autonomous} two-cell CNN in Figure \ref{fig:cell-original} \cite{Chua1998}.
It is driven by an independent current source.    
The dynamics of the circuit is given by 
\begin{equation}
\left. 
 \begin{array}{cll}
  \displaystyle C \frac{dv_{1}}{dt} 
    &=& \displaystyle - \frac{v_{1}}{R} - \, g(v_{1}) - h( v_{2} ) + j(t),\vspace{2mm} \\
  \displaystyle C \frac{dv_{2}}{dt} 
    &=& \displaystyle - \frac{v_{2}}{R} - \, g(v_{2}) + h( v_{1} ), \vspace{2mm} \\   
 \end{array}
\right \} 
\label{eqn: chaotic-cell-original} 
\end{equation}
where 
\begin{equation} 
\left. 
 \begin{array}{cll}
  R  &=& 1, \ C=1, \vspace{2mm} \\
  j(t) &=& 4.04 \sin \left(  \frac{\pi}{2} t \right ), \vspace{2mm} \\
  g(x) &=&  - ( \, |x + 1 | -  |x - 1 | \, ),  \vspace{2mm} \\
  h(x) &=&  0.6 \, ( \, |x + 1 | -  |x - 1 | \,).   \vspace{2mm} \\
 \end{array}   
 \right \}
\end{equation} 
In this circuit, the two nonlinear resistors have the same current-voltage characteristics,   
that is, 
\begin{equation} 
\left.
 \begin{array}{cllll}
   i_{R} &=& g(v_{R}) &=& -  ( \, |v_{R} + 1 | - |v_{R} - 1 | \, ), \vspace{2mm} \\
   i_{r} &=& g(v_{r}) &=& -  ( \, |v_{r} + 1 | - |v_{r} - 1 | \, ). 
 \end{array}
\right \} 
\end{equation} 
The two voltage-controlled current sources are defined by 
\begin{equation}
\left.
 \begin{array}{ccrrr}
   i_{3} &=& - h(v_{2}) &=& - 0.6 \,( \, |v_{2} + 1 | -  |v_{2} - 1 | \, ),  \vspace{2mm} \\
   i_{4} &=&   h(v_{1}) &=&   0.6 \,( \, |v_{1} + 1 | -  |v_{1} - 1 | \, ).
 \end{array}
\right \}
\end{equation} 
Equation (\ref{eqn: chaotic-cell-original}) is recast into the second-order non-autonomous CNN equation \cite{Chua1998}:
\begin{center}
\begin{minipage}[t]{8.5cm}
\begin{itembox}[l]{Dynamics of non-autonomous CNN}
\begin{equation} 
 \begin{array}{cll}
  \displaystyle \frac{dv_{1}}{dt} 
    &=& \displaystyle - v_{1} + a_{0,0} f(v_{1}) + a_{0,1} f( v_{2} ) + j(t),\vspace{2mm} \\
  \displaystyle \frac{dv_{2}}{dt} 
    &=& \displaystyle - v_{2} + a_{0,-1} f(v_{1}) +  a_{0,0} f( v_{2} ), \vspace{2mm} \\   
 \end{array}
\label{eqn: chaotic-cell-recast}  
\end{equation} 
\end{itembox}
\end{minipage}
\end{center}
where $a_{0,0}=2, \ a_{0,-1} = - a_{0,1} = 1.2$, and 
\begin{equation}
 f(x) = 0.5 \bigl ( \, |x+1|-|x-1| \, \bigr ).  
\end{equation}
We show the well-known chaotic trajectory and the associated Poincar\'e map of Eq. (\ref{eqn: chaotic-cell-original}) in Figure \ref{fig:chaos-original} \cite{Chua1998}.

\begin{figure}[hpbt]
 \begin{center} 
  \psfig{file=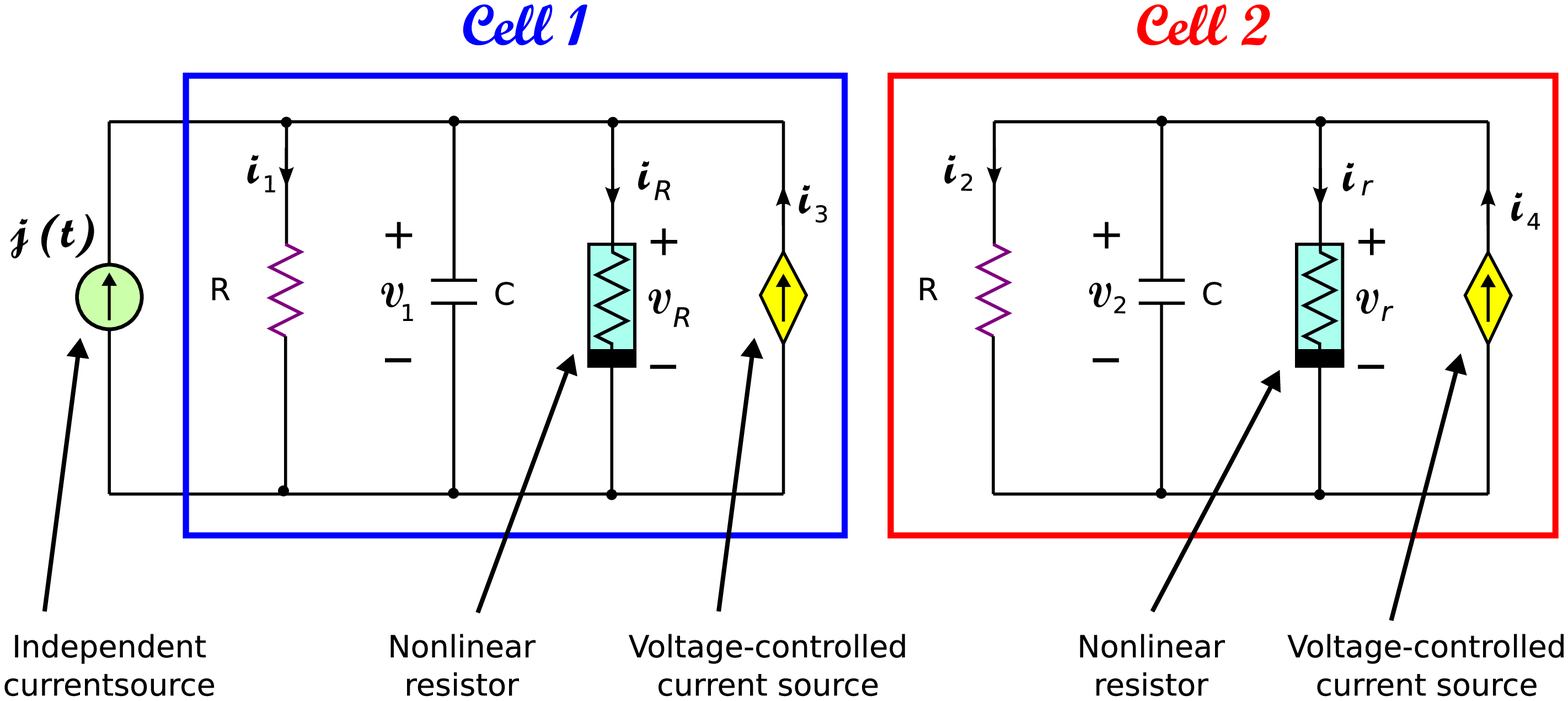,width=14.0cm}
  \caption{Two-cell CNN with an independent current source $j(t) = 4.04 \sin \left(  \frac{\pi}{2} t \right )$.   
   Parameters: $R=1, \ C=1$.  \ 
   \newline
   (1) The two nonlinear resistors (cyan) have the same current-voltage characteristics:  
   $i_{R} = g(v_{R}) \stackrel{\triangle}{=} -  ( |v_{R} + 1 | -  |v_{R} - 1 | )$ (left) and 
   $i_{r} = g(v_{r}) \stackrel{\triangle}{=} -  ( |v_{r} + 1 | -  |v_{r} - 1 | )$ (right).  \
   \newline
   (2) The two voltage-controlled current sources (yellow) are defined by 
   $i_{3} = - h(v_{2}) \stackrel{\triangle}{=} -  0.6 \,( |v_{2} + 1 | -  |v_{2} - 1 | )$ (left) and 
   $i_{4} =   h(v_{1}) \stackrel{\triangle}{=}    0.6 \,( |v_{1} + 1 | -  |v_{1} - 1 | )$ (right). }
 \label{fig:cell-original}
 \end{center}
\end{figure}
%

\begin{figure}[hpbt]
 \centering
  \begin{tabular}{cc}
  \psfig{file=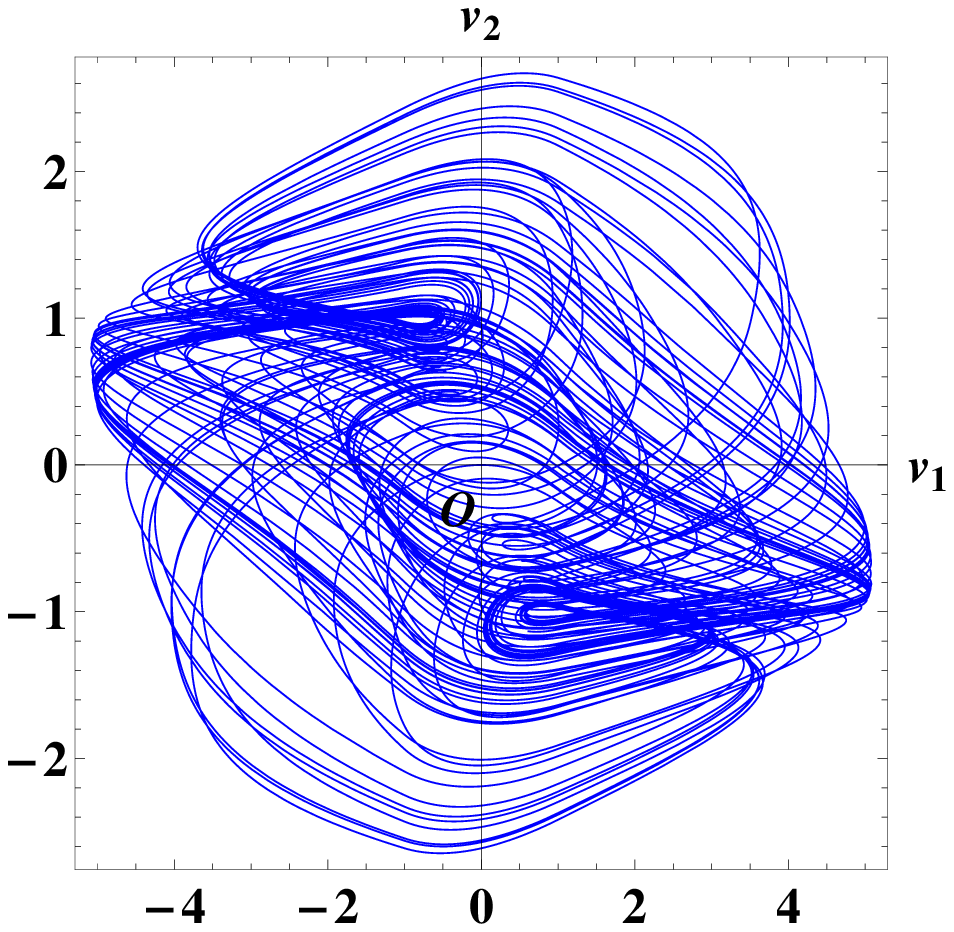,width=5.0cm} & 
  \psfig{file=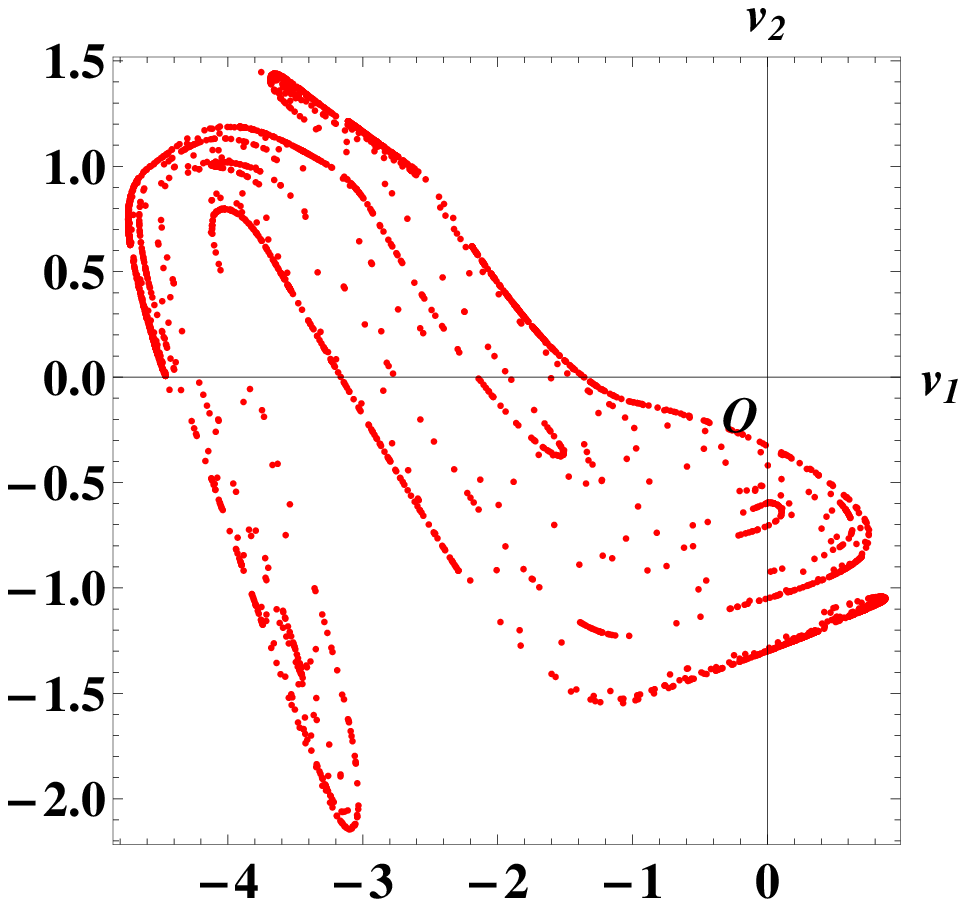,width=5.0cm}  \vspace{1mm} \\
  (a) Chaotic trajectory on the $(v_{1}, \, v_{2})$-plane   & (b) Associated Poincar\'e map     
  \end{tabular}
  \caption{Chaotic trajectory of Eq. (\ref{eqn: chaotic-cell-original}) and associated Poincar\'e map.  
  The attractor in Figure \ref{fig:chaos-original}(b) is called the ``Lady's shoe attractor''.  
  Initial condition: $v_{1}(0)=0, \, v_{2}=0$.}
 \label{fig:chaos-original}
\end{figure}
\newpage

%
\begin{figure}[hpbt]
 \begin{center} 
  \psfig{file=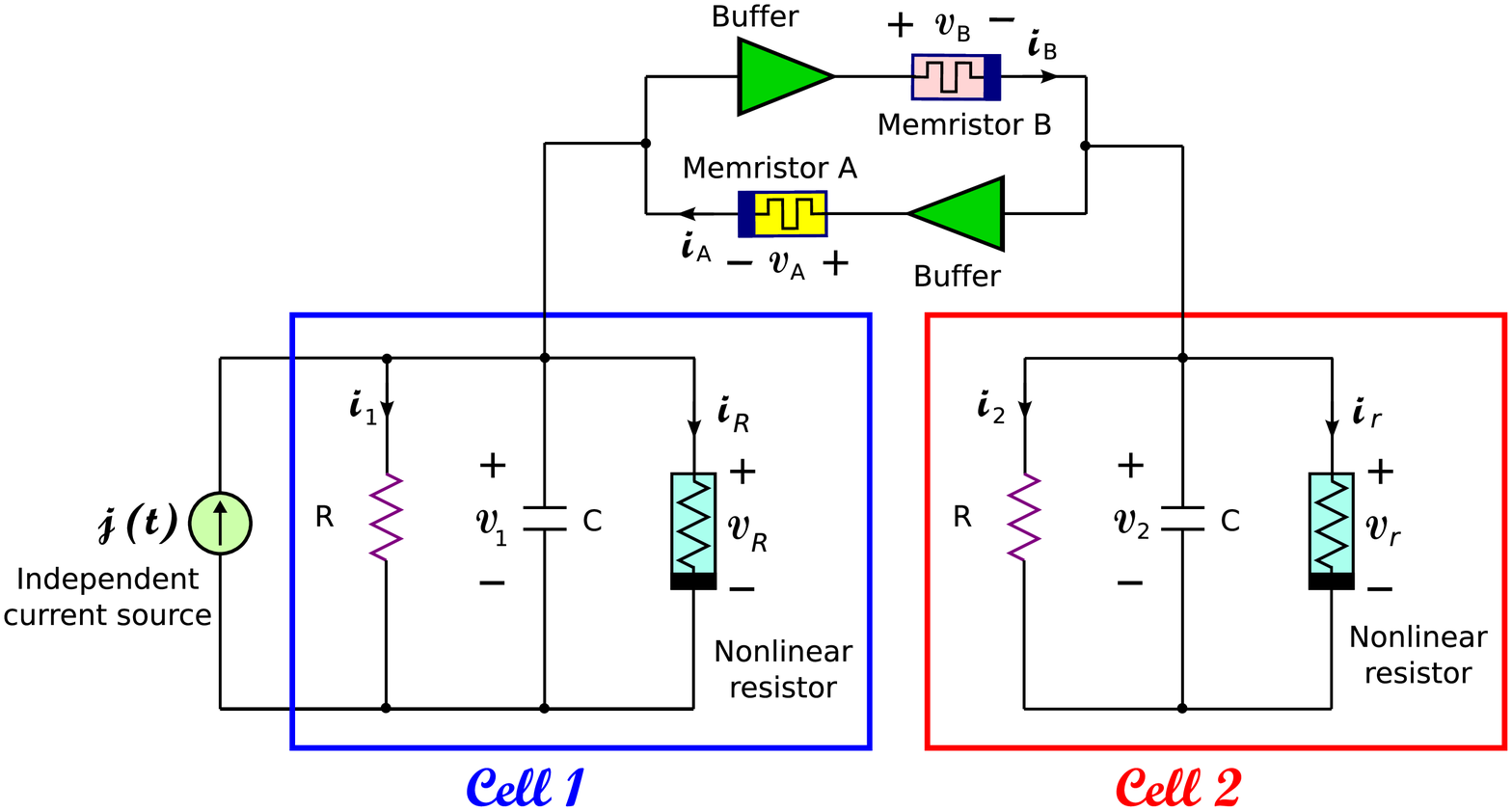,width=13.5cm}
  \caption{Two-cell CNN which is coupled with flux-controlled memristors.  
   \newline
   (1) Parameters: $R=1, \ C=1$. 
   \newline 
   (2) The cell {\it 1} (left) is driven by an independent current source (light-green):
   $j(t) = 4 \sin \left(  \frac{2 \pi}{3} t \right )$.  
   \newline 
   (3) The memristor $A$ (yellow) is passive, while the memristor $B$ (pink) is active.  
   \newline 
   (4) The terminal currents and voltages of the memristor $A$ and the memristor $B$ satisfy 
   $i_{A} = W_{A}( \varphi_{A} ) \, v_{A} = W_{A}( \varphi_{A} )(v_{2} - v_{1})$ and 
   $i_{B} = W_{B}( \varphi_{B} ) \, v_{B} = W_{B}( \varphi_{B} )(v_{1} - v_{2})$, respectively, 
   where 
   $\displaystyle \frac{d\varphi_{A}}{dt} = v_{A} = v_{2} - v_{1}$, 
   $\displaystyle \frac{d\varphi_{B}}{dt} = v_{B} = v_{1} - v_{2}$, 
   $W_{A}( \varphi_{A} ) = - 1.2 \,(\mathfrak{s}[\varphi _{A}+ 2] - \mathfrak{s}[\varphi_{A} - 2])$, and 
   $W_{B}( \varphi_{B} ) =   1.2 \,( \mathfrak{s}[\varphi_{B} + 2] + \mathfrak{s}[\varphi_{B} - 2])$.  
   \newline
   The symbol $\mathfrak{s}  [\, z \,]$ denotes the \emph{unit step} function, 
   equal to $0$ for $z < 0$ and 1 for $z \ge 0$.  
   \newline
   (5) The two nonlinear resistors (cyan) have the same current-voltage characteristics, 
   that is, 
   $i_{R} = g(v_{R}) \stackrel{\triangle}{=} -  ( |v_{R} + 1 | -  |v_{R} - 1 | )$ and 
   $i_{r} = g(v_{r}) \stackrel{\triangle}{=} -  ( |v_{r} + 1 | -  |v_{r} - 1 | )$. 
   \newline
   (6) The buffer (dark green) is a op-amp circuit which has a voltage gain of $1$, 
   that is, the output voltage is the same as the input voltage.  
   It offers input-output isolation.    }
 \label{fig:cell-chaos}
 \end{center}
\end{figure}
%
%

%
\subsection{Memristor-coupled two-cell CNNs}
\label{sec: memristor-coupled}
%
Let us consider the non-autonomous CNN in Figure \ref{fig:cell-chaos}, whose cells are coupled with memristors.  
The dynamics of the circuit in Figure \ref{fig:cell-chaos} is given by 
\begin{equation}
\left. 
 \begin{array}{cll}
  \displaystyle C \frac{dv_{1}}{dt} 
    &=& \displaystyle - \frac{v_{1}}{R} + g(v_{1}) + W_{A}( \varphi_{A} )(v_{2} - v_{1}) + j(t),\vspace{2mm} \\
  \displaystyle C \frac{dv_{2}}{dt} 
    &=& \displaystyle - \frac{v_{2}}{R} + g(v_{2}) + W_{B}( \varphi_{B} )(v_{1} - v_{2}), \vspace{2mm} \\
  \displaystyle\frac{ d\varphi_{A}}{dt}
    &=& v_{A} = v_{2} - v_{1},  \vspace{2mm} \\
  \displaystyle\frac{ d\varphi_{B}}{dt}
    &=& v_{B} = v_{1} - v_{2},      
 \end{array}
\right \} 
\label{eqn: chaotic-cell} 
\end{equation}
where 
\begin{equation} 
\left. 
 \begin{array}{cll}
   R  &=& 1, \ C=1, \vspace{2mm} \\
   j(t) &=& 4 \sin \left(  \frac{2 \pi}{3} t \right ), \vspace{2mm} \\
   g(x) &=& - ( |x + 1 | -  |x - 1 | ),   \vspace{2mm} \\

   W_{A}( \varphi_{A} ) 
    &=& - 1.2 \,( \, \mathfrak{s}[\varphi _{A}+ 2] -   \mathfrak{s}[\varphi_{A} - 2] \, ) \vspace{2mm} \\
    &=& \left \{ 
    \begin{array}{clcc}
      -1.2 & \ for \ & \ & -2 \le \varphi_{A} < 2,  \vspace{2mm} \\
      0 & \ for \ & \ & \varphi_{A} < -2  \text{~and~}  2 \le \varphi_{A}, 
    \end{array}
    \right.   \vspace{2mm} \\

   W_{B}( \varphi_{B} )  
    &=& 1.2 \,( \, \mathfrak{s}[\varphi_{B} + 2] - \mathfrak{s}[\varphi_{B} - 2] \, ) \vspace{2mm} \\
    &=& \left \{ 
    \begin{array}{clcc}
        1.2 & \ for \ & \ & -2 \le \varphi_{B} < 2,  \vspace{2mm} \\
        0 & \ for \ & \ & \varphi_{B} < -2  \text{~and~} 2 \le \varphi_{B}.  
    \end{array}
    \right.   
 \end{array}
 \right \}
\end{equation} 
Here, $\varphi_{A}$ and $\varphi_{B}$ denote the flux of the memristors $A$ and $B$, respectively, 
$W_{A}( \varphi_{A} )$ and  $W_{B}( \varphi_{B} )$ denote the memductances of the memristors $A$ and $B$, respectively,  
$j(t)$ denotes an independent current source,   
and $\mathfrak{s}  [\, z \,]$ denotes the \emph{unit step} function, equal to $0$ for $z < 0$ and 1 for $z \ge 0$.  
The two nonlinear resistors have the same current-voltage characteristics,   
that is, 
\begin{equation} 
 \begin{array}{cll}
   i_{R} &=& g(v_{R}) = -  ( |v_{R} + 1 | -  |v_{R} - 1 | ), \vspace{2mm} \\
   i_{r} &=& g(v_{r}) = -  ( |v_{r} + 1 | -  |v_{r} - 1 | ). 
 \end{array}
\end{equation} 
The terminal currents and voltages of the memristors $A$ and $B$ satisfy the relation
\begin{equation} 
\left. 
 \begin{array}{cllll}
   i_{A} &=& W_{A}( \varphi_{A} ) \, v_{A} &=& W_{A}( \varphi_{A} ) \, (v_{2} - v_{1}),  \vspace{2mm} \\
   i_{B} &=& W_{B}( \varphi_{B} ) \, v_{4} &=& W_{B}( \varphi_{B} ) \, (v_{1} - v_{2}),   
 \end{array}
\right \}
\end{equation} 
respectively.  

We show the constitutive relations and memdauctances of the flux-controlled memristors in Figures \ref{fig:phi-q-3-4} and \ref{fig:W-3-4}, respectively.  
Equation (\ref{eqn: chaotic-cell}) is recast into the form: 
\begin{center}
\begin{minipage}[t]{8.7cm}
\begin{itembox}[l]{Dynamics of memristor-coupled two-cell CNN}
\begin{equation}
 \begin{array}{cll}
  \displaystyle \frac{dv_{1}}{dt} 
    &=& \displaystyle - v_{1} + g(v_{1}) + W_{A}( \varphi_{A} )(v_{2} - v_{1}) + j(t),\vspace{2mm} \\
  \displaystyle \frac{dv_{2}}{dt} 
    &=& \displaystyle - v_{2} + g(v_{2}) + W_{B}( \varphi_{B} )(v_{1} - v_{2}), \vspace{2mm} \\
  \displaystyle\frac{ d\varphi_{A}}{dt}
    &=& v_{A} = v_{2} - v_{1},  \vspace{2mm} \\
  \displaystyle\frac{ d\varphi_{B}}{dt}
    &=& v_{B} = v_{1} - v_{2}.        
 \end{array}
\end{equation}
\end{itembox}
\end{minipage}
\end{center}

Note that the memristor $A$ is active since $W_{A}( \varphi_{A} ) \le 0$. 
On the contrary, the memristor $B$ is passive since $W_{B}( \varphi_{B} ) \ge 0$.  
Observe that the memristors switch ``off'' and ``on'' at irregular intervals as shown in Figure \ref{fig:wave-3-4}. 
Furthermore, the two cells are connected when either or both of the memristors switches ``on''.
We show the trajectory of Eq. (\ref{eqn: chaotic-cell}) and its associated Poincar\'e map in Figure \ref{fig:chaos-1}.  
Observe that Eq. (\ref{eqn: chaotic-cell}) can exhibit a chaotic attractor. 
Thus, we conclude as follow:
%
%
\begin{center}
\begin{minipage}{12cm}
\begin{shadebox}
The non-autonomous memristor-coupled two-cell CNN defined by Eq. (\ref{eqn: chaotic-cell}) can exhibit a chaotic attractor.  
The two flux-controlled memristors in Figure \ref{fig:cell-chaos} switch ``off'' and ``on'' at irregular intervals. 
Furthermore, the two cells are connected when either or both of the memristors switches ``on''. 
 \end{shadebox}
\end{minipage}
\end{center}
%
%

\begin{figure}[htbp]
 \centering
  \begin{tabular}{cc}
  \psfig{file=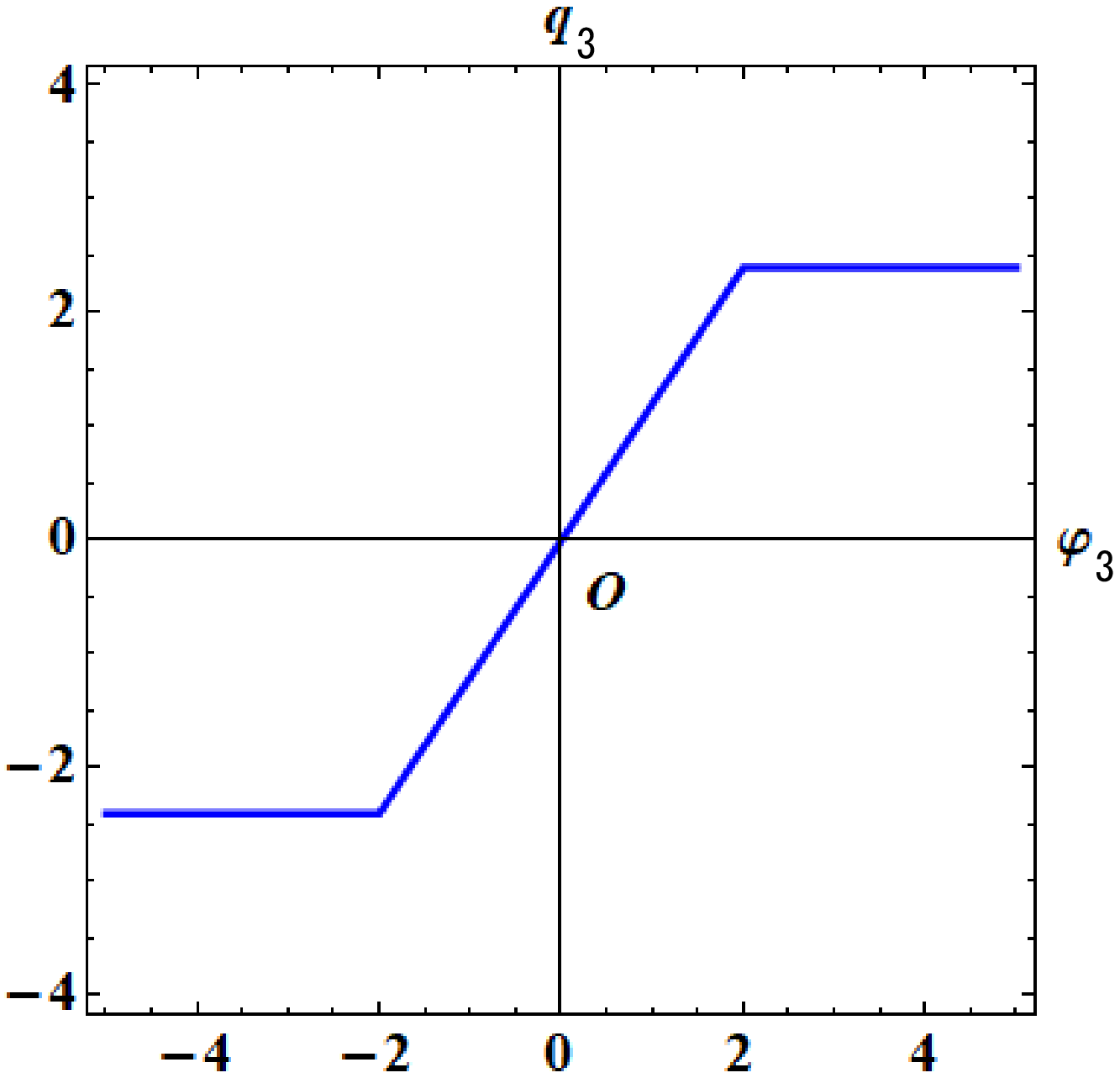,width=6.0cm} &
  \psfig{file=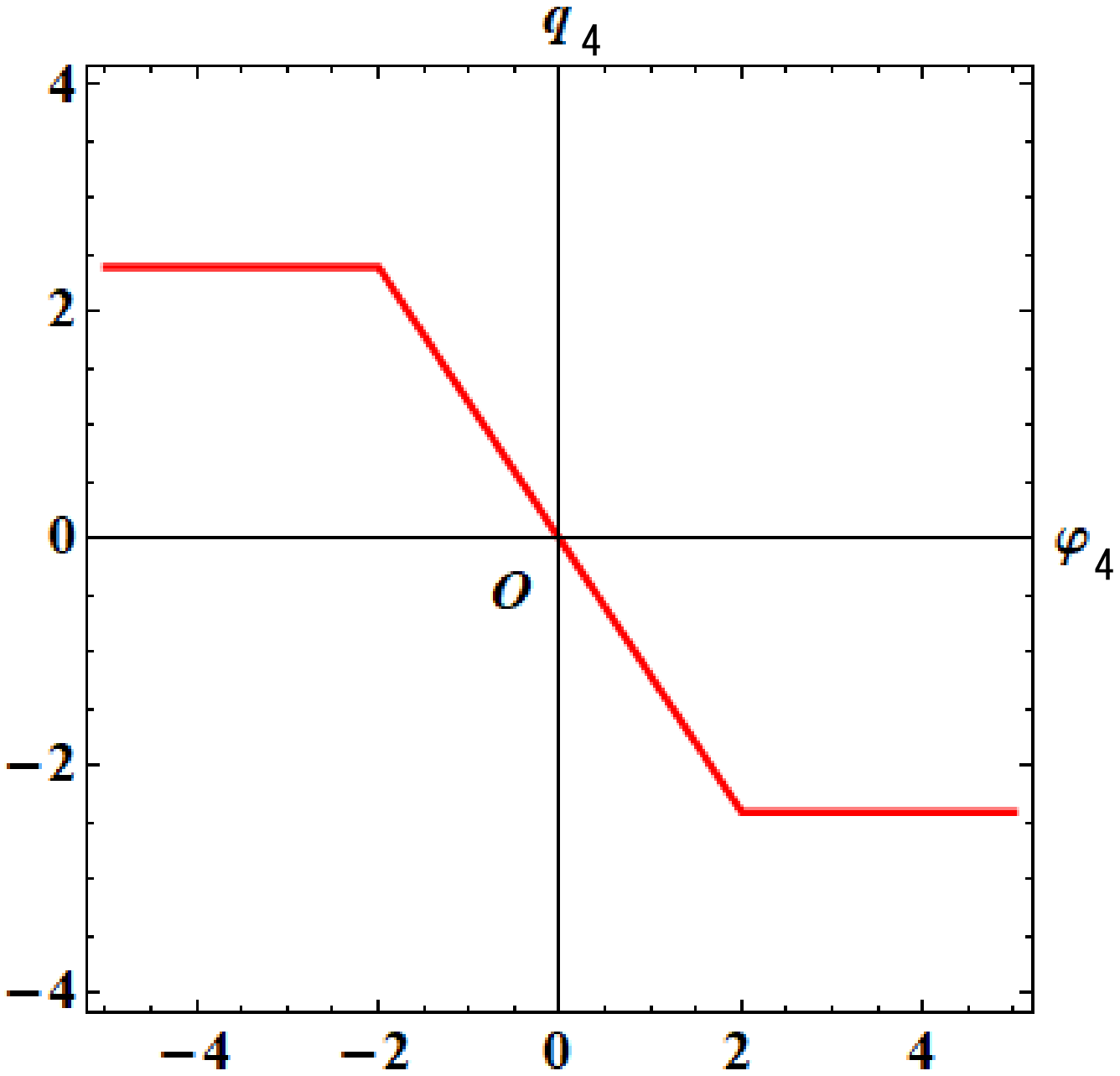,width=6.0cm}\vspace{1mm} \\
  (a) passive memristor $A$ & (b) active memristor $B$     
  \end{tabular}  
  \caption{Constitutive relations of the memristor $A$ and the memristor $B$.}
  \label{fig:phi-q-3-4} 
\end{figure}
%
%

\begin{figure}[htbp]
 \centering
  \begin{tabular}{cc}
  \psfig{file=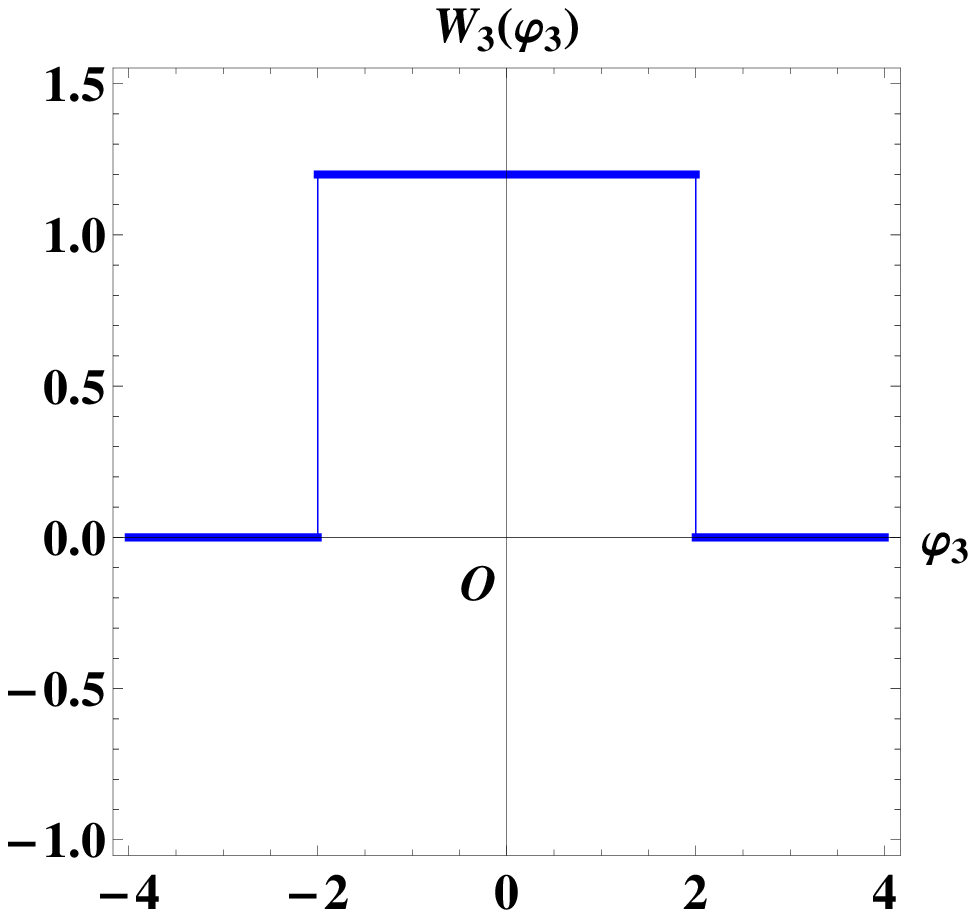,width=6.0cm} & 
  \psfig{file=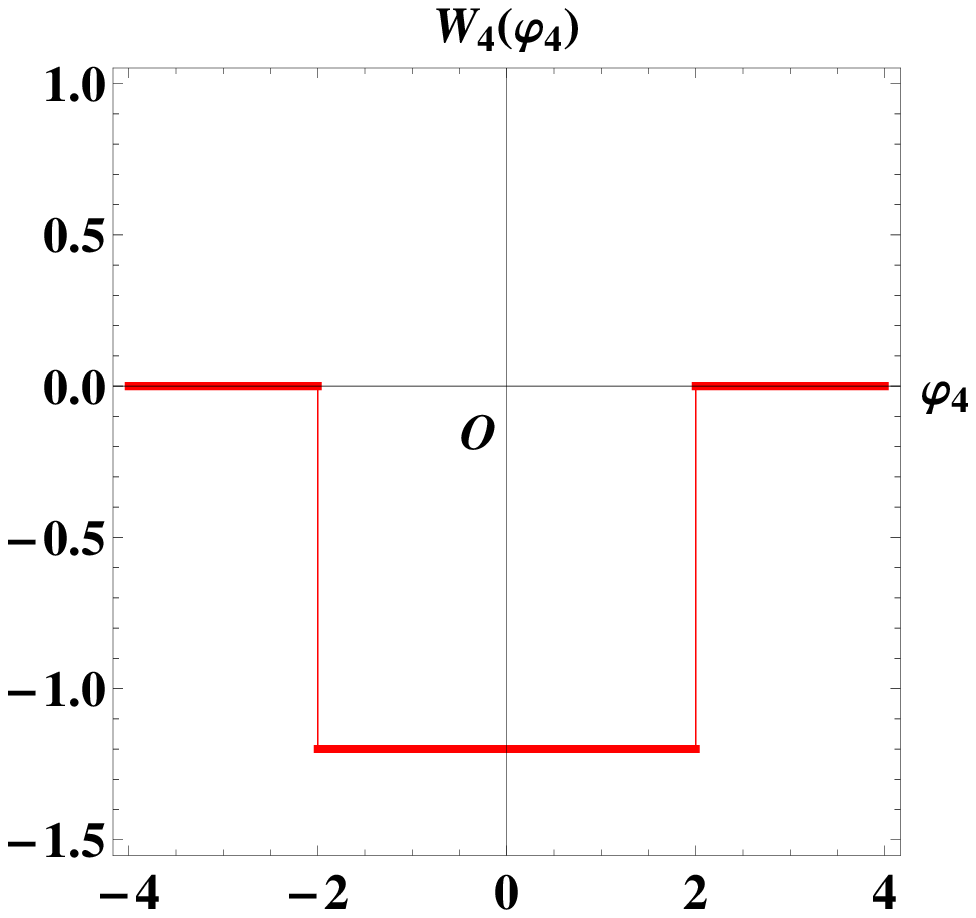,width=6.0cm}  \vspace{1mm} \\
  (a) $W_{3}( \varphi_{3} )$    & (b) $W_{4}( \varphi_{4} )$     
  \end{tabular}
  \caption{The memductance $W_{3}( \varphi_{3} )$ of the passive memristor $A$ 
   and the memductance $W_{4}( \varphi_{4} )$ of the active memristor $B$.   }
 \label{fig:W-3-4}
\end{figure}
%
%

\begin{figure}[htbp]
 \centering
  \begin{tabular}{c}
  \psfig{file=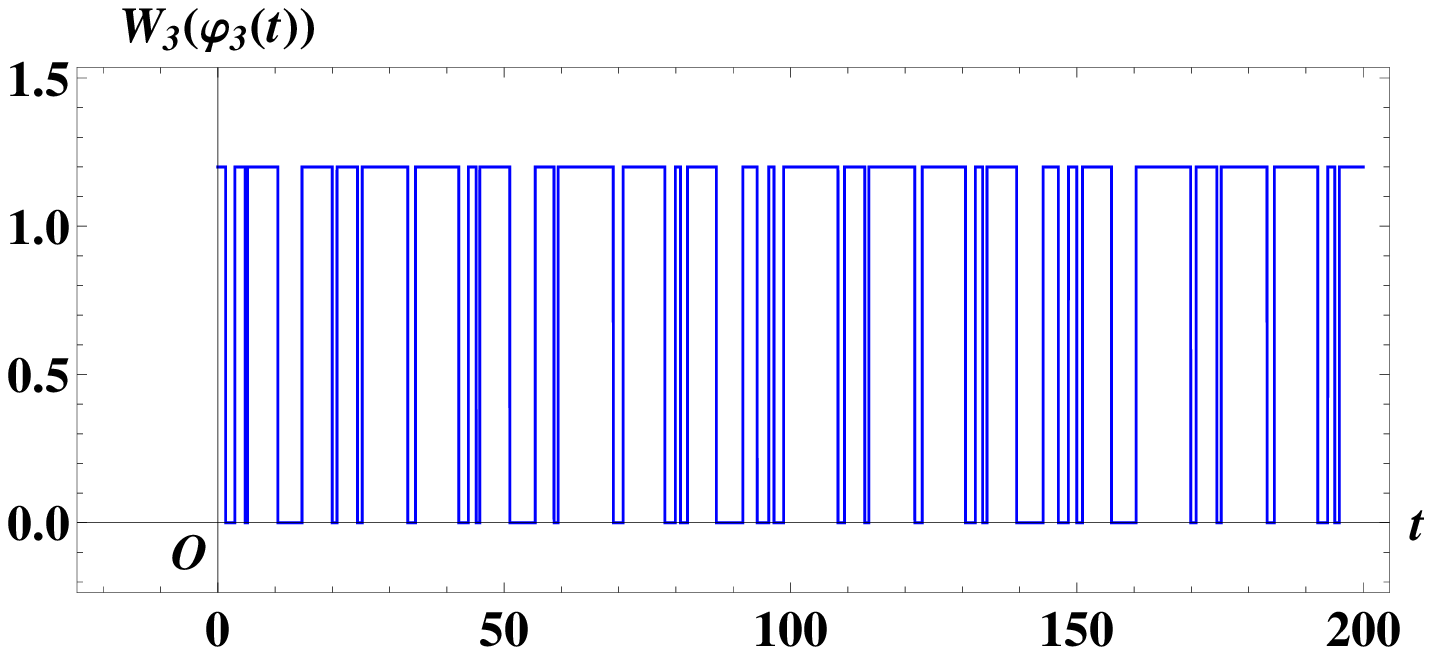,width=10.0cm}  \\
   (a) $W_{3}( \varphi_{3}(t) )$            \vspace{1mm} \\
  \psfig{file=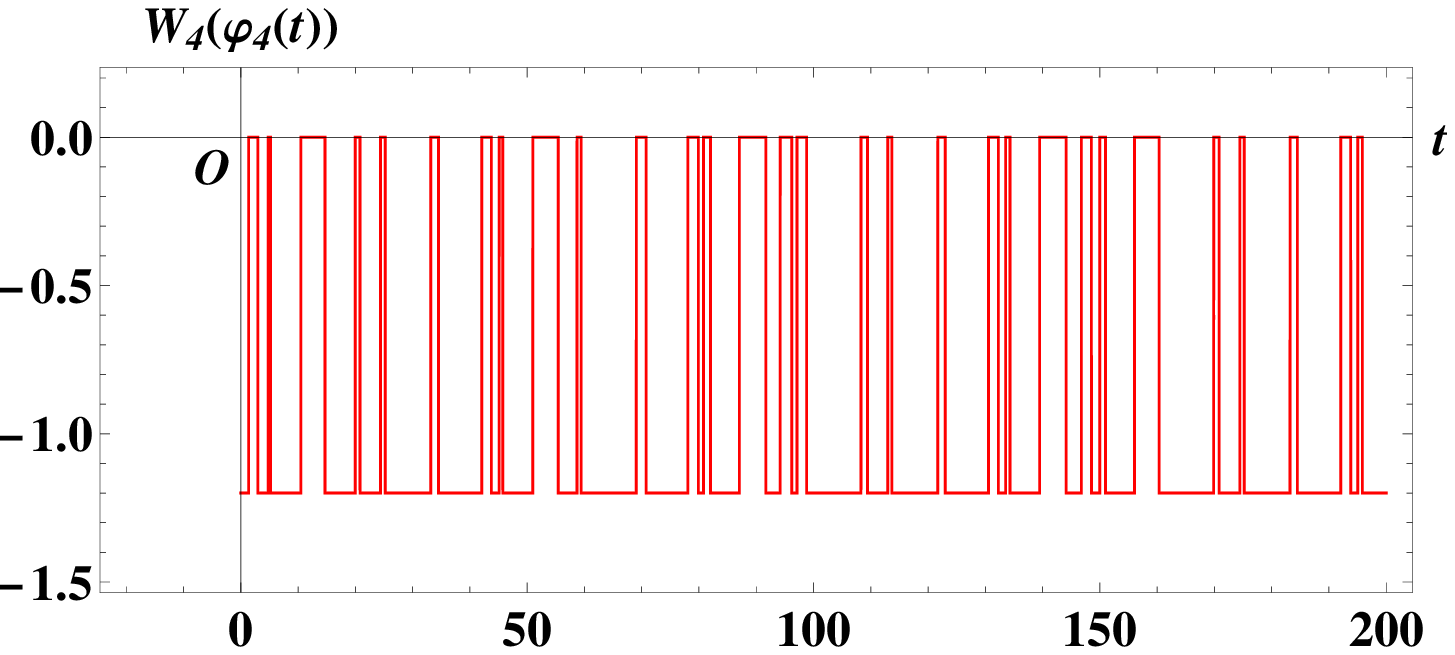,width=10.0cm}   \\
   (b) $W_{4}( \varphi_{4}(t) )$     
  \end{tabular}
  \caption{Waveforms of the memductance $W_{3}( \varphi_{3} )$ of the passive memristor $A$ (top)
   and the memductance $W_{4}( \varphi_{4} )$ of the active memristor $B$ (bottom).  
   These memristors switch ``off'' and ``on'' irregularly.  }
 \label{fig:wave-3-4}
\end{figure}
%
%

\begin{figure}[htbp]
 \centering
  \begin{tabular}{cc}
  \psfig{file=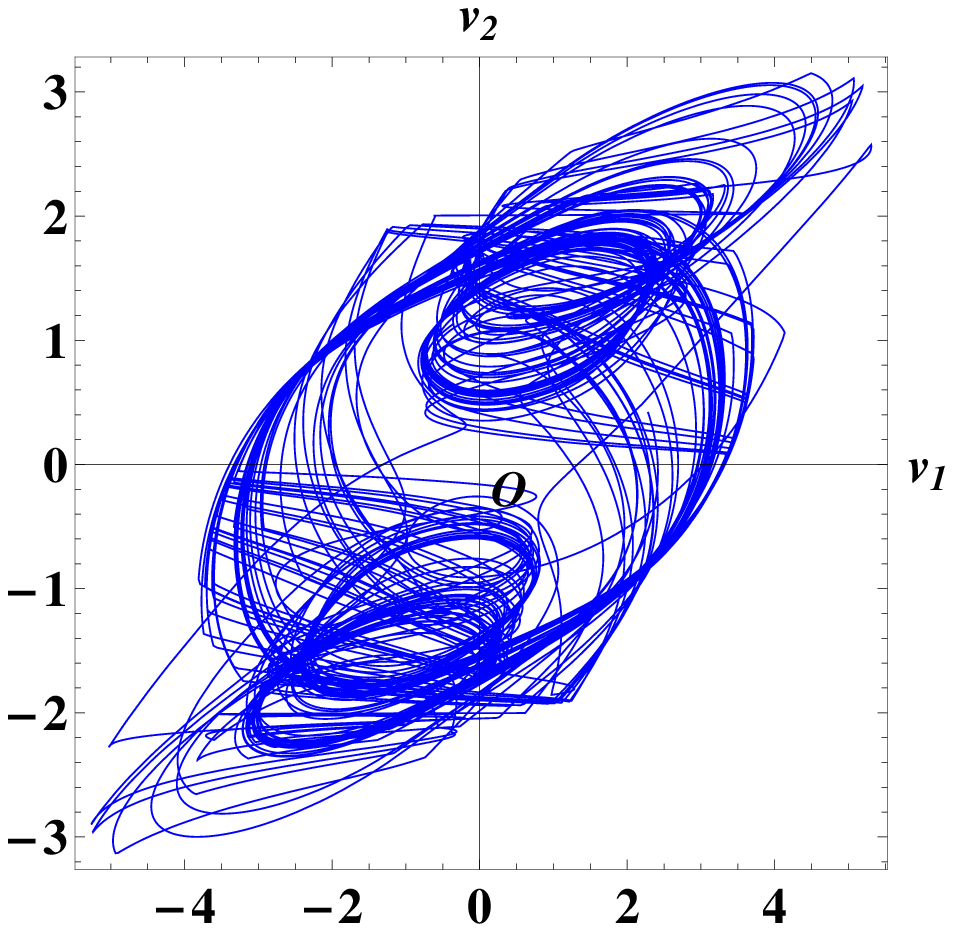,width=7.0cm} & 
  \psfig{file=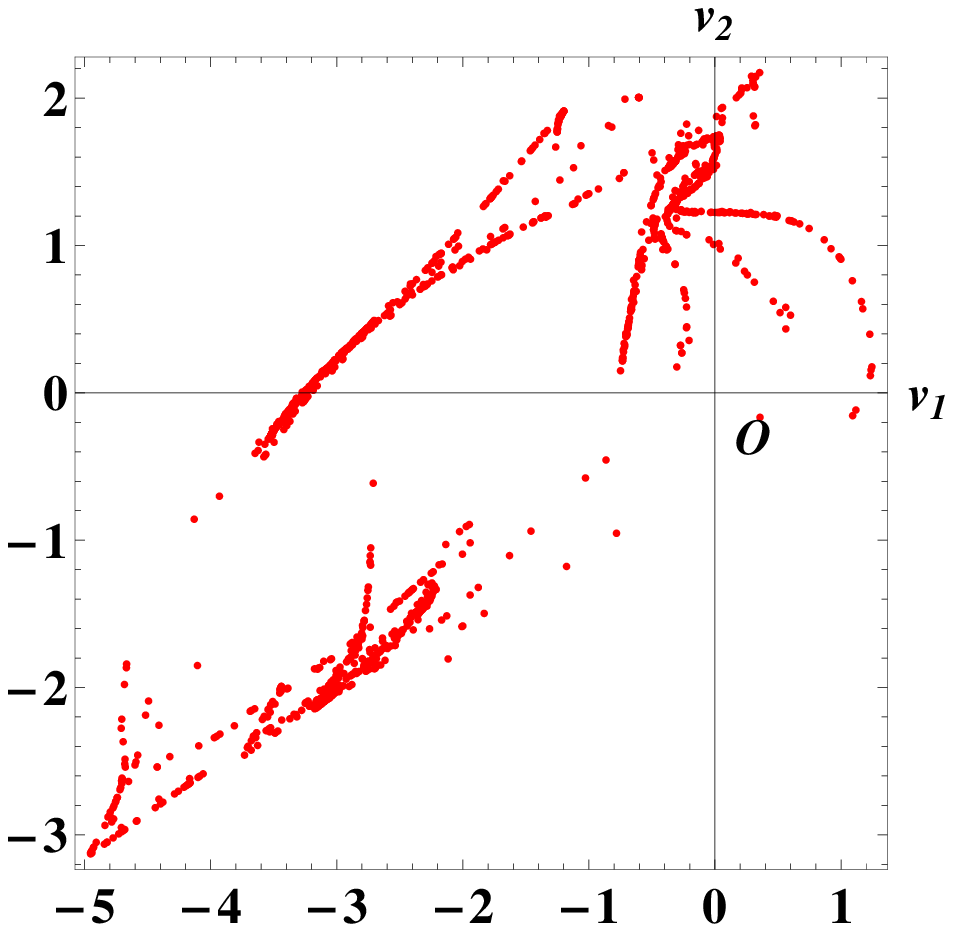,width=7.0cm}  \vspace{1mm} \\
  (a) Chaotic trajectory on the $(v_{1}, \, v_{2})$-plane   & (b) Associated Poincar\'e map     
  \end{tabular}
  \caption{Chaotic trajectory of Eq. (\ref{eqn: chaotic-cell}) and associated Poincar\'e map.  
  Compare the two attractors in Figures \ref{fig:chaos-original} and \ref{fig:chaos-1}.  
  Initial condition: $v_{1}(0)=0, \, v_{2}=0, \, \varphi_{3}(0)=0, \, \varphi_{4}(0)=0$.}
 \label{fig:chaos-1}
\end{figure}

\newpage
%
\subsection{Similarity between memristors and neurons}
\label{sec: similarity}
%
In this subsection, we show that there is the similarity between memristors and neurons.  
The neuron has an ``excitatory'' synapse and an ``inhibitory'' synapse.  
Synapses are junctions that allow a neuron to transmit a signal to another cell. 
They can either be excitatory or inhibitory.  
The inhibitory synapses decrease the likelihood of the firing action potential of a cell, 
while the excitatory synapses increase its likelihood of the firing action potential of a cell. 

The memristors in Figure \ref{fig:cell-chaos} transmit signals from one cell to another cell at irregular intervals.  
The instantaneous powers of the memristor $A$ and $B$ are given by 
\begin{equation}
 \begin{array}{lll}
  p_{A}(t) &\stackrel{\triangle}{=}& v_{A}(t) \, i_{A}(t) \vspace{1mm} \\
           &=& W_{A}( \varphi_{A} (t)) \, (v_{2}(t) - v_{1}(t))^{2} \le 0,  
 \end{array}
\end{equation}
and
\begin{equation}
 \begin{array}{lll}
  p_{B}(t) &\stackrel{\triangle}{=}&  v_{B}(t) \, i_{B}(t) \vspace{1mm} \\
           &=& W_{B}( \varphi_{B}(t) ) \, (v_{1}(t) - v_{2}(t))^{2} \ge 0, 
 \end{array}
\end{equation}
respectively, where  $W_{A}( \varphi_{A} ) \le 0$ and $W_{B}( \varphi_{B} ) \ge 0$. 

It follows that the instantaneous power $p_{B}(t)$ \emph{flows into} the cell {\it 2}.     
Similarly, the instantaneous power $p_{A}(t)$ flows into the cell {\it 1} with the current source $j(t)$, that is, $ - p_{A}(t)$ \emph{flows out} of it, since $p_{A}(t)$ is negative.\footnote{Note the direction of the power flow and the buffer in Figure \ref{fig:cell-chaos} provides an input-output isolation.}
Thus, the memristor $B$ corresponds to the ``excitatory synapse'', 
and the memristor $A$ corresponds to the ``inhibitory synapse''.    
The chaotic oscillation of Eq. (\ref{eqn: chaotic-cell}) depends on a delicate balance between the powers $p_{A}(t)$ and $p_{B}(t)$.

%
\begin{figure}[ht]
 \begin{center} 
  \psfig{file=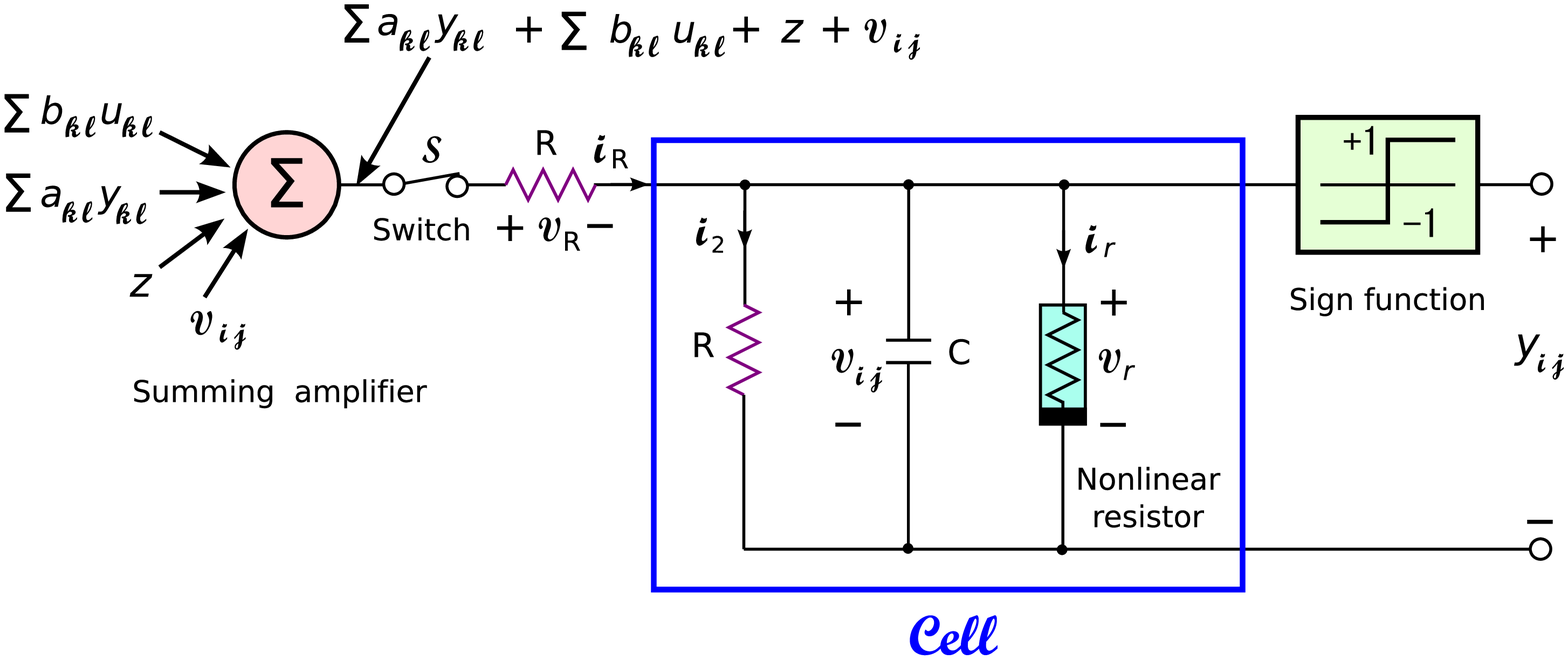,width=15cm}
  \caption{Modified CNN circuit.  The switch $S$ is used to disconnect the cell from the summing amplifier (pink). 
   \newline
   (1) Parameters: $R=1, \ C=1$.  
   \newline
   (2) The $v-i$ characteristic of the nonlinear resistor (light blue) is given by 
   $i_{r} =  f_{r}(v_{r}) \stackrel{\triangle}{=} - 0.5 \, a \, ( |v_{r}+ 1 | -  |v_{r} - 1 | )$, 
   where $a$ is a constant. 
   \newline 
   (3) The output voltage of the summing amplifier is given by  
   $\displaystyle \sum_{k, \, l \in N_{ij}, \  k \ne i, \ l \ne j} a_{kl} \, \operatorname {sgn} (v_{kl}) 
   + \sum_{k, \, l \in N_{ij}}b_{k l} \ u_{kl} + z + v_{ij}$. 
   \newline  
   (4) The above output voltage contains the voltage $v_{ij}$ of the capacitor $C$ (the last term).     
   \newline
   (5) The above sum 
   $\displaystyle \sum_{k, \, l \in N_{ij}, \  k \ne i, \ l \ne j} a_{kl} \, \operatorname {sgn} (v_{kl}) $ 
   does \emph{not} contain the term 
   $ a_{ij} \, \operatorname {sgn} (v_{ij})$ \ ($k=i, \ l=j$).  
   \newline
   (6) The output $y_{ij}$ and the state $v_{ij}$ of each cell are related via the sign function (green): 
    $y_{ij} = \operatorname {sgn} (v_{ij})$.  
   \newline
   (7) The two linear resistors (purple) have the same resistance $R$.  
   Thus, we used the same symbol $R$ and color. 
   }
 \label{fig:CNN-model}
 \end{center}
\end{figure}
%
%

%
%
%
\section{Modified CNN}
\label{sec: modified-CNN}
%
%
%
Consider the modified CNN shown in Figure \ref{fig:CNN-model}.  
In this system, the output $y_{ij}$ and the state $v_{ij}$ of each cell is related via the sign function\footnote{The sign function can be approximated by using the saturation non-linearity of the Op amp, and by normalizing its output voltage.} 
\begin{equation}
  y_{ij} = \operatorname {sgn} (v_{ij}) 
  \stackrel{\triangle}{=} 
    \begin{cases}
      1  & v_{ij}>0, \\
      0  & v_{ij}=0, \\
     -1  & v_{ij}<0.
    \end{cases}
\end{equation}
The switch $S$ is used to disconnect all signals, namely, the output $y_{kl}$, the input $u_{kl}$, the threshold $z$, and the state $v_{ij}$. 

Let us calculate the voltage $v_{R}$ across the resistor $R$, which is connected to the switch $S$. 
It is given by 
\begin{equation}
 \begin{array}{l}
  \scalebox{0.83}{$\displaystyle v_{R} =\displaystyle  \displaystyle  \left ( \sum_{k, \, l \in N_{ij}, \ k \ne i, \ l \ne j}
         a_{k l} \ y_{kl} +  \sum_{k, \, l \in N_{ij}} b_{k l} \ u_{kl} + z + v_{ij} \right )  - v_{ij} $} \vspace{3mm} \\
   \scalebox{0.9}{$\displaystyle  =  \displaystyle  \left ( \sum_{k, \, l \in N_{ij}, \ k \ne i, \ l \ne j}
         a_{k l} \ y_{kl} + \displaystyle  \sum_{k, \, l \in N_{ij}} b_{k l} \ u_{kl} + z \right ).  $}
 \end{array}
\end{equation}
Thus, the dynamics of the modified CNN is given by 
\begin{equation}
 \begin{array}{l}
  \displaystyle C \frac{dv_{ij}}{dt} 
   = \displaystyle - \frac{v_{ij}}{R} - i_{r} + \frac{v_{R}}{R} \vspace{3mm} \\
 
   = \displaystyle - \frac{v_{ij}}{R} - f_{r}(v_{r}) 
     \scalebox{0.9}{$\displaystyle 
         + \ \displaystyle \frac{ \displaystyle  \left ( \sum_{k, \, l \in N_{ij}, \ k \ne i, \ l \ne j}
         a_{k l} \ y_{kl} + \sum_{k, \, l \in N_{ij}} b_{k l} \ u_{kl} + z \right )  }{R}, 
    $} \vspace{4mm} \\  
    \hspace*{10mm}  (i, \, j) \in \{ 1, \cdots , M \} \times \{ 1, \cdots , N \},  
 \end{array}
\label{eqn: cnn10} 
\end{equation}
where 
\begin{enumerate}
\renewcommand{\labelenumi}{(\alph{enumi})}
\item $C=R=1$.  \\
\item $v_{ij}$ denotes the voltage across the capacitor $C$.  \\
\item The symbols $y_{kl}$ and $u_{kl}$ denote the output and input of cell $C_{ij}$, respectively. \\
\item The symbol $N_{ij}$ denotes the $r$-neighborhood of cell $C_{ij}$. \\ 
\item The symbols  $a_{kl}, \, b_{kl}$, and $z$ denote the feedback, control, and threshold template parameters, respectively. 
The matrices $A = [ a_{kl} ]$ and $B = [ b_{kl} ]$ are referred to as the feedback template $A$ and the feed-forward (input) template $B$, respectively. \\
\item 
The characteristic of the nonlinear resistor is given by 
\begin{equation}
  i_{r} = f_{r}(v_{r}) = - 0.5 \, a \, ( |v_{r}+ 1 | -  |v_{r} - 1 | ),  
\label{eqn: nl-resistor-2} 
\end{equation}
where $a$ is a constant.\footnote{If we use the standard piecewise-linear nonlinearity defined by Eq. (\ref{eqn: nl-resistor-2}), then we can apply almost all templates of the CNN, since the CNN templates are designed for this nonlinearity \cite{{Chua1998},{Roska},{Itoh2003}}.} 

\end{enumerate}
Substituting Eq. (\ref{eqn: nl-resistor-2}) into Eq.(\ref{eqn: cnn10}), 
and using the relations $C=R=1$, $v_{r} = v_{ij}$, $a_{0, 0} = a$, 
we obtain 
\begin{center}
\begin{minipage}[]{12cm}
\begin{itembox}[l]{Dynamics of the modified CNN}
\begin{equation}
 \begin{array}{l}
  \displaystyle  \frac{dv_{ij}}{dt} = - v_{ij} + 0.5 \, a_{0, 0} \, ( |v_{ij} + 1 | -  |v_{ij} - 1 | ) \vspace{2mm} \\
      \scalebox{0.95}{$\displaystyle + \sum_{k, \, l \in N_{ij}, \  k \ne i, \ l \ne j} a_{kl} \, \operatorname {sgn} (v_{kl})  
     \displaystyle \ + \sum_{k, \, l \in N_{ij}}b_{k l} \ u_{kl} + z, $} \vspace{4mm} \\
   \hspace*{5mm} (i, \, j) \in \{ 1, \cdots , M \} \times \{ 1, \cdots , N \}.   
 \end{array}
\label{eqn: cnn11} 
\end{equation}
\end{itembox}
\end{minipage}
\end{center}

If we restrict the neighborhood radius of every cell to $1$, and if we assume that the feedback and control template parameters do not vary with space, then the template $\{ A, B, z \}$ is fully specified by $19$ parameters, which are the elements of two $3 \times 3$ matrices $A$ and $B$, namely 
\begin{equation}
 A =
 \begin{array}{|c|c|c|}
  \hline
   a_{-1, -1} & a_{-1, 0} & a_{-1,1} \\
  \hline
   a_{0, -1}  & a_{0, 0}  & a_{0,1}  \\  
  \hline 
   a_{1, -1}  & a_{1, 0}  & a_{1,1}  \\ 
  \hline
  \end{array} \ , \ \ \ 
 B = 
   \begin{array}{|c|c|c|}
  \hline
   b_{-1, -1} & b_{-1, 0} & b_{-1,1} \\
  \hline
   b_{0, -1} & b_{0, 0} & b_{0,1}    \\  
  \hline 
   b_{1, -1} & b_{1, 0} & b_{1,1}    \\ 
  \hline
  \end{array} \ , 
\label{eqn: AB-modified}
\end{equation}
and a real number $z$ \cite{{Chua1998},{Roska}}. 
As stated in Sec. \ref{sec: CNN}, the feedback and control template parameters can be described as follow:
\begin{equation}
  a_{k l} \stackrel{\triangle}{=} a_{k-i, \, l-j}, \ \  b_{k l} \stackrel{\triangle}{=} b_{k-i, \, l-j}. 
\label{eqn: a-kl-1}
\end{equation}
%
%

%
%
\subsection{Template parameter $a_{0,  0}$}
\label{sec: a00}
%
The following sum in Eq. (\ref{eqn: cnn11}):
\begin{equation}
  \sum_{k, \, l \in N_{ij}, \  k \ne i, \ l \ne j} a_{kl} \, \operatorname {sgn} (v_{kl}), 
\label{eqn: sum-a-kl}
\end{equation}
does \emph{not} contain the term: 
\begin{equation}
  a_{kl} \, \operatorname {sgn} (v_{kl}) \Bigr |_{k=i, \ l=j} =  a_{ij} \, \operatorname {sgn} (v_{ij}).   
\label{eqn: a-kl-2}
\end{equation}
Substituting Eq. (\ref{eqn: a-kl-1}) into the left-hand side of Eq. (\ref{eqn: a-kl-2}), we obtain 
\begin{equation}
 \begin{array}{l}
  a_{kl} \, \operatorname {sgn} (v_{kl}) \Bigr |_{k=i, \ l=j} 
    = a_{k-i, \, l-j} \, \operatorname {sgn} (v_{kl})  \Bigr |_{k=i, \ l=j} 
  \vspace{2mm} \\ 
   = a_{0, 0} \, \operatorname {sgn} (v_{ij}).   
 \end{array}
\end{equation}
Thus, $a_{0, 0} \, \operatorname {sgn} (v_{ij})$ is not included in Eq. (\ref{eqn: sum-a-kl}).  
That is, the element $a_{0, 0}$ of the feedback template $A$ in Eq. (\ref{eqn: AB-modified}) is not given.   
Therefore, \underline{we define  $a_{0, 0}$ by setting $a_{0, 0} = a$}, 
where $a$ is the parameter of the nonlinear resistor, which is given by Eq. (\ref{eqn: nl-resistor-2}). 
\begin{center}
\begin{minipage}[t]{12cm}
\begin{itembox}[l]{Element $a_{0, 0}$ of template $A$}
The element $a_{0, 0}$ of the feedback template $A$ is defined by 
\begin{equation}
 a_{0, 0} = a, 
\end{equation}
where $a$ is the parameter of the nonlinear resistor defined by 
\begin{equation}
  i_{r} = - 0.5 \, a \, ( |v_{r}+ 1 | -  |v_{r} - 1 | ).    
\label{eqn: nl-resistor-a00} 
\end{equation}
The other elements $a_{kl}$ of $A$ are defined by the coefficients of the output $y_{kl}$, that is,  
\begin{equation}
  a_{k l} \ y_{kl} = a_{kl} \, \operatorname {sgn} (v_{kl}).   
\label{eqn: a-kl-20}
\end{equation}
\end{itembox}
\end{minipage}
\end{center}

%
%
\subsection{Gray-scale edge detection}
\label{sec: edge}
%
Let us consider the \emph{gray-scale edge detection template} \cite{{Chua1998}, {Roska}}
\begin{equation}
 A =
 \begin{array}{|c|c|c|}
  \hline
   ~0~   &  ~0~   &  ~0~   \\
  \hline
   ~0~   &  ~2~  & ~0~   \\  
  \hline 
   ~0~   &  ~0~   & ~0~  \\ 
  \hline
  \end{array} \ , \ \ \ 
 B = 
   \begin{array}{|c|c|c|}
  \hline
   -1 &  -1  & -1 \\
  \hline
   -1 &   8  & -1   \\  
  \hline 
   -1 &  -1  & -1    \\ 
  \hline
  \end{array} \ ,  \ \ \ 
 z =
  \begin{array}{|c|}
  \hline
   - 0.5 \\
  \hline
  \end{array} \ .  
\label{eqn: template11} 
\end{equation}
The initial condition for $v_{ij}$ is  given by 
\begin{equation}
  v_{ij}(0)  = 0,   
\end{equation}
and the input $u_{kl}$ is equal to a given gray scale image.  
The boundary condition is given by 
\begin{equation}
  v_{k^{*}l^{*}}  = 0,  \  u_{k^{*}l^{*}}  = 0,
\end{equation}
where $k^{*}l^{*}$ denotes boundary cells.  
This template can extract edge of objects in gray scale image as shown in Figure \ref{fig:output-1}.  
Observe that the modified CNN (\ref{eqn: cnn11}) can hold a binary output image,  
even if the switch $S$ in Figure \ref{fig:CNN-model} is turned off at $t=80$.    
That is, the output binary state can not be changed by switching off.

%
\subsection{Shadow projection}
\label{sec: shadow}
%
Consider the \emph{shadow projection template} \cite{{Chua1998},{Roska}}

\begin{equation}
 A =
 \begin{array}{|c|c|c|}
  \hline
   ~0~   &  ~0~   &  ~0~   \\
  \hline
   ~0~   &  ~2~   & ~2~   \\  
  \hline 
   ~0~   &  ~0~   & ~0~  \\ 
  \hline
  \end{array} \ , \ \ \ 
 B = 
   \begin{array}{|c|c|c|}
  \hline
   ~0~   &  ~0~   &  ~0~   \\
  \hline
   ~0~   &  ~2~     &  ~0~   \\  
  \hline 
   ~0~   &  ~0~   &  ~0~   \\
  \hline
  \end{array} \ ,  \ \ \ 
 z =
  \begin{array}{|c|}
  \hline
   ~0~ \\
  \hline
  \end{array} \ . 
\label{eqn: template12} 
\end{equation}
The initial condition for $v_{ij}$ is  given by 
\begin{equation}
  v_{ij}(0)  = 1,   
\end{equation}
and the input $u_{kl}$ is equal to a given binary image.  
The boundary condition is given by 
\begin{equation}
  v_{k^{*}l^{*}}  = 0,  \  u_{k^{*}l^{*}}  = 0,
\end{equation}
where $k^{*}l^{*}$ denotes boundary cells.  
This template can project onto the left shadow of all objects illuminated from the right as shown in Figure \ref{fig:output-2}.  
Observe that the modified CNN (\ref{eqn: cnn11}) can hold a binary output image,  
even if the switch $S$ is turned off at $t=40$.    
That is, the output binary state can not be changed by switching off $S$ as stated above.    

\begin{figure}[p]
 \centering
  \begin{tabular}{ccc}
  \psfig{file=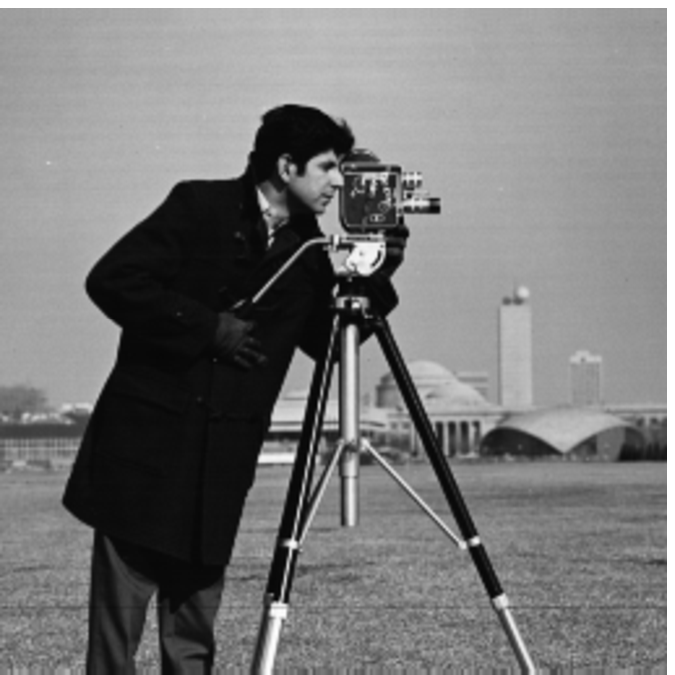,width=5cm} &
  \psfig{file=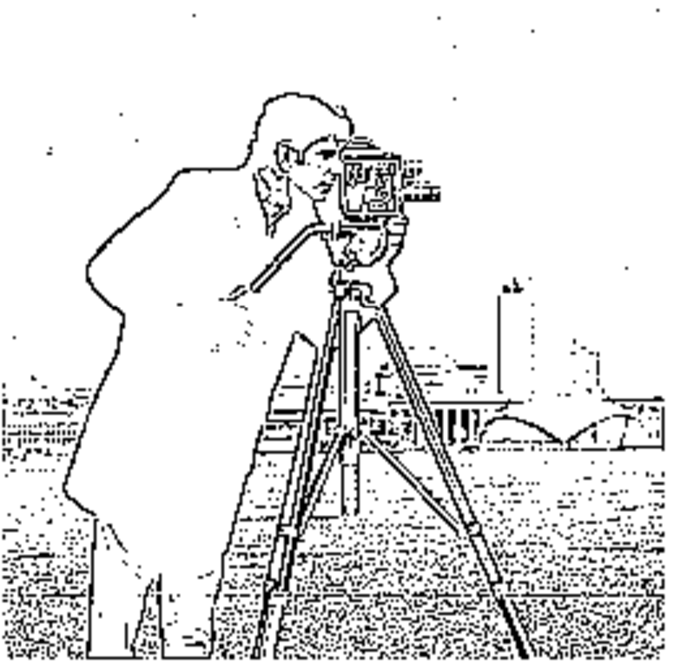,width=5cm} &
  \psfig{file=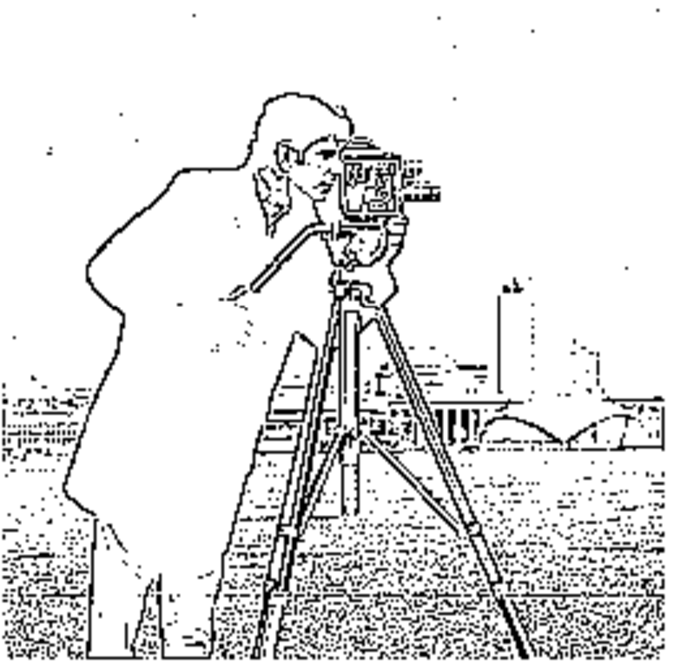,width=5cm} \vspace{1mm} \\
  (a) input image   & (b) output image ($t=70$) & (c) output image ($t=150$)  \\ 
   & \underline{\emph{before}} switch $S$ off & \underline{\emph{after}} switch $S$ off 
  \end{tabular}
  \caption{Image-holding property of the modified CNN (\ref{eqn: cnn11}) with the gray-scale edge detection template (\ref{eqn: template11}).   
  This template can extract edge of objects in gray scale image
  The modified CNN (\ref{eqn: cnn11}) can hold a binary output image even if all signals are disconnected from the cells 
  (The switch $S$ in Figure \ref{fig:CNN-model} is turned off at $t=80$).  
  Figures \ref{fig:output-1}(b) and (c) show the output images ``before'' and ``after'' the switch is turned off, respectively.  
  Observe that they are same.   Image size is $256 \times 256$. }
 \label{fig:output-1}
\end{figure}
%
%

\begin{figure}[hpbt]
 \centering
  \begin{tabular}{ccc}
  \psfig{file=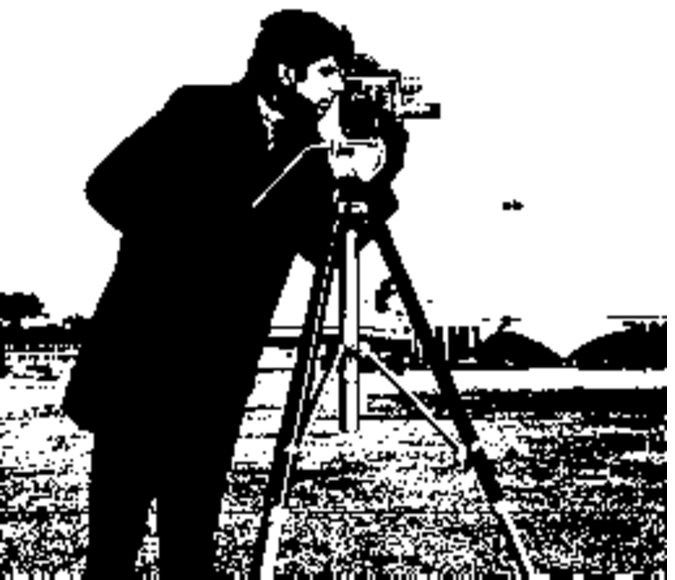,width=5cm} &
  \psfig{file=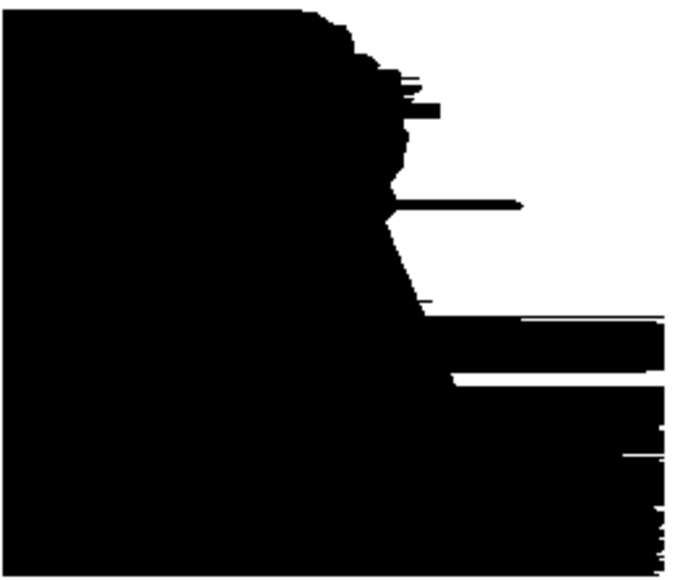,width=5cm} &
  \psfig{file=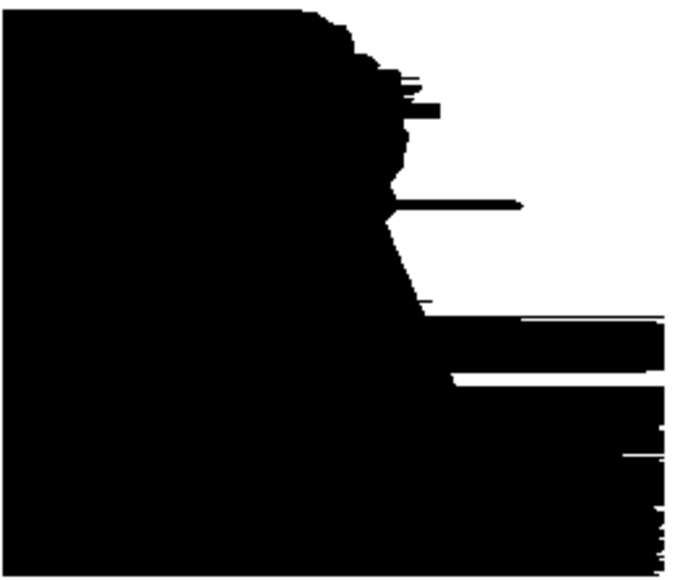,width=5cm} \vspace{1mm} \\
  (a) input image   & (b) output image ($t=35$) & (c) output image ($t=50$) \\ 
   & \underline{\emph{before}} switch $S$ off & \underline{\emph{after}} switch $S$ off
  \end{tabular}
  \caption{Image-holding property of the modified CNN (\ref{eqn: cnn11}) with the shadow projection template (\ref{eqn: template12}).   
  This template can project onto the left shadow of all objects illuminated from the right 
  The modified CNN (\ref{eqn: cnn11}) can hold a binary output image even if all signals are disconnected from the cells 
  (The switch $S$ in Figure \ref{fig:CNN-model} is turned off at $t=40$).  
  Figures \ref{fig:output-2}(b) and (c) show the output images ``before'' and ``after'' the switch is turned off, respectively.  
  Observe that they are same.  Image size is $256 \times 256$. }
 \label{fig:output-2}
\end{figure}
\newpage

%
%
\subsection{Behavior of the isolated cell after switch-off}
\label{sec: isolated-cell-2}
%

Assume that the switch $S$ in Figure \ref{fig:CNN-model} is turned off at $t=t_{0}$.   
Then, the dynamics of the cell is given by 
\begin{equation}
 \begin{array}{l}
  \displaystyle  \frac{dv_{ij}}{dt} = - v_{ij} + 0.5 \, a_{0, 0} \, ( |v_{ij} + 1 | -  |v_{ij} - 1 | ), 
   \vspace{4mm} \\
   \hspace*{5mm} (i, \, j) \in \{ 1, \cdots , M \} \times \{ 1, \cdots , N \}.   
 \end{array}
\label{eqn: cnn21} 
\end{equation}
Let us study the behavior of Eq. (\ref{eqn: cnn21}) for the three cases: 
$a_{0, 0} > 1$, $a_{0,0} =1 $, and $a_{0,0} \le 1 $.
\begin{enumerate}
\item {\bf Case 1.} \ $a_{0, 0} > 1$. \\
The driving-point plot of Eq. (\ref{eqn: cnn21}) for $a_{0, 0} = 2$ is shown in Figure \ref{fig:stability}.  
From Eq. (\ref{eqn: equilibrium-1}), we obtain  
\begin{equation} 
 \left.
    \begin{array}{cl}
      \text{if~~} v_{ij}(t_{0})<0,  & \text{then} \displaystyle \lim_{t \to \infty} v_{ij}(t) \to -2, \vspace{2mm} \\
      \text{if~~} v_{ij}(t_{0})>0,  & \text{then} \displaystyle \lim_{t \to \infty} v_{ij}(t) \to 2, \vspace{2mm} \\
      \text{if~~} v_{ij}(t_{0})=0,  & \text{then~~} v_{ij}(t) = 0,
    \end{array}
 \right \}
\end{equation}
where $t>t_{0}$.  
Since the solution $v_{ij}(t)$ can not move across the origin,  
the output $y_{ij}(t)$  satisfies the following relations:  
\begin{equation} 
 \left.
  \begin{array}{ll}
   (1) & \text{ if~} y_{ij}(t_{0}) = \operatorname {sgn} (v_{ij}(t_{0})) = -1,  \vspace{2mm} \\
       & \text{~then~} y_{ij}(t) = \operatorname {sgn} (v_{ij}(t))= -1. \vspace{4mm} \\
   (2) & \text{if~} y_{ij}(t_{0}) = \operatorname {sgn} (v_{ij}(t_{0})) = 1,   \vspace{2mm} \\
       &  \text{~then~}  y_{ij}(t) = \operatorname {sgn} (v_{ij}(t))= 1. \vspace{4mm} \\
   (3) & \text{ if~} y_{ij}(t_{0}) = \operatorname {sgn} (v_{ij}(t_{0})) = 0,   \vspace{2mm} \\
       & \text{~then~} y_{ij}(t) = \operatorname {sgn} (v_{ij}(t))= 0,
  \end{array}
 \right \}
\label{eqn: y-ij-t}
\end{equation}
where $t>t_{0}$. 
Thus, the output binary state can not be changed even if we turn off the switch $S$ at $t=t_{0}$.  \\

\item { \bf Case 2.}  \ $a_{0,0} < 1 $. \\
We show the driving-point plot of Eq. (\ref{eqn: cnn21}) for $a_{0,0} = -1$ in Figure \ref{fig:stability-a-0}.  
Observe that the origin is a stable equilibrium point.  
Thus, any trajectory tends to the origin, that is,  $v_{ij}(t) \to 0$ as $t \to \infty$.  
Since the solution $v_{ij}(t)$ for $t>t_{0}$ can not move across the origin, Eq. (\ref{eqn: y-ij-t}) also holds.  
That is, the output $y_{ij}(t) = \operatorname {sgn} (v_{ij}(t))$ does not change (except for $t=\infty$), even if we turn off the switch $S$ at $t=t_{0}$.  \\

\item {\bf Case 3.} \ $a_{0,0}=1$. \\
We show the driving-point plot of Eq. (\ref{eqn: cnn21}) for $a_{0,0}=1$ in Figure \ref{fig:stability-a-1}.  
Observe that Eq. (\ref{eqn: cnn21}) has an invariant set $D$.  
Any trajectories outside of $D$ tends to the boundary of this set.  
Furthermore, the solution $v_{ij}(t)$ for $t>t_{0}$ can not move across $D$. 
Thus, Eq. (\ref{eqn: y-ij-t}) also holds,   
that is, the output binary state can not be changed even if we turn off the switch $S$ at $t=t_{0}$.

\end{enumerate}
We conclude that the modified CNN (\ref{eqn: cnn11}) can hold a binary output image  
even if all cells are disconnected from the summing amplifier and no signal is supplied to the cell.   
However, we should assume that $a_{0,0}>1$.   
It is due to the reason that if $a_{0,0}< 1$, then the physical circuit, for example, the sign function circuit, may not work properly in the presence of noise when $|v_{ij}(t)|$ becomes sufficiently small.    
Furthermore, if $a_{0,0}=0$, then the qualitative behavior of Eq. (\ref{eqn: cnn21}) is greatly changed by the small perturbation of $a_{0,0}$. 

Note that the modified CNN requires power to maintain the output image, since the isolated cell has a nonlinear active resistor. 
That is, the output image is lost immediately (\emph{volatile}) when the power is interrupted. 
From our computer simulations\footnote{If the input image size is $256 \times 256$, 
we have to solve a system of 65536 first-order differential equations.  
Thus, we used the simple Euler method for solving Eq. (\ref{eqn: cnn11}).  It is the most basic method for numerical integration.}, we obtain the following result: 
%
%
\begin{center}
\begin{minipage}{14cm}
\begin{shadebox}
Assume $a_{00} > 1$.  
Then the modified CNN (\ref{eqn: cnn11}) can hold a binary output image  
even if all cells are disconnected from the summing amplifier and no signal is supplied to the cell after a certain point of time.  
The modified CNN requires power to maintain the output image, that is, it is \emph{volatile}.    
\end{shadebox}
\end{minipage}
\end{center}
%
%

\begin{figure}[h]
 \begin{center} 
  \psfig{file=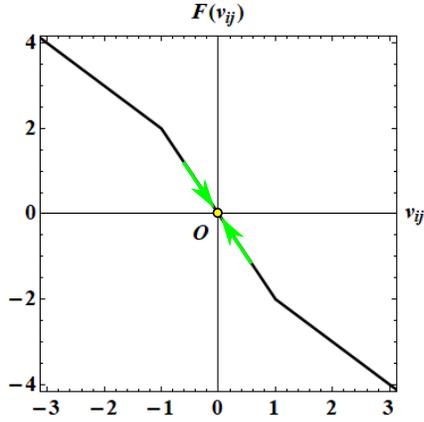,width=6.0cm}
  \caption{Driving-point plot of Eq. (\ref{eqn: cnn21}) with $a_{0,0} = -1$.  
   Any trajectory tends to the origin, that is, the origin is a stable equilibrium point.}
 \label{fig:stability-a-0}
 \end{center}
\end{figure}
%
%

\begin{figure}[h]
 \begin{center} 
  \psfig{file=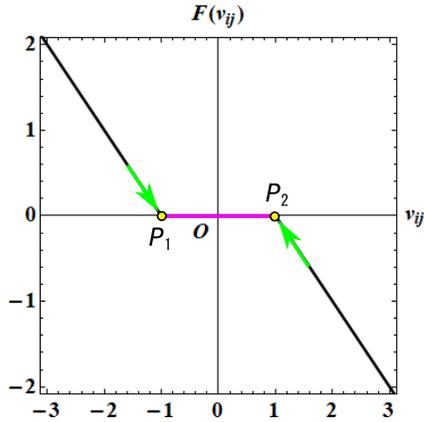,width=6.0cm}
  \caption{Driving-point plot of Eq. (\ref{eqn: cnn21}) with $a_{0,0}=1$.  
   The region $D \stackrel{\triangle}{=} [ P_{1} \le v_{ij} \le P_{2} ]$ is an invariant set. 
   That is, any trajectory starting inside $D$ cannot escape from it.  
   Any trajectories outside of $D$ tends to the boundary of this set, that is, $P_{1}$ or $P_{2}$.}
 \label{fig:stability-a-1}
 \end{center}
\end{figure}
\newpage
%
%
%
\section{Memoristor CNN}
\label{sec: memoristor-CNN}
%
%
%
The memristor can switch ``off'' and ``on'', depending on the value of the flux.
In this section, we realize the switch $S$ in Figure \ref{fig:CNN-model} by using memristors.

\begin{figure}[t]
 \begin{center} 
  \psfig{file=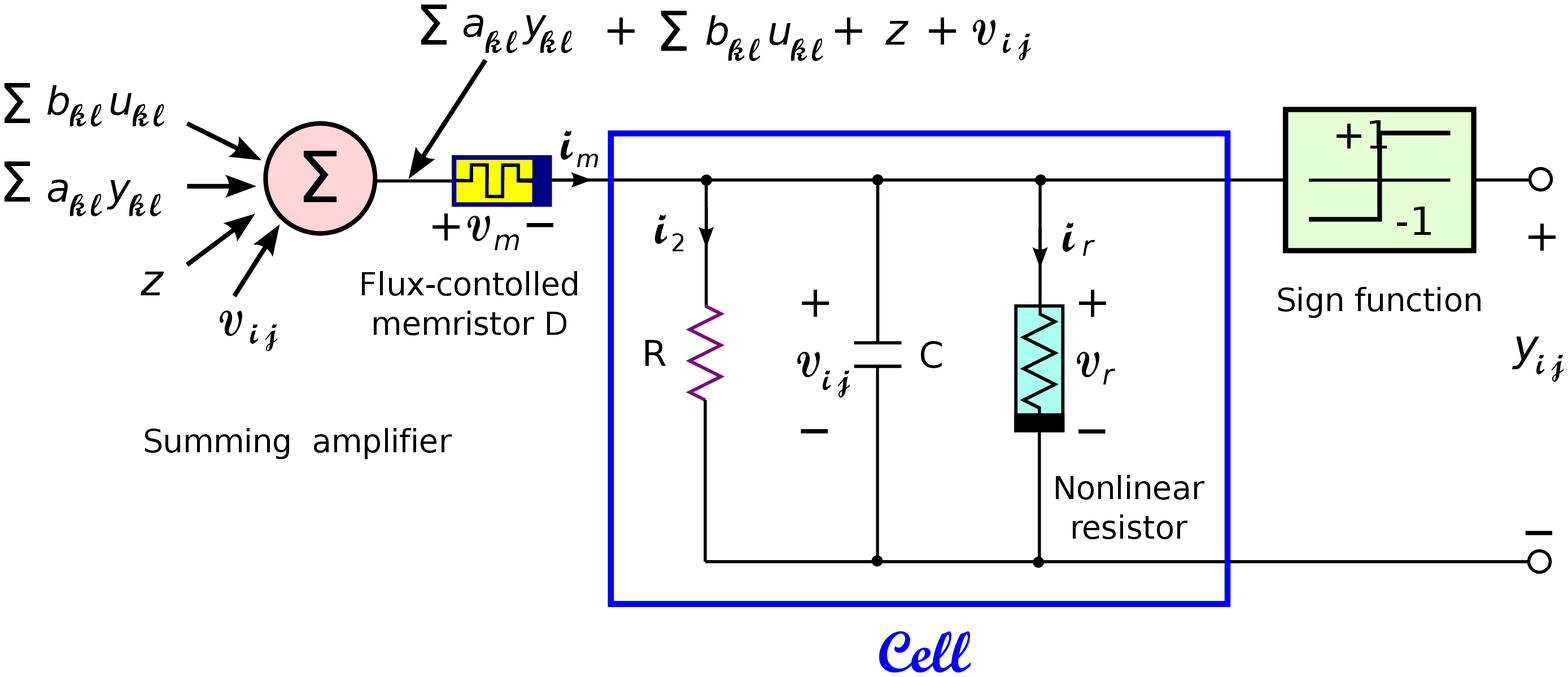,width=15cm}
  \caption{Memristor CNN circuit.  
   \newline
   (1) Parameters: $R=1, \ C=1$.  
   \newline 
   (2) The $v-i$ characteristic of the nonlinear resistor (light blue) is given by 
   $i_{r} =  f_{r}(v_{r}) \stackrel{\triangle}{=} - 0.5 \, a \, ( |v_{r}+ 1 | -  |v_{r} - 1 | )$, 
   where $a$ is a constant.  
   \newline
   (3) The terminal currents and voltages of the flux-controlled memristor $D$ (yellow) satisfies 
   $i_{m} = W( \varphi_{m} ) \, v_{m}$, where $W( \varphi_{m} ) = 1 $ for $-1< \varphi_{m} < 1$, 
       and $W( \varphi_{m} ) = 0$ for $\varphi_{m} \le -1$ and $\varphi_{m} \ge 1$. 
   \newline
   (4) The output voltage of the summing amplifier (pink) is given by  
   $\displaystyle \sum_{k, \, l \in N_{ij}, \  k \ne i, \ l \ne j} a_{kl} \, \operatorname {sgn} (v_{kl}) 
   + \sum_{k, \, l \in N_{ij}}b_{k l} \ u_{kl} + z + v_{ij}$.  
   \newline
   (5) The above output voltage contains the voltage $v_{ij}$ of the capacitor $C$ (the last term).   
   \newline
   (6) The above sum 
   $\displaystyle \sum_{k, \, l \in N_{ij}, \  k \ne i, \ l \ne j} a_{kl} \, \operatorname {sgn} (v_{kl}) $ 
   does \emph{not} contain the term 
   $ a_{ij} \, \operatorname {sgn} (v_{ij})$ \ ($k=i, \ l=j$).   
   \newline  
   (7) The output $y_{ij}$ and the state $v_{ij}$ of each cell are related via the sign function (green): 
    $y_{ij} = \operatorname {sgn} (v_{ij})$.  }  
 \label{fig:cnn-memistor}
 \end{center}
\end{figure}

Consider the circuit shown in Figure \ref{fig:cnn-memistor}.  
The dynamics of this circuit is given by 
\begin{equation}
 \begin{array}{l}
  \displaystyle C \frac{dv_{ij}}{dt} 
   = \displaystyle - \frac{v_{ij}}{R} -  i_{r}  +  i_{m} , \vspace{2mm} \\
    (i, \, j) \in \{ 1, \cdots , M \} \times \{ 1, \cdots , N \},
 \end{array}
\label{eqn: memristor-cnn1} 
\end{equation}
where $C=R=1$, and the symbols $v_{ij}$, $i_{r}$, and $i_{m}$ denote the voltage across the capacitor $C$, the current through the nonlinear resistor, and the current through the memristor, respectively.  
Furthermore, this circuit satisfies the following relations:  
%
%
\begin{enumerate}
\renewcommand{\labelenumi}{(\alph{enumi})}
%
\item The output $y_{ij}$ and the state $v_{ij}$ of each cell is related via the sign function 
\begin{equation}
  y_{ij} = \operatorname {sgn} (v_{ij}).  
\end{equation}
%
%
\item The characteristic of the nonlinear resistor is given by 
\begin{equation}
  i_{r}= f_{r}(v_{r}) = - 0.5 \, a \, ( |v_{r}+ 1 | -  |v_{r} - 1 | ),  
\label{eqn: nl-resistor-3} 
\end{equation}
where $a$ is a constant.  \\
\item The terminal voltage $v_{m}$ and the terminal current $i_{m}$ of the memristor is given by 
\begin{equation}
  i_{m} = W( \varphi_{m} )v_{m},
\label{eqn: mem-w} 
\end{equation}
where the flux $\varphi_{m}$ of the memristor is defined by 
\begin{equation}
   \varphi_{m}  =  \int_{- \infty}^{t} \, v_{m} \, dt.   
\label{eqn: phi-m-1}
\end{equation}
%
%
\item The voltage $v_{m}$ across the memristor is given by  
\begin{equation}
   v_{m} = 
   \scalebox{0.9}{$\displaystyle  \left ( \displaystyle \sum_{k, \, l \in N_{ij}, \ k \ne i, \ l \ne j}
      a_{k l} \ y_{kl} + \sum_{k, \, l \in N_{ij}} b_{k l} \ u_{kl} + z \right ).  $}
\label{eqn: vm-parasitic}
\end{equation}
Note that the first term does not contain $ a_{ij} \, \operatorname {sgn} (v_{ij})$, where $k=i, \ l=j$. \\ 
%
%
\item Each cell has \emph{only one memristor}. 
\end{enumerate} 
%
%

Substituting Eqs. (\ref{eqn: nl-resistor-3}), (\ref{eqn: mem-w}) into Eq. (\ref{eqn: memristor-cnn1}), and using the relations $C=R=1$, $v_{r} = v_{ij}$,  we obtain  
\begin{center}
\begin{minipage}[]{12cm}
\begin{itembox}[l]{Dynamics of the memristor CNN}
\begin{equation}
 \begin{array}{lll}
  \displaystyle \frac{dv_{ij}}{dt} 
   &=& \displaystyle - v_{ij} + 0.5 \, a \, \bigl ( \, |v_{ij}+1|-|v_{ij}-1| \, \bigr ) 
     \displaystyle + \ W (\varphi_{m})v_{m},  \vspace{2mm} \\
   && (i, \, j) \in \{ 1, \cdots , M \} \times \{ 1, \cdots , N \}.
 \end{array}
\label{eqn: memristor-cnn1-recast} 
\end{equation}
\end{itembox}
\end{minipage}
\end{center}
As stated in Sec. \ref{sec: CNN}, if we restrict the neighborhood radius of every cell to $1$, and if we assume that the feedback and control template parameters do not vary with space, then the templates are fully specified by $19$ parameters, which are the elements of two $3 \times 3$ matrices $A$ and $B$, namely 
\begin{equation}
 \scalebox{0.95}{$\displaystyle
 A =
 \begin{array}{|c|c|c|}
  \hline
   a_{-1, -1} & a_{-1, 0} & a_{-1,1} \\
  \hline
   a_{0, -1}  & a_{0, 0}  & a_{0,1}  \\  
  \hline 
   a_{1, -1}  & a_{1, 0}  & a_{1,1}  \\ 
  \hline
  \end{array} \ , \ 
 B = 
   \begin{array}{|c|c|c|}
  \hline
   b_{-1, -1} & b_{-1, 0} & b_{-1,1} \\
  \hline
   b_{0, -1} & b_{0, 0} & b_{0,1}    \\  
  \hline 
   b_{1, -1} & b_{1, 0} & b_{1,1}    \\ 
  \hline
  \end{array} \ ,  $}
\end{equation}
and a real number $z$, where $a_{0, 0} = a$ ($a$ is the parameter of the nonlinear resistor defined by Eq. (\ref{eqn: nl-resistor-3})).\footnote{  
The term $a_{ij} \, \operatorname {sgn} (v_{ij})= a_{0, 0} \, \operatorname {sgn} (v_{ij})$ 
is not included in Eq. (\ref{eqn: vm-parasitic}), as shown in Sec. \ref{sec: modified-CNN}.  
Therefore, we have to define the element $a_{0, 0}$ of the template $A$ by setting $a_{0, 0} = a$. }

%
\subsection{Dilation}
\label{sec: dilation}
%
Let us consider the \emph{dilation template} \cite{{Chua1998},{Roska}}
\begin{equation}
\begin{array}{l}
 A =
 \begin{array}{|c|c|c|}
  \hline
   ~0~   &  ~0~   &  ~0~   \\
  \hline
   ~0~   &  ~2~  & ~0~   \\  
  \hline 
   ~0~   &  ~0~   & ~0~  \\ 
  \hline
  \end{array} \ , \ \ \ 
 B = 
   \begin{array}{|c|c|c|}
  \hline
   ~0~ &  ~1~  &  ~0~ \\
  \hline
   ~1~ &  ~1~  &  ~1~   \\  
  \hline 
   ~0~ &  ~1~  &  ~0~    \\ 
  \hline
  \end{array} \ ,  \ \ \ 
 z =
  \begin{array}{|c|}
  \hline
    4.5 \\
  \hline
  \end{array} \ . 
\end{array}
\label{eqn: template31} 
\end{equation}
The initial condition for $v_{ij}$ is  given by 
\begin{equation}
  v_{ij}(0)  = 0,   
\end{equation}
and the input $u_{kl}$ is equal to a given binary image.  
The boundary condition is given by 
\begin{equation}
  v_{k^{*}l^{*}}  = -1,  \  u_{k^{*}l^{*}}  = -1,
\end{equation}
where $k^{*}l^{*}$ denotes boundary cells. 
This template can be used to grow a layer of pixels around objects.

Assume that the memductance $W (\varphi_{m})$ is given  by 
\begin{equation}
  W (\varphi_{m}) 
  \stackrel{\triangle}{=} 
    \begin{cases}
      1  & -1 \le \varphi_{m} \le 1, \\
      0  &  \varphi_{m} < -1, \ \varphi_{m} > 1.     
    \end{cases}
\label{eqn: w-phi-10}
\end{equation}
The constitutive relation and memductance of the memristor is shown in Figure \ref{fig:relation-10}.
The terminal voltage $v_{m}$ satisfies 
\begin{equation}
\begin{array}{lll}
  |v_{m}| &=& \scalebox{0.95}{$\displaystyle    \left | \displaystyle \sum_{k, \, l \in N_{ij}, \ k \ne i, \ l \ne j}
      a_{k l} \ y_{kl} + \sum_{k, \, l \in N_{ij}} b_{k l} \ u_{kl} + z \right |  $} \vspace{2mm} \\
          &=&  \displaystyle   \left |  \sum_{k, \, l \in N_{ij}} b_{k l} \ u_{kl} + 4.5 \right |   \vspace{2mm} \\ 
          &=&  \displaystyle  \Bigl | u_{i-1, \, j}  +  u_{i, \, j-1} + u_{i, \, j} + u_{i, \, j+1}        \vspace{2mm} \\ 
           &&   ~~~+ u_{i+1, \, j} + 4.5  \Bigr |  \ge  0.5,
\label{eqn: abs-vm-1}
\end{array}
\end{equation}
where $u_{kl} = \pm 1$.  
For example, assume that 
\begin{equation}
  u_{i-1, \, j} =  u_{i, \, j-1} = u_{i, \, j} = u_{i, \, j+1} = u_{i+1, \, j}  = - 1.  
\end{equation}
Then, we obtain
\begin{equation}
\begin{array}{lll}
  v_{m} &=& u_{i-1, \, j}  +  u_{i, \, j-1} + u_{i, \, j} + u_{i, \, j+1} + u_{i+1, \, j} + 4.5  \vspace{2mm} \\
        &=& -5+ 4.5  = - 0.5.
\end{array}
\end{equation}
From Eq. (\ref{eqn: phi-m-1}), we obtain 
\begin{equation}
  \varphi_{m}(t) = - 0.5t, 
\end{equation}
where we assumed $\varphi_{m}(0)=0$.  
Thus, 
\begin{equation}
  W (\varphi_{m}(t)) 
  =
    \begin{cases}
      1  & 0 \le t \le 2, \\
      0  & t > 2.      
    \end{cases}
\end{equation}
Hence, the memristor switches ``off'' for $t > 2$. 
Our computer simulations are given in Figure \ref{fig:output-3}.  
Observe that the template (\ref{eqn: template31}) can be used to grow a layer of pixels around objects.
It can hold a binary output image for $t >2$, at which all memristors switched ``off''.    
Note that all memristors do \emph{not} switch ``off'' \emph{synchronously} since their terminal flux are not always identical.

\begin{figure}[hpbt]
 \centering
  \begin{tabular}{cc}
   \psfig{file=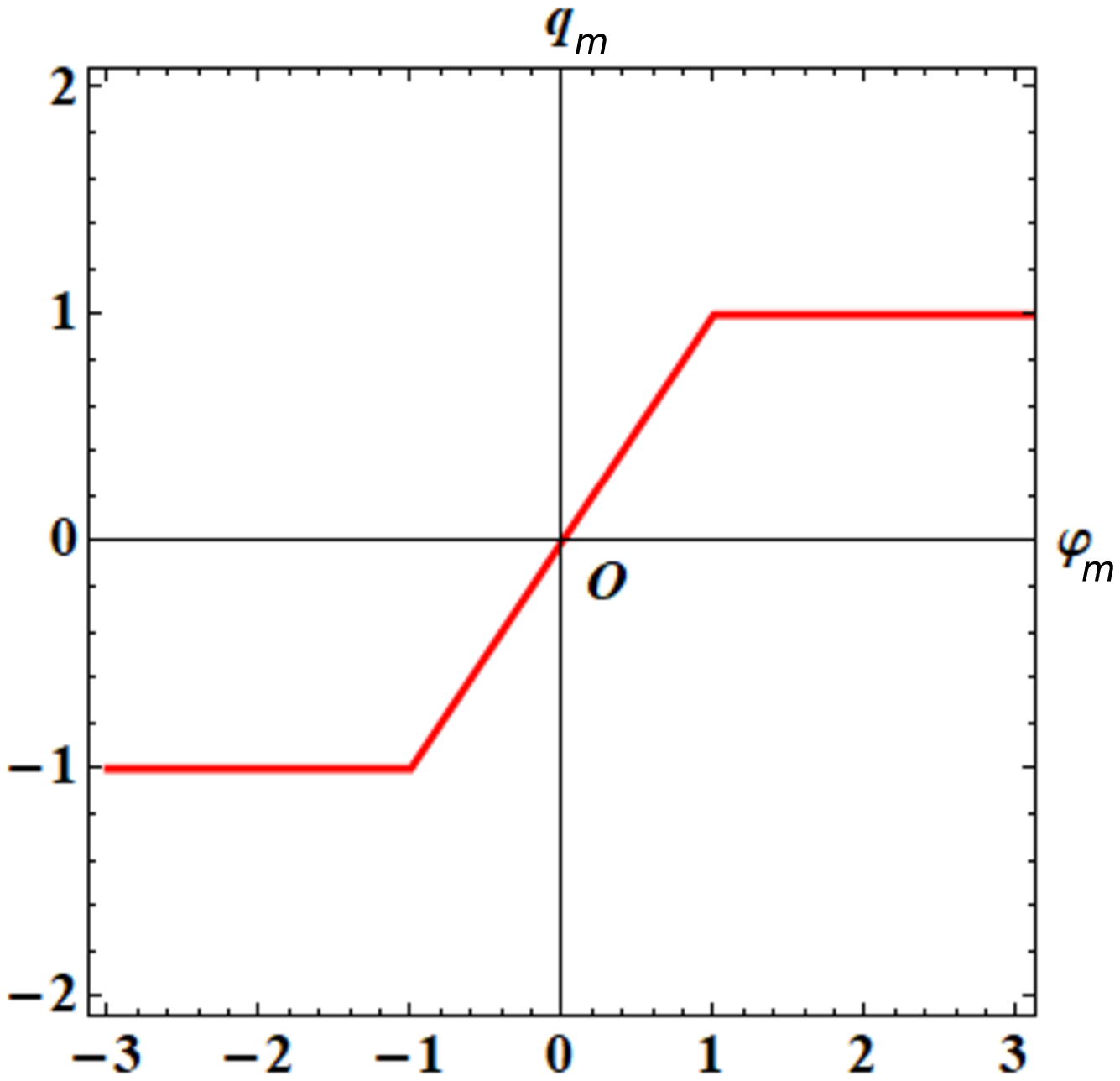, width=6.0cm} & 
   \psfig{file=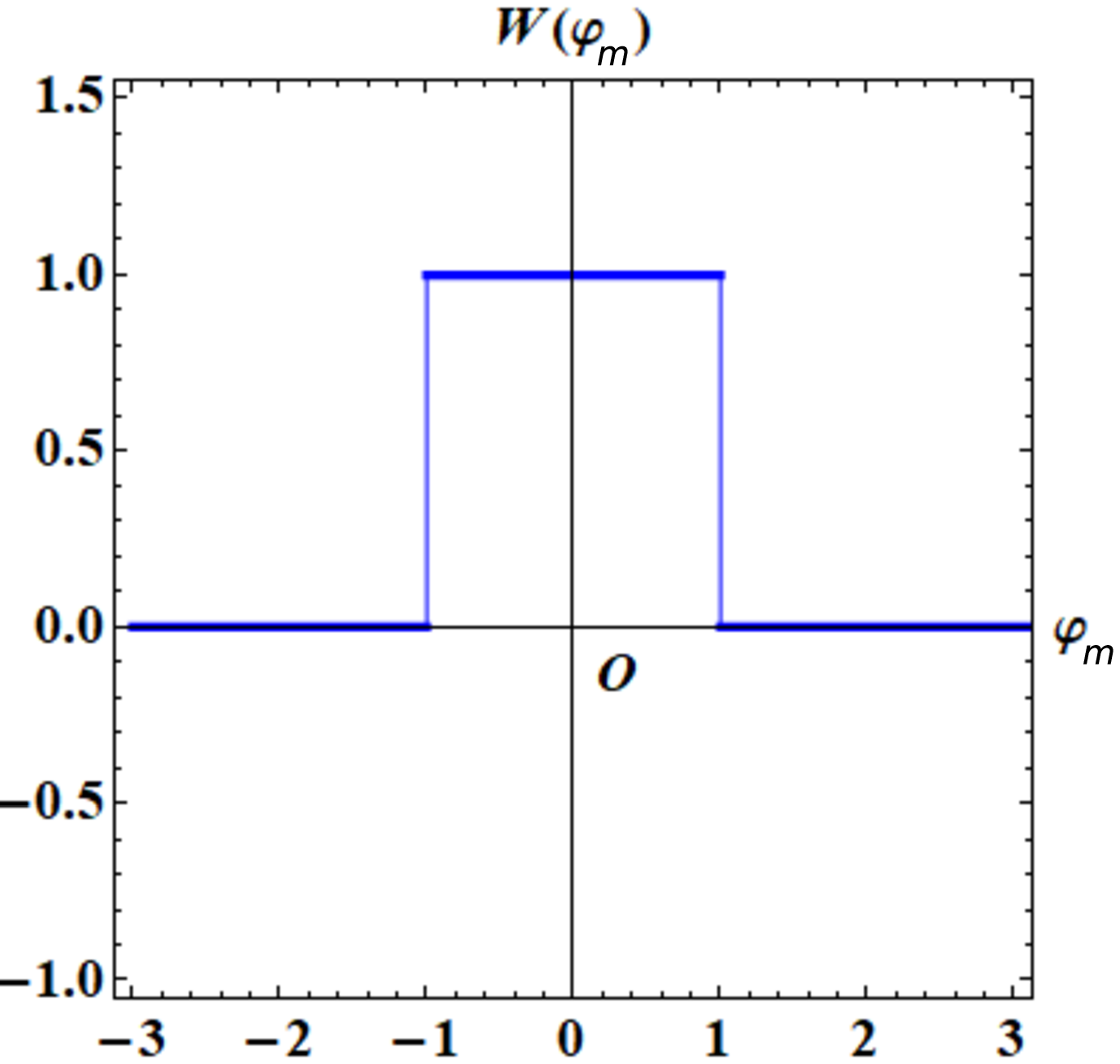, width=6.0cm} \vspace{1mm} \\
   (a) constitutive relation $q_{m} = h (\varphi_{m})$ & 
   (b) memductance  $W( \varphi_{m} )$ \\
  \end{tabular} 
  \caption{Constitutive relation and memductance of the flux-controlled memristor.  
   The memristor switches ``off'' and ``on'' depending on the value of the flux $\varphi_{m}$. 
   \newline 
   (a) The constitutive relation of the memristor, which  is given by 
       $q_{m} = h(\varphi_{m}) \stackrel{\triangle}{=} 0.5 ( |\varphi_{m} + 1| - |\varphi_{m} - 1| )$.      
   \newline
   (b) Memductance $W( \varphi )$ of the  memristor, which is defined by   
       $\displaystyle W( \varphi_{m} ) \stackrel{\triangle}{=} \frac{dh(\varphi_{m})}{d\varphi_{m}}$.      
       Thus, $W( \varphi_{m} ) = 1 $ for $-1< \varphi_{m} < 1$,   
       and $W( \varphi_{m} ) = 0$ for $\varphi_{m} \le -1$ and $\varphi_{m} \ge 1$. }
 \label{fig:relation-10}
\end{figure}
%
%

\begin{figure}[hpbt]
 \centering
  \begin{tabular}{ccc}
  \psfig{file=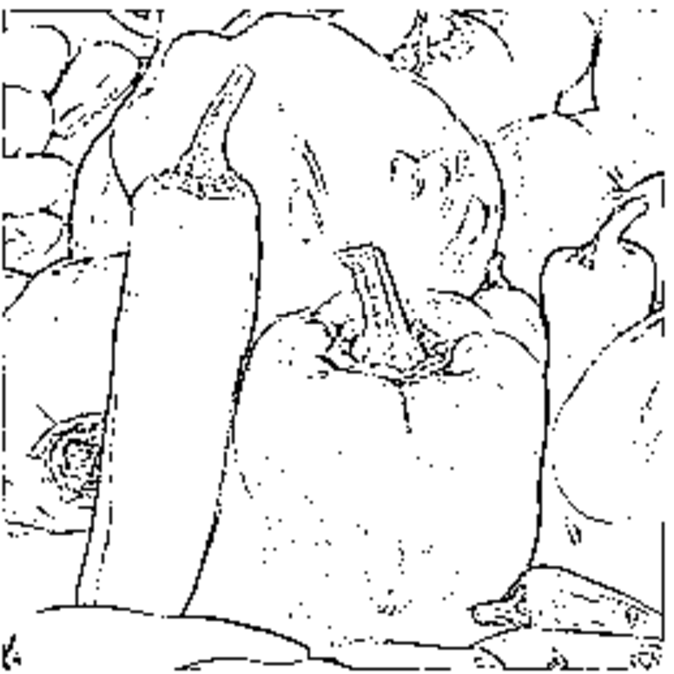,width=5cm} &
  \psfig{file=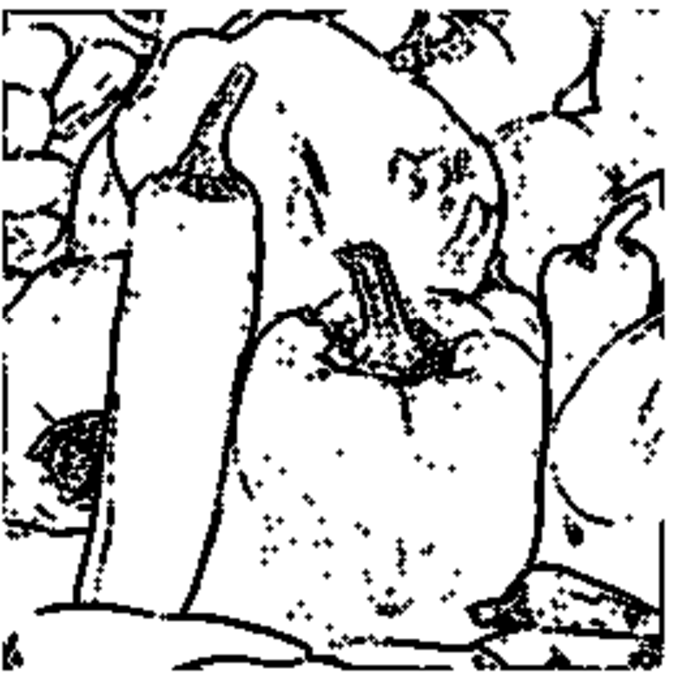,width=5cm} &
  \psfig{file=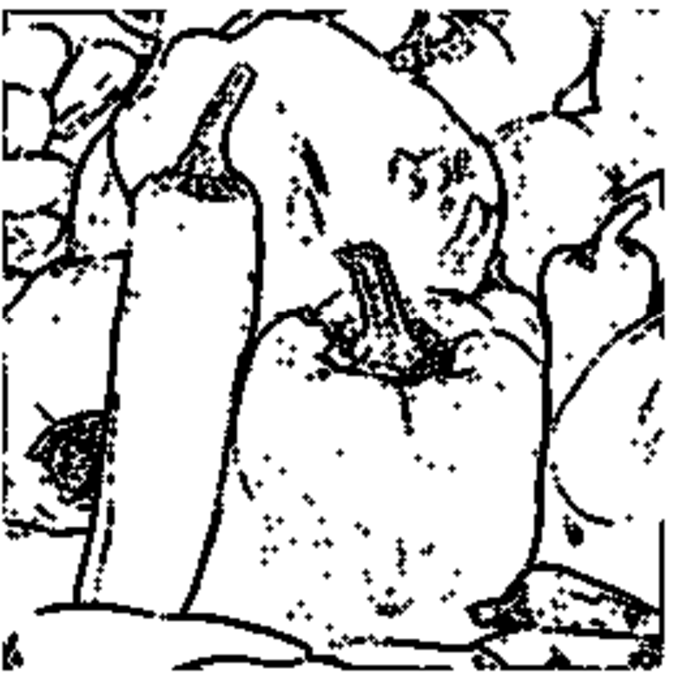,width=5cm} \vspace{1mm} \\
  (a) input image   & (b) output image \emph{{\bf before}} all & (c) output image \emph{{\bf after}} all  \\ 
   & memristors turned off  ($t=1$) &  memristors turned off ($t=5$) 
  \end{tabular}
  \caption{Image-holding property of the memoristor CNN (\ref{eqn: memristor-cnn1}) 
  with the dilation template (\ref{eqn: template31}).  
  This template can be used to grow a layer of pixels around objects.
  The memoristor CNN (\ref{eqn: memristor-cnn1}) can hold a binary output image even if all memristos are switched ``off''.  
  The memristor in Figure \ref{fig:cnn-memistor} is turned off at $t>2$.  
  Figures \ref{fig:output-3}(b) and (c) show the output images ``before'' and ``after'' all memristors are turned off, respectively.  
  Observe that they are same.   
  The memristor CNN cell size is $256 \times 256$, since the input image size is $256 \times 256$.}
 \label{fig:output-3}
\end{figure}
\newpage

%
\subsection{Sharpening with binary output}
\label{sec: sharpening}
%
Let us consider the \emph{sharpening with binary output template} 
\begin{equation}
\begin{array}{l}
 A =
 \begin{array}{|c|c|c|}
  \hline
   ~0~   &  ~0~   &  ~0~   \\
  \hline
   ~0~   &  ~2~  & ~0~   \\  
  \hline 
   ~0~   &  ~0~   & ~0~  \\ 
  \hline
  \end{array} \ , \ \ \ 
 B = 
   \begin{array}{|c|c|c|}
  \hline
   ~~~0 &  ~-1~  &  ~~~0 \\
  \hline
   ~-1~  & ~~~5  &  ~-1~   \\  
  \hline 
   ~~~0 &  ~-1~  &  ~~~0    \\ 
  \hline
  \end{array} \ ,  \vspace{5mm} \\
 z =
  \begin{array}{|c|}
  \hline
    0.5 \\
  \hline
  \end{array} \ . 
\end{array}
\label{eqn: template33} 
\end{equation}
The initial condition for $v_{ij}$ is  given by 
\begin{equation}
  v_{ij}(0)  = 0,   
\end{equation}
and the input $u_{kl}$ is equal to a given \emph{gray-scale} image.  
The boundary condition is given by 
\begin{equation}
  v_{k^{*}l^{*}}  = 0,  \  u_{k^{*}l^{*}}  = 0,
\end{equation}
where $k^{*}l^{*}$ denotes boundary cells. 
This template can be used to make an image appear sharper by enhancing edges and convert into a binary image.  

Assume that the memductance $W (\varphi_{m})$ is given by  Eq. (\ref{eqn: w-phi-10}).  
The constitutive relation and memductance of the memristor is shown in Figure \ref{fig:relation-10}.
The terminal voltage $v_{m}$ satisfies 
\begin{equation}
\begin{array}{l}
  |v_{m}| = \scalebox{0.95}{$\displaystyle    \left | \displaystyle \sum_{k, \, l \in N_{ij}, \ k \ne i, \ l \ne j}
      a_{k l} \ y_{kl} + \sum_{k, \, l \in N_{ij}} b_{k l} \ u_{kl} + z \right |  $} \vspace{2mm} \\
          =  \displaystyle   \left |  \sum_{k, \, l \in N_{ij}} b_{k l} \ u_{kl} - 0.5 \right |   \vspace{2mm} \\ 
          =  \displaystyle  \Bigl | 
                       -  u_{i-1, \, j-1}  - u_{i-1, \, j}  - u_{i-1, \, j+1}     \vspace{2mm} \\     
                    ~~~~ -  u_{i, \, j-1}    + 9 u_{i, \, j}  - u_{i, \, j+1}       \vspace{2mm} \\ 
                    ~~~~ -  u_{i+1, \, j-1}  - u_{i+1, \, j}  - u_{i+1, \, j+1} - 0.5  \Bigr |  \ge  0,
\label{eqn: abs-vm-10}
\end{array}
\end{equation}
where $ -1 \le u_{kl} \le 1$ (gray-scale image).  
Thus, if $ v_{m}(t)$ becomes zero at $t=t_{0}$, then $ \varphi_{m}(t)$ and $W( \varphi_{m}(t) )$ does not change until $ v_{m}(t) \ne 0$ for $t > t_{0}$. 
In this case, all memristor may not switch ``off''.  

We show our computer simulations in Figure \ref{fig:sharpening}.  
Observe that the template (\ref{eqn: template33}) can make the image sharper by enhancing edges and convert into a binary image.   
Thus, the memoristor CNN (\ref{eqn: memristor-cnn1}) can hold a binary output image, even if almost memristos switched ``off'' as shown in Figure \ref{fig:sharpening}(c).  
Thus, we conclude as follow:  
%
%
\begin{center}
\begin{minipage}{12cm}
\begin{shadebox}   
Assume $a_{00} > 1$.  
Then the memristor CNN in Figure \ref{fig:cnn-memistor} can hold a binary output image, even if almost all memristors switch off.
\end{shadebox}
\end{minipage}
\end{center}
%
%

\begin{figure}[h]
 \centering
  \begin{tabular}{ccc}
  \psfig{file=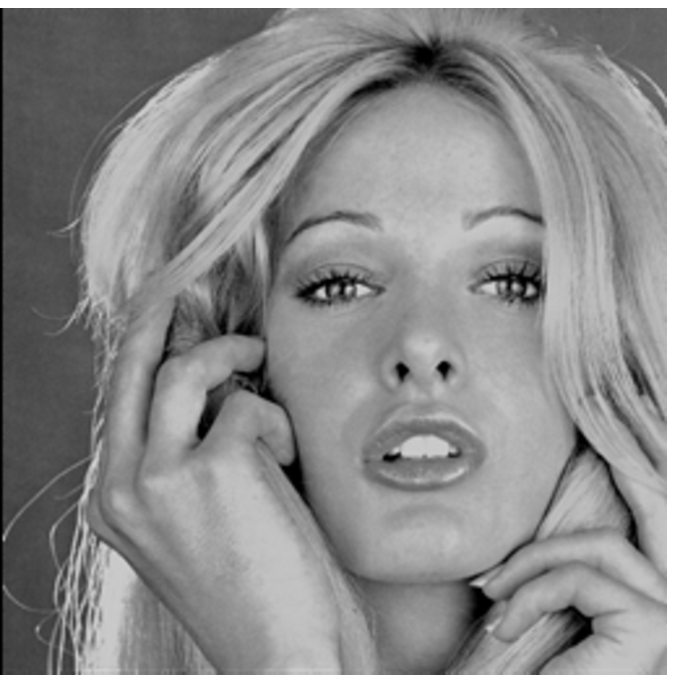,width=5cm} &
  \psfig{file=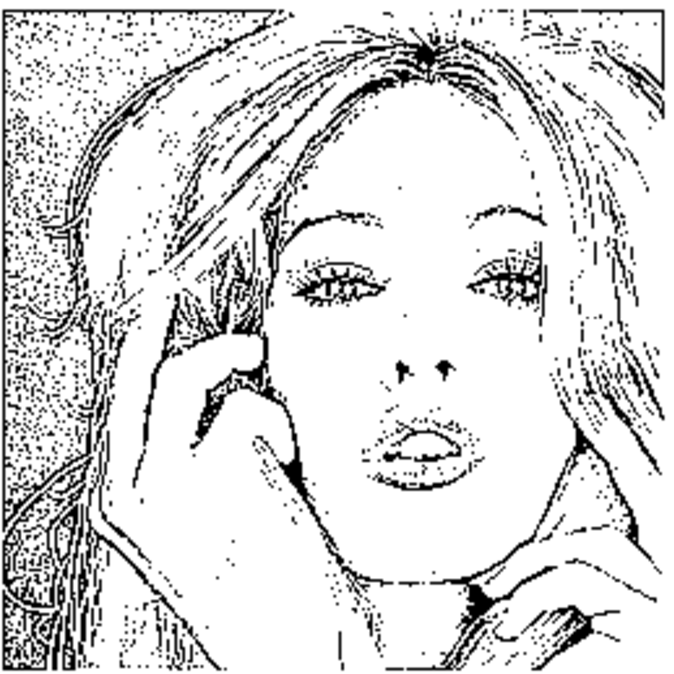,width=5cm} &
  \psfig{file=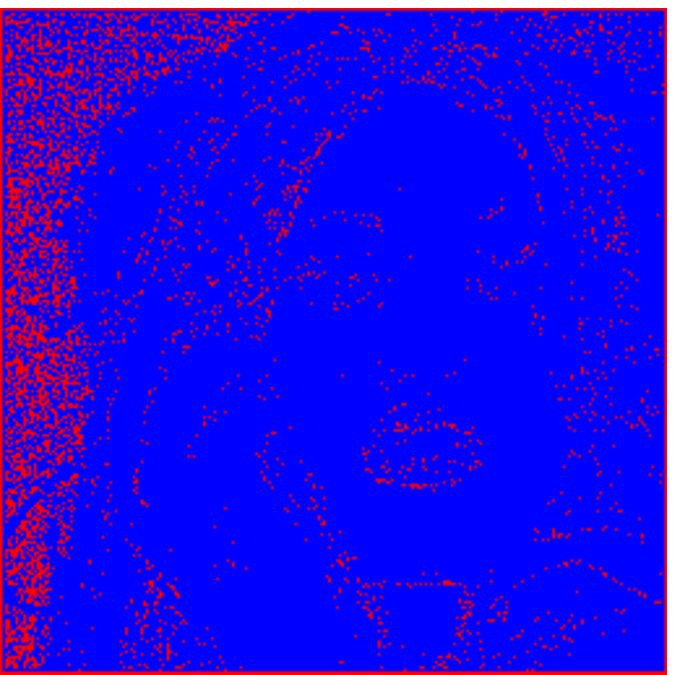,width=5cm} \vspace{1mm} \\
  (a) input image   & (b) output image at $t=100$ & (c)  disconnected cells at $t=100$ \\ 
   && (printed in blue)
  \end{tabular}
  \caption{Image-holding property of the memoristor CNN (\ref{eqn: memristor-cnn1}) 
  with the sharpening with binary output (\ref{eqn: template33}).  
  This template can be used to make an image appear sharper by enhancing edges and convert into a binary image, 
  as shown in Figure \ref{fig:sharpening}(b).  
  The memoristor CNN (\ref{eqn: memristor-cnn1}) can hold a binary output image, 
  even if almost memristos switched ``off'' (printed in blue) 
  as shwon in Figure \ref{fig:sharpening}(c).  
  Here, the red point indicates the cell whose memristor still switches ``on''.   
  The blue point indicates the cell whose memristor switched ``off'',  
  that is, the cell is disconnected from the summing amplifier in Figure \ref{fig:cnn-memistor}.   
  The memristor CNN cell size is $256 \times 256$, since the input image size is $256 \times 256$.  
   }
 \label{fig:sharpening}
\end{figure}
\newpage

%
%
%
\section{Effect of a Parasitic Conductance}
\label{sec: parasitic}
%
%
%

Consider the case where the memristor in Figure \ref{fig:cnn-memistor} has a parallel parasitic conductance $G$ as shown in Figure \ref{fig:para-1}.  
Then, the dynamics of the circuit in Figure \ref{fig:cnn-memistor} is modified into 
\begin{equation}
 \begin{array}{l}
  \displaystyle C \frac{dv_{ij}}{dt} 
   = \displaystyle - \frac{v_{ij}}{R} - i_{r} - i_{m} - i_{G}, \vspace{2mm} \\
    (i, \, j) \in \{ 1, \cdots , M \} \times \{ 1, \cdots , N \},
 \end{array}
\label{eqn: memristor-cnn1-para} 
\end{equation}
where $C=R=1$, and the symbols $v_{ij}$, $i_{m}$, $i_{r}$, and $i_{G}$ denote the voltage across the capacitor $C$, the current through the memristor, and the current trough the nonlinear resistor, and the current through the parallel parasitic conductance $G$, respectively.

\begin{figure}[ht]
 \begin{center} 
  \psfig{file=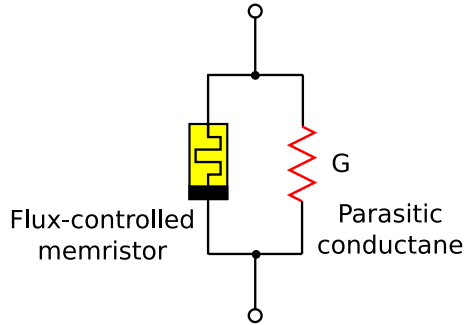,width=6cm}
  \caption{Parasitic conductance $G$ ($0 < G \ll 1$) of the flux-conrolled memristor.  }
 \label{fig:para-1}
 \end{center}
\end{figure}

Furthermore, this circuit satisfies the following relations:  
\begin{enumerate}
\renewcommand{\labelenumi}{(\alph{enumi})}
\item The output $y_{ij}$ and the state $v_{ij}$ of each cell is related via the sign function 
\begin{equation}
  y_{ij} = \operatorname {sgn} (v_{ij}).  
\end{equation}

\item The voltage $v_{m}$ across the memristor is given by  
\begin{equation}
  \scalebox{0.95}{$\displaystyle  v_{m} =   \displaystyle \sum_{k, \, l \in N_{ij}, \ k \ne i, \ l \ne j}
      a_{k l} \ y_{kl} + \sum_{k, \, l \in N_{ij}} b_{k l} \ u_{kl} + z,  $}
\end{equation}
where the first sum does not contain $ a_{ij} \, \operatorname {sgn} (v_{ij})$, where $k=i, \ l=j$.
The current $i_{m}$ through the memristor is given by 
\begin{equation}
 \begin{array}{l}
  i_{m} = W (\varphi_{m})v_{m} \vspace{2mm} \\
        =   \scalebox{0.85}{$\displaystyle   W (\varphi_{m}) \left ( \sum_{k, \, l \in N_{ij}, \ k \ne i, \ l \ne j}
      a_{k l} \ y_{kl} + \sum_{k, \, l \in N_{ij}} b_{k l} \ u_{kl} + z \right ). $}
 \end{array}
\label{eqn: im-para}
\end{equation}
where $W (\varphi_{m})$ denotes the memductance of the memristor and the flux $\varphi_{m}$ of the memristor is defined by 
\begin{equation}
   \varphi_{m}  =  \int_{- \infty}^{t} \, v_{m} \, dt.   
\label{eqn: phi-m-p}
\end{equation}

\item The current $i_{G}$ through the conductance $G$ is given by 
\begin{equation}
   i_{G} = G v_{m}, 
\label{eqn: iG-para}
\end{equation}
where $0 < G \ll 1$. \\

\item The characteristic of the nonlinear resistor is given by 
\begin{equation}
  i_{r}=  f_{r}(v_{r}) = - 0.5 \, a \, ( |v_{r}+ 1 | -  |v_{r} - 1 | ),  
\label{eqn: ir-para}
\end{equation}
where $a$ is a constant.
\end{enumerate} 
Substituting Eqs. (\ref{eqn: im-para}), (\ref{eqn: iG-para}), and (\ref{eqn: ir-para}) into Eq. (\ref{eqn: memristor-cnn1-para}), 
we obtain 
\begin{equation}
 \begin{array}{l}
  \displaystyle \frac{dv_{ij}}{dt} 
   = \displaystyle - v_{ij} + 0.5 \, a \, ( |v_{ij} + 1 | -  |v_{ij} - 1 | ) \vspace{1mm} \\
    \hspace*{15mm}+ ( W (\varphi_{m})  + G )\, v_{m}. \vspace{2mm} \\
 \end{array}
\label{eqn: memristor-cnn1-para-10} 
\end{equation}
Let us assume that the memductance $W (\varphi_{m})$ is given by  Eq. (\ref{eqn: w-phi-10}), that is,  
\begin{equation}
  W (\varphi_{m}) 
  \stackrel{\triangle}{=} 
    \begin{cases}
      1  & -1 \le \varphi_{m} \le 1, \\
      0  &  \varphi_{m} < -1, \ \varphi_{m} > 1.     
    \end{cases}
\label{eqn: w-phi-10-b}
\end{equation}
Then the dynamics of Eq. (\ref{eqn: memristor-cnn1-para-10}) can be described as follow: 
\begin{enumerate}
\item {\bf Case 1.} \ $W (\varphi_{m}) =1$. \\  
From Eq. (\ref{eqn: memristor-cnn1-para-10}), we obtain 
\begin{equation}
  \displaystyle \frac{dv_{ij}}{dt} 
   = \displaystyle - v_{ij} + 0.5 \, a \, ( |v_{ij} + 1 | -  |v_{ij} - 1 | ) + (1 +  \epsilon ) v_{m}, 
\label{eqn: memristor-cnn1-para-3} 
\end{equation}
where $0 < \epsilon \stackrel{\triangle}{=} G \ll 1$.  
Substituting Eq. (\ref{eqn: vm-parasitic}) into Eq. (\ref{eqn: memristor-cnn1-para-3}), we obtain
\begin{equation}
 \begin{array}{l}
  \displaystyle \frac{dv_{ij}}{dt} 
   =   \displaystyle - v_{ij} + 0.5 \, a \, ( |v_{ij} + 1 | -  |v_{ij} - 1 | )  \vspace{2mm} \\
   +  \scalebox{0.85}{$\displaystyle  ( 1 + \epsilon )\left ( \sum_{k, \, l \in N_{ij}, \ k \ne i, \ l \ne j}
      a_{k l} \ y_{kl} + \sum_{k, \, l \in N_{ij}} b_{k l} \ u_{kl} + z  \right ).  $} 
 \end{array}
\label{eqn: memristor-cnn1-para-4} 
\end{equation}
Assume that the qualitative behavior of the modified CNN (\ref{eqn: cnn11})  is not affected by the small perturbation of the template $\{ A, B, z \}$.  
Then the output of Eq. (\ref{eqn: memristor-cnn1-para-4}) is identical to that of the modified CNN, which is given by  
\begin{equation}
 \begin{array}{l}
  \displaystyle \frac{dv_{ij}}{dt} 
   =   \displaystyle - v_{ij} + 0.5 \, a \, ( |v_{ij} + 1 | -  |v_{ij} - 1 | )  \vspace{2mm} \\
   +  \scalebox{0.9}{$\displaystyle  \left ( \sum_{k, \, l \in N_{ij}, \ k \ne i, \ l \ne j}
      a_{k l} \ y_{kl} + \sum_{k, \, l \in N_{ij}} b_{k l} \ u_{kl} + z  \right ). $} 
 \end{array}
\label{eqn: memristor-cnn1-modify-2} 
\end{equation}
\\ 

\item {\bf Case 2.} \ $W (\varphi_{m}) =0$. \\
From Eq. (\ref{eqn: memristor-cnn1-para-10}), we obtain 
\begin{equation}
  \displaystyle \frac{dv_{ij}}{dt} 
   = \displaystyle - v_{ij} + 0.5 \, a \, ( |v_{ij} + 1 | -  |v_{ij} - 1 | ) +  \epsilon  v_{m}, \vspace{2mm} \\
\label{eqn: memristor-cnn1-para-5} 
\end{equation}
where $0 < \epsilon \ll 1$.  
Let us choose the parameter $a$ so that the qualitative behavior of the isolated cell is not affected by the small perturbation $\epsilon  v_{m}$. 
For example, choose $a >1$.  
Then, the qualitative behavior of the driving-point plot of Eqs. (\ref{eqn: memristor-cnn1-para-3}) and (\ref{eqn: memristor-cnn1-para-5}) is not changed, since $\epsilon  v_{m}$ and $(1 +  \epsilon ) v_{m}$ have the same sign.  
Thus, the output $y_{ij}(t) = \operatorname {sgn} (v_{ij}(t))$ is not changed by this small perturbation $\epsilon  v_{m}$.   
In this case, Eq. (\ref{eqn: memristor-cnn1-para-5}) is approximated by 
\begin{equation}
  \displaystyle \frac{dv_{ij}}{dt} 
   = \displaystyle - v_{ij} + 0.5 \, a \, ( |v_{ij} + 1 | -  |v_{ij} - 1 | ).         
\label{eqn: memristor-cnn1-para-6} 
\end{equation}
\end{enumerate} 
It follows that there is not much of a difference between the output images of Eqs. (\ref{eqn: memristor-cnn1}) and (\ref{eqn: memristor-cnn1-para}).   
Note that the CNN template $\{ A, B, z \}$ is usually designed such that the qualitative behavior is not affected by the small perturbation.  
In fact, we could not find any difference between the output images of Eqs. (\ref{eqn: memristor-cnn1}) and (\ref{eqn: memristor-cnn1-para}) when the template is given by  Eq. (\ref{eqn: template31}) or Eq. (\ref{eqn: template33}).    
Thus, we conclude as follow: 

%
%
\begin{center}
\begin{minipage}{12cm}
\begin{shadebox}
Assume that the memductance $W (\varphi_{m})$ is given  by Eq. (\ref{eqn: w-phi-10}), 
and assume that the template is given by Eq. (\ref{eqn: template31}) or Eq. (\ref{eqn: template33}).
Then we could not find any difference between the output images of Eqs. (\ref{eqn: memristor-cnn1}) and (\ref{eqn: memristor-cnn1-para}).     
\end{shadebox}
\end{minipage}
\end{center}
%
%

%
\section{Neuron-like Behavior}
\label{sec: neuron-like}
%

The neurons cannot respond to inputs quickly and they cannot generate outputs rapidly,   
since charging or discharging the membrane potential energy can take time.  
Furthermore, after firing, the neurons have \emph{refractory period}.  
In this subsection, we show that the memoristor CNN (\ref{eqn: memristor-cnn1}) can exhibit the similar behavior.

%
\subsection{Smoothing with binary output}
\label{sec: smoothing}
%
Let us consider the \emph{smoothing with binary output template} \cite{Roska1997}
\begin{equation}
 A =
 \begin{array}{|c|c|c|}
  \hline
   ~0~   &  ~1~   &  ~0~   \\
  \hline
   ~1~   &  ~2~   & ~1~   \\  
  \hline 
   ~0~   &  ~1~   &  ~0~  \\ 
  \hline
  \end{array} \ , \ \ \ 
 B = 
   \begin{array}{|c|c|c|}
  \hline
   ~0~ &  ~0~  &  ~0~ \\
  \hline
   ~0~ &  ~0~  &  ~0~   \\  
  \hline 
   ~0~ &  ~0~  &  ~0~    \\ 
  \hline
  \end{array} \ ,  \ \ \ 
 z =
  \begin{array}{|c|}
  \hline
    0 \\
  \hline
  \end{array} \  .
\label{eqn: template32} 
\end{equation}
The initial condition for the state $v_{ij}$ is equal to a given gray-scale image.  
The boundary condition is given by 
\begin{equation}
  v_{k^{*}l^{*}}  = 0,  \  u_{k^{*}l^{*}}  = 0,
\end{equation}
where $k^{*}l^{*}$ denotes boundary cells. 
This template can be used to smooth (average) a gray-scale image and convert into a binary image.  
It also deletes the noise from the image as shown in Figure \ref{fig:output-4}.

\begin{figure}[h]
 \begin{center} 
  \begin{tabular}{cc}
  \psfig{file=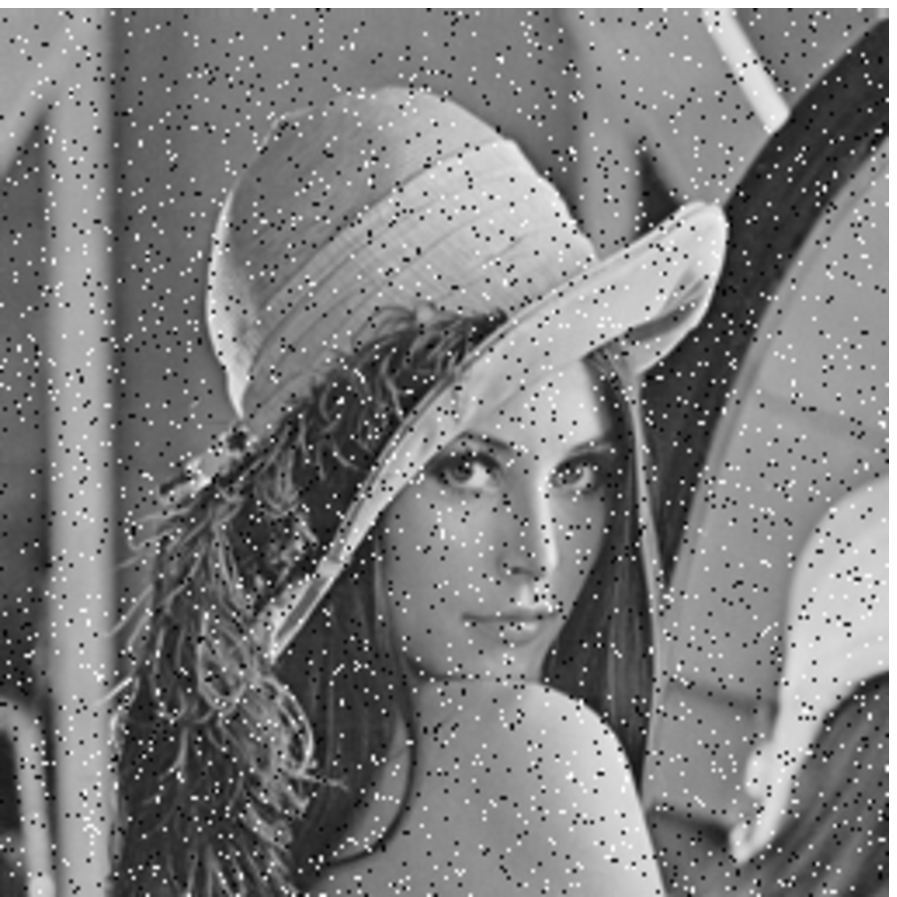,width=6cm} &
  \psfig{file=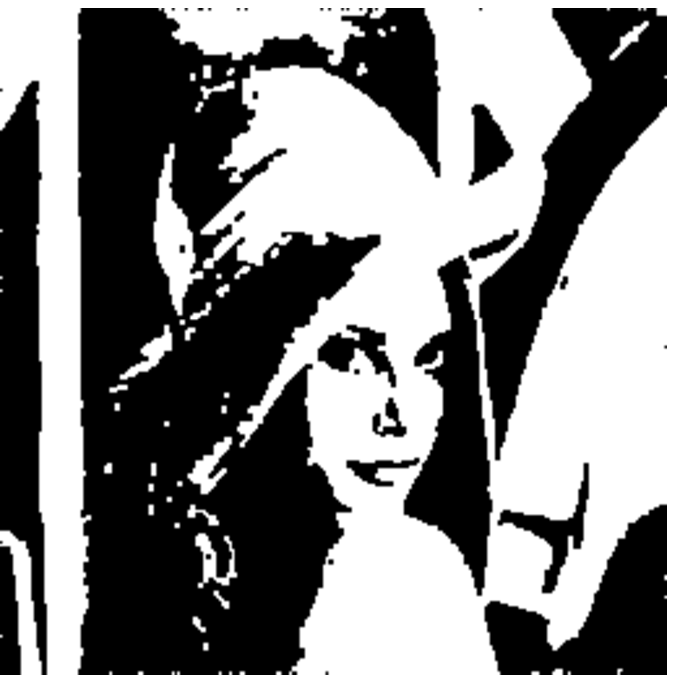,width=6cm} \vspace{1mm} \\
  (a) input image   & (b)   output image   \\ 
  \end{tabular}
 \end{center}
  \caption{Input and output images for the smoothing with binary output template (\ref{eqn: template32}).  
  It can be used to smooth (average) a gray-scale image and convert into a binary image.  
  Hence, it can delete the noise from the image.  Image size is $256 \times 256$. }
 \label{fig:output-4}
\end{figure}
%
%

\begin{figure}[h]
 \centering
  \begin{tabular}{cc}
   \psfig{file=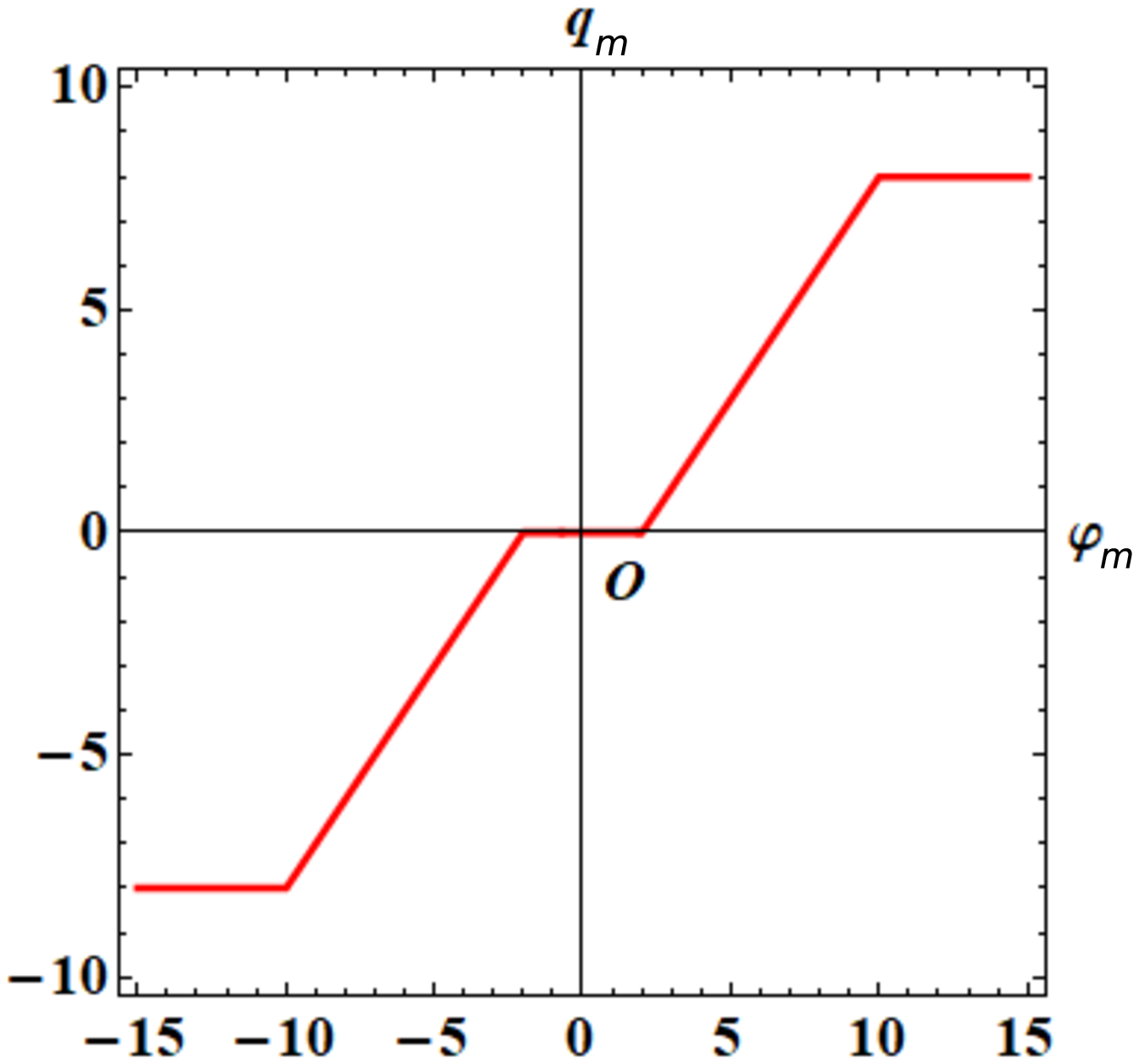, width=6.0cm} & 
   \psfig{file=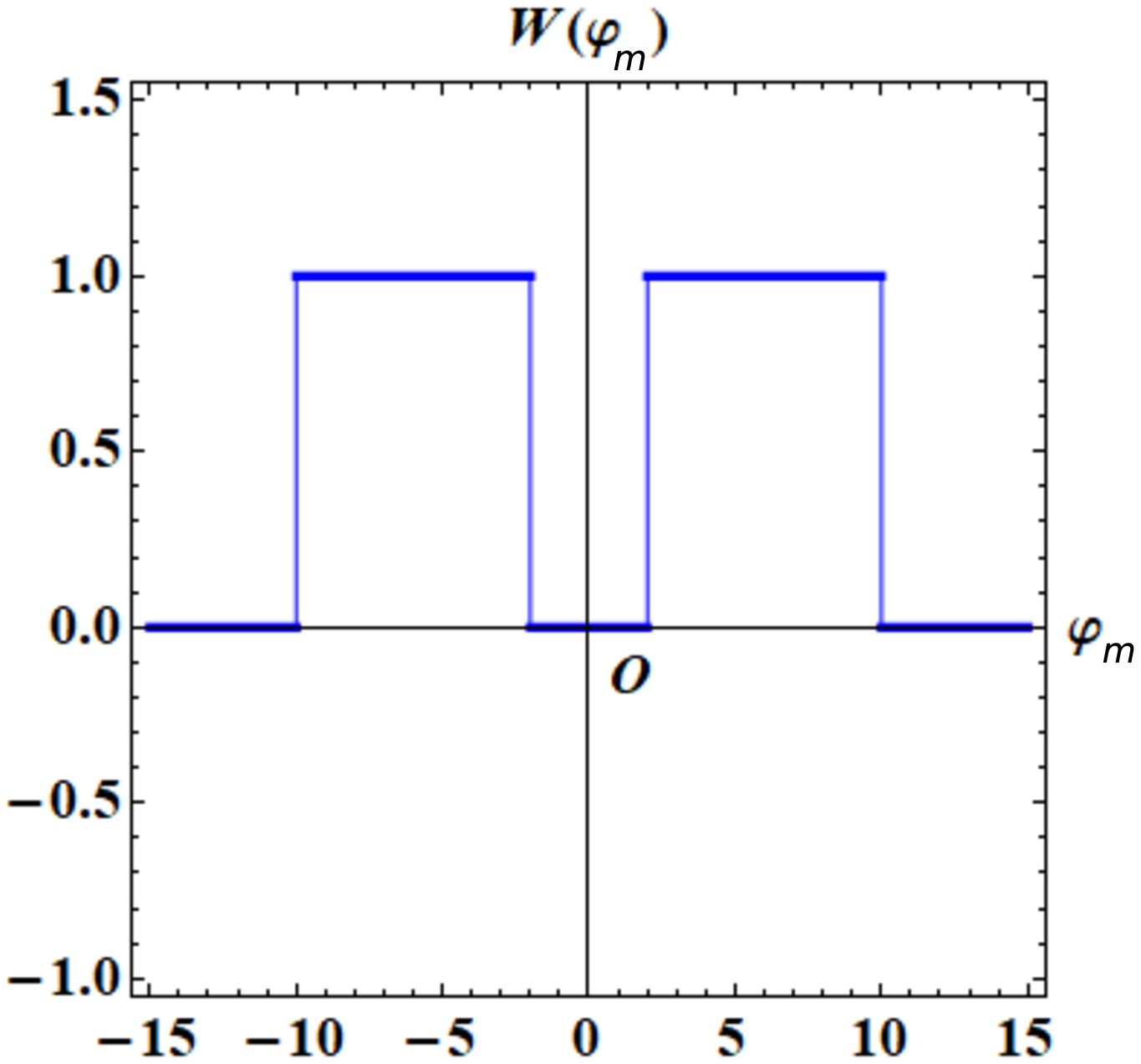, width=6.0cm} \vspace{1mm} \\
   (a) constitutive relation $q_{m} = h (\varphi_{m})$ & 
   (b) memductance  $W( \varphi_{m} )$ \\
  \end{tabular} 
  \caption{Constitutive relation and memductance of the flux-controlled memristor.  
   The memristor switches off and on depending on the value of the flux $\varphi_{m}$. 
   \newline 
   (a) The constitutive relation of the memristor, which is defined by  
       $q_{m} = h(\varphi_{m}) \stackrel{\triangle}{=} 
       0.5( |\varphi _{m} - 2| - |\varphi_{m} - 10| ) - 0.5( |\varphi _{m} + 2| - |\varphi_{m} + 10| )$.   
   \newline
   (b) Memductance $W( \varphi )$ of the memristor, which is defined by   
       $\displaystyle W( \varphi_{m} ) \stackrel{\triangle}{=} \frac{dh(\varphi_{m})}{d\varphi_{m}} 
       = (\mathfrak{s}[|\varphi _{m}| - 2] - \mathfrak{s}[|\varphi_{m}| - 10])$. 
       The symbol $\mathfrak{s}  [\, z \,]$ denotes the \emph{unit step} function, 
       equal to $0$ for $z < 0$ and 1 for $z \ge 0$.   }
 \label{fig:relation-30}
\end{figure}
%
%

Consider the memristor CNN circuit in Figure \ref{fig:cnn-memistor}.  
Suppose that the flux-controlled memristor has the following constitutive relation and memductance:  
\begin{equation}
\begin{array}{llc}
 q_{m} = h(\varphi_{m}) &\stackrel{\triangle}{=} & 0.5 \bigl ( \, |\varphi _{m} - 2| - |\varphi_{m} - 10| \, \bigr )  \vspace{2mm} \\      
        && - 0.5 \bigl ( \, |\varphi _{m} + 2| - |\varphi_{m} + 10| \, \bigr ), 
\end{array}
\label{eqn: qm-hm}
\end{equation}
and 
\begin{equation}
 \begin{array}{l}
 W( \varphi_{m} ) = \displaystyle \frac{dh(\varphi_{m})}{d\varphi_{m}} \vspace{2mm} \\ 
  \stackrel{\triangle}{=}  
   \scalebox{0.95}{$\displaystyle\left ( \mathfrak{s} \Bigl [ \, | \varphi _{m}| - 2 \, \Bigr ] 
   - \mathfrak{s} \Bigl [ \, |\varphi_{m}| - 10 \Bigr ] \right )  $} \vspace{4mm} \\
     = \left \{
     \begin{array}{ccccc} 
      0~~~~              & & \varphi_{m} & \le -10,        \vspace{2mm} \\ 
      1~~~~              &  -10 <  & \varphi_{m} & < -2,   \vspace{2mm} \\ 
      0~~~~              &  -2 \le & \varphi_{m} & \le 2,  \vspace{2mm} \\ 
      1~~~~              &   2 <   & \varphi_{m} & < 10,   \vspace{2mm} \\  
      0~~~~              &  10 \le & \varphi_{m}, &    
    \end{array}
    \right. 
\end{array}
\label{eqn: wm}
\end{equation}
respectively (see Figure \ref{fig:relation-30}).  
Its terminal voltage $v_{m}$  satisfies 
\begin{equation}
 \begin{array}{l}
 v_{m}(t) = \scalebox{0.95}{$\displaystyle 
                \displaystyle \sum_{k, \, l \in N_{ij}, \ k \ne i, \ l \ne j} a_{k l} \ y_{kl}(t) $} \vspace{2mm} \\
          = a_{-1, \, 0}~y_{i-1, \, j}(t)  + a_{0, \, -1}~y_{i, \, j-1}(t) + a_{0, \, 1}~y_{i, \, j+1}(t) \vspace{2mm} \\
          ~~ + a_{1, \, 0}~y_{i+1, \, j} (t)   
             \vspace{3mm} \\
          =  y_{i-1, \, j}(t) +  y_{i, \, j-1} (t) +  y_{i, \, j+1} (t) +  y_{i+1, \, j}(t),  
  \end{array}
\end{equation}
where 
\begin{equation}
 \begin{array}{l}
   a_{-1, \, 0} = a_{0, \, -1} = a_{0, \, 1} = a_{1, \, 0} = 1,  \vspace{2mm} \\
   y_{kl}(t) \in  \{ -1, \, 0, \, 1 \}.  
  \end{array}
\end{equation}
Thus, the terminal voltage $v_{m}$ satisfies 
\begin{equation}
 |v_{m}(t)| \ge 0.  
\label{eqn: abs-vm-2}
\end{equation}
If $ v_{m}(t)$ becomes zero at $t=t_{0}$, then $W( \varphi_{m}(t) )$ does not change until $ v_{m}(t) \ne 0$ for $t > t_{0}$. 
Compare Eq. (\ref{eqn: abs-vm-2}) with Eq. (\ref{eqn: abs-vm-1}).

Our computer simulations are shown in Figures \ref{fig:output-6} and \ref{fig:output-5}.  
We explain their results briefly.  
\begin{enumerate}
\item The memoristor CNN can \emph{not} remove the noise from the input image at the initial stage.    
It is due to the reason that charging flux to memristors can take time, and the memristor can not switch ``on'' rapidly.  

\item When time $t$ is increased, almost memristors change from the ``switch-off'' state to the ``switch-on'' state, and    
the memoristor CNN (\ref{eqn: memristor-cnn1}) can delete the noise from the given image by smoothing.  
It can also hold the output image even if almost memristors switched ``off''.  
\end{enumerate}
The detailed behavior of the memoristor CNN (\ref{eqn: memristor-cnn1}) can be described as follow:
\begin{enumerate}
\item  $t = 0.01$ \\    
All memristors switched ``off''.  
That is, all cells are printed in blue as shown in Figure \ref{fig:output-6}(b).  
Thus, the memoristor CNN (\ref{eqn: memristor-cnn1}) can only convert the given gray-scale image to the binary image.   
It cannot delete the noise from the image.   
It is due to the reason that the memoristor CNN (\ref{eqn: memristor-cnn1}) cannot respond to the input quickly, since the memristors cannot switch ``on'' rapidly.  
\vspace{2mm}\\ 

\item $t = 0.5$,  $t = 2$, and $t = 5$ \\ 
Many memristors change to the ``switch-on'', and then they switch ``off'' again, as shown in Figures \ref{fig:output-6}(c), (d), and (e).  
Note that all memristors do not change between the ``switch-off'' and ''switch-on'' states, \emph{synchronously}.  
Furthermore, only the cell with a \emph{``switched-on memristor''} can smooth (average) the given image, and delete the noise.  
\vspace{2mm}\\ 

\item $t = 15$ \\ 
The memoristor CNN (\ref{eqn: memristor-cnn1}) can hold the output image, even if almost all memristors switched ``off''.   
Several cells still switch ``on'' (they are marked by yellow circles and arrows in Figure \ref{fig:output-6}(f)).  
%
%
\end{enumerate}
%
%
\begin{figure}[p]
 \centering
  \begin{tabular}{ccc}
   \psfig{file=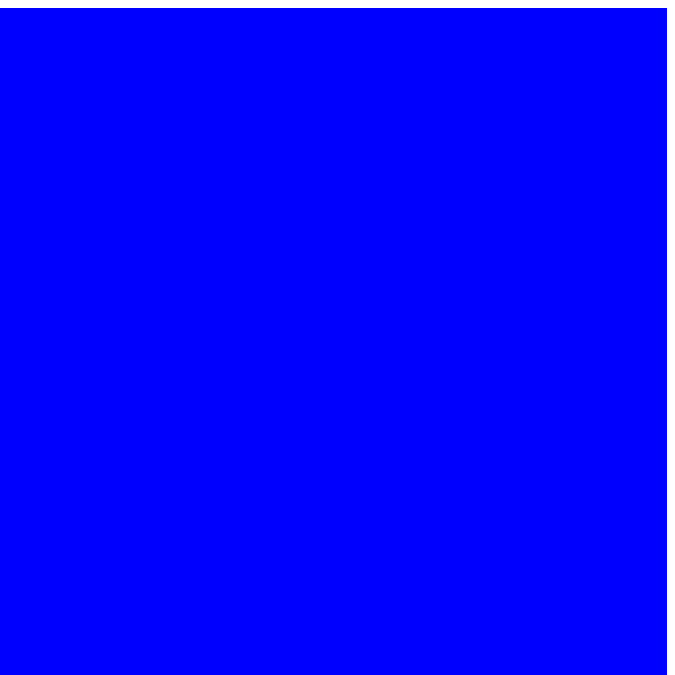,width=5cm} &
   \psfig{file=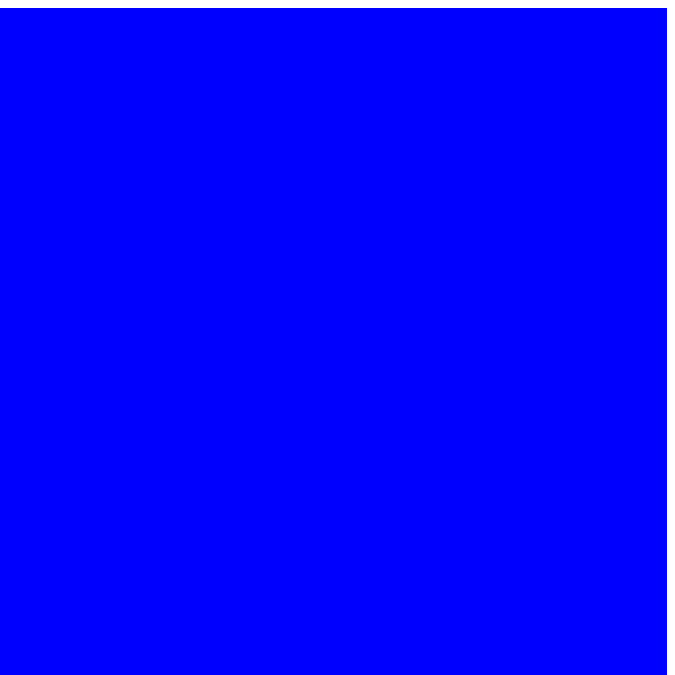,width=5cm} &
   \psfig{file=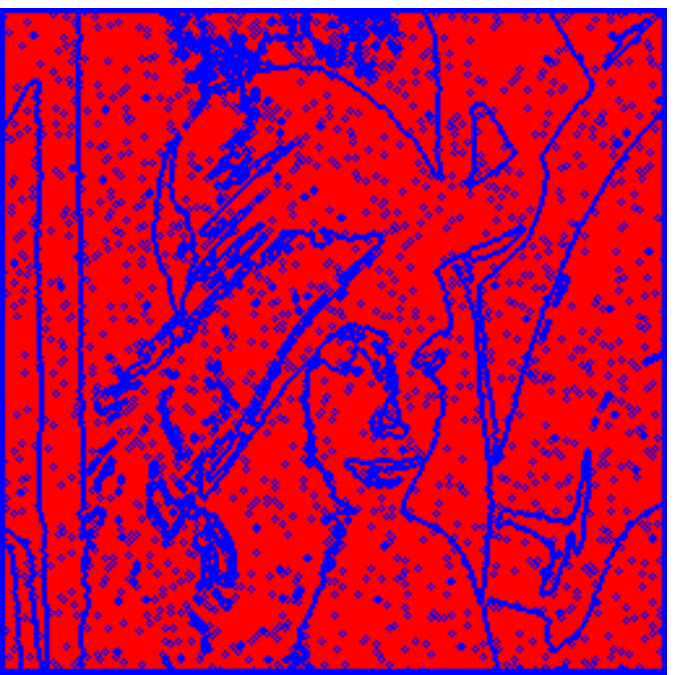,width=5cm} \vspace{2mm} \\
  (a) $t=0$  & (b) $t=0.01$ &  (c) $t=0.5$  \vspace{5mm} \\
  \psfig{file=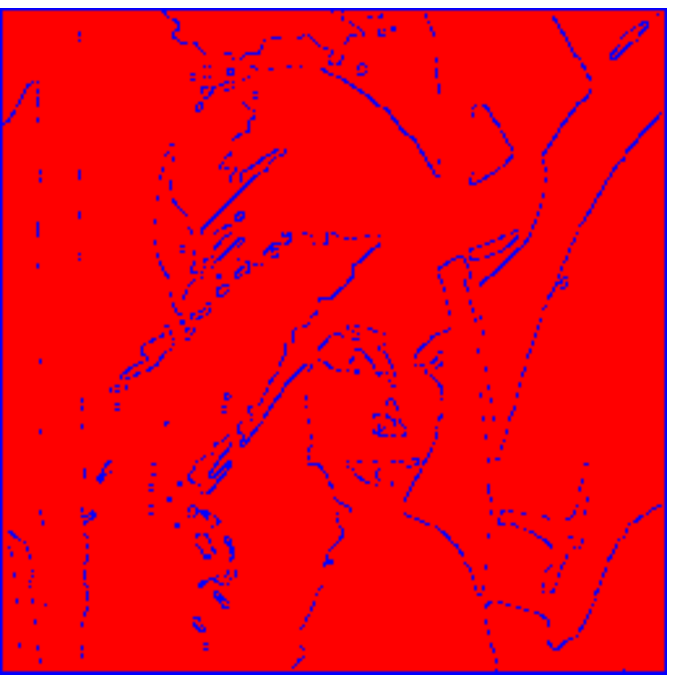,width=5cm} &  
  \psfig{file=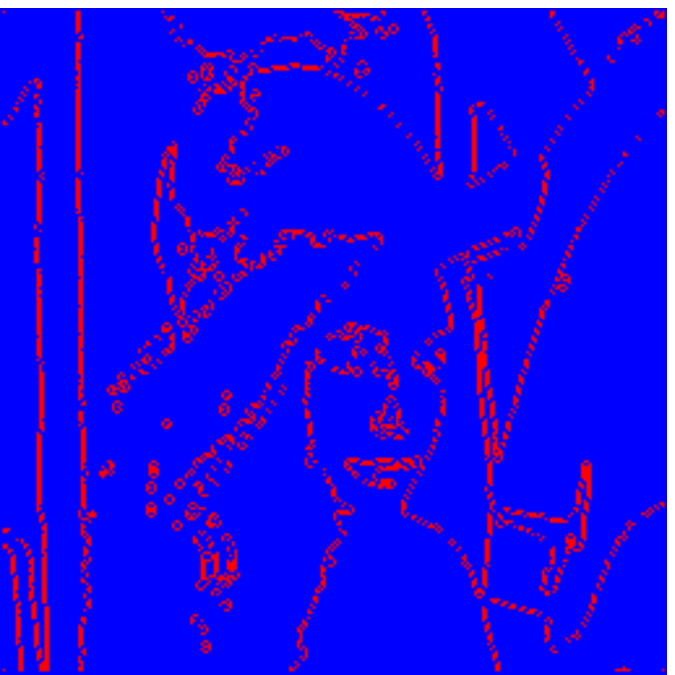,width=5cm} &
  \psfig{file=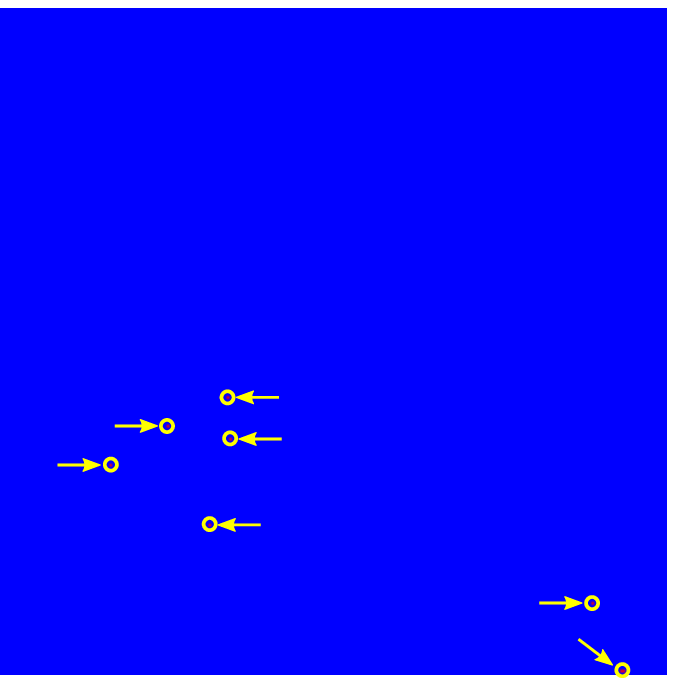,width=5cm} \vspace{2mm} \\  
  (d) $t=2$ & (e) $t=5$ &  (f) $t=15$  \\
  \end{tabular}
  \caption{``Switch-on'' and ``switch-off'' states.    
   Red point indicates the cell whose memristor switched ``on'', 
   and the blue point indicates the cell whose memristor switched ``off''.   
   The memristor CNN cell size is $256 \times 256$, since the input image size is $256 \times 256$. 
   \newline
   (a) All memristors switch ``off''.  Thus, all cells are printed in blue.  
   \newline  
   (b) All memristors switch ``off'', since the memristor can not switch ``on'' rapidly.  
   \newline
   (c) Many memristors is turning from the ``switch-off'' state (printed in blue) to the ``switch-on'' state (printed in red).   
   \newline
   (d) Most memristos turn to the ``switch-on'' (printed in red).    
   \newline
   (e) Most memristos turn from the ``switch-on'' state (printed in red) to the ``switch-off'' state (blue), again.          
   \newline  
   (f) Almost all memristos are switched ``off'' (printed in blue), except several cells (marked by yellow circles and arrows).  }
 \label{fig:output-6}
\end{figure}
%
%

\begin{figure}[p]
 \centering
  \begin{tabular}{ccc}
   \psfig{file=./figure/noise5.eps,width=5cm} &
   \psfig{file=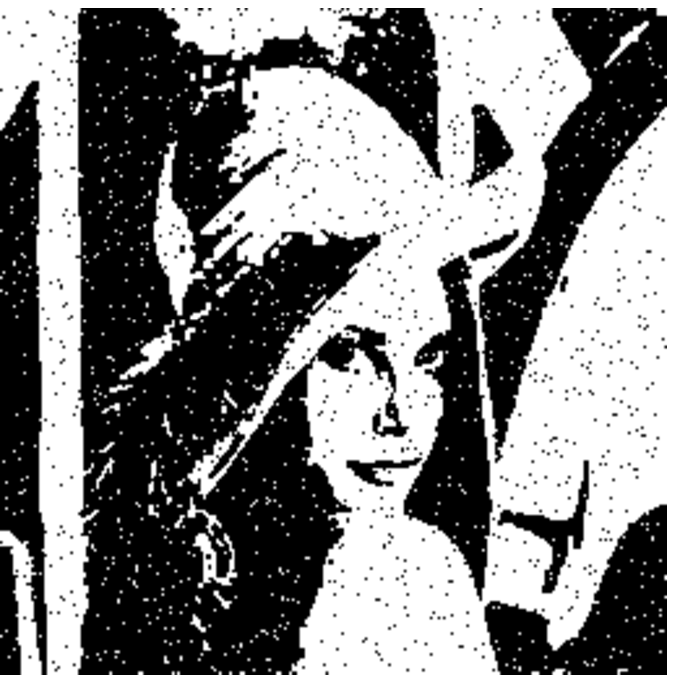,width=5cm}  &
   \psfig{file=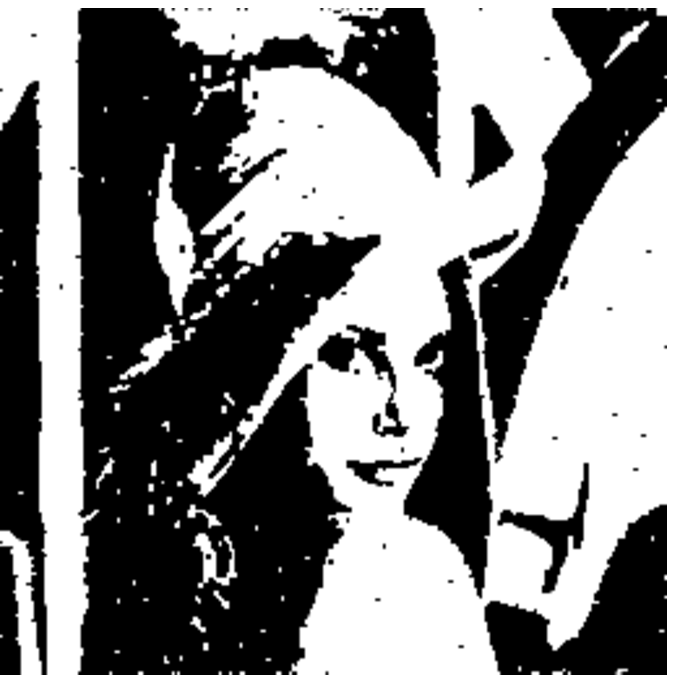,width=5cm} \vspace{2mm} \\
   (a) input gray-scale image ($t=0$)  & (b) output image ($t=0.01$) & (c) output image ($t=0.5$) \vspace{5mm} \\ 
   \psfig{file=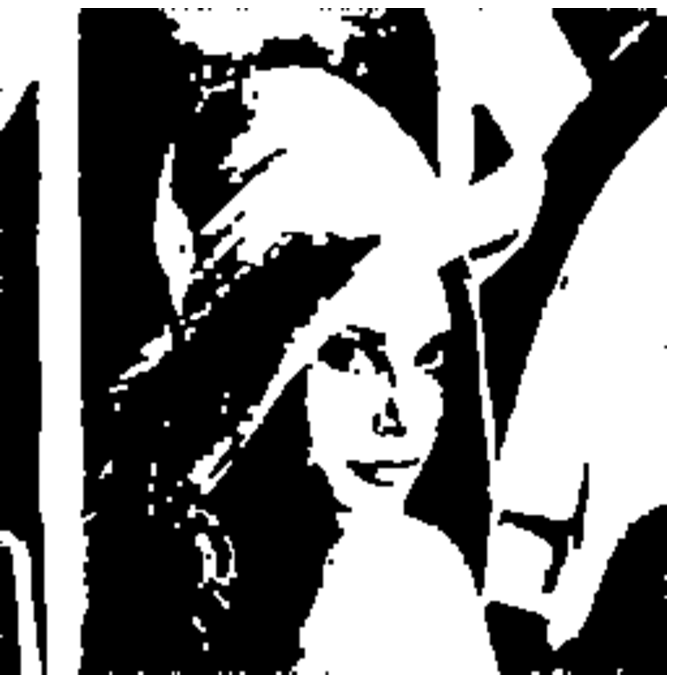,width=5cm} & 
   \psfig{file=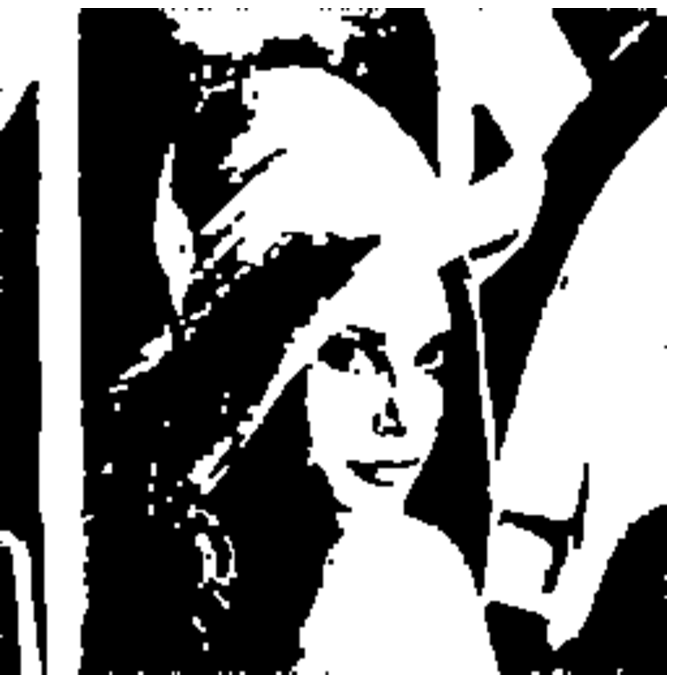,width=5cm} &
   \psfig{file=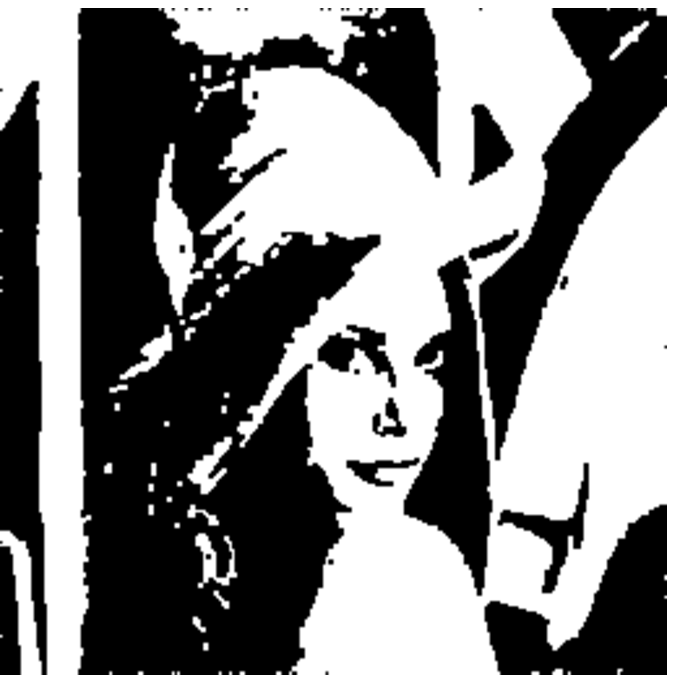,width=5cm}  \\  
   (d) output image ($t=2$) & (e) output image ($t=5$) &  (f) output image ($t=15$)  \\
  \end{tabular}
  \caption{Neuron-like behavior for the memoristor CNN (\ref{eqn: memristor-cnn1}) 
   with the smoothing with binary output template (\ref{eqn: template32}).   
   Observer that the memoristor CNN (\ref{eqn: memristor-cnn1}) cannot respond to the input quickly, 
   and it can hold the output image even if almost all memristors switch ``off''.  
   \newline
   (a)  Gray-scale input image for the memoristor CNN (\ref{eqn: memristor-cnn1}).  
        All memristors switch ``{\bf off}''.
   \newline  
   (b) The memoristor CNN (\ref{eqn: memristor-cnn1}) does \emph{not} delete the noise from the image, 
       since {\bf all memristors} still switch ``{\bf off}'', that is, the memristor can not switch ``on'' rapidly. 
   \newline
   (c) The memoristor CNN (\ref{eqn: memristor-cnn1}) can \emph{not} delete \emph{all noise} from the image,    
       though {\bf many memristors} are turning from the ``switch-off'' state to the ``{\bf switch-on}'' state.    
   \newline
   (d) The memoristor CNN (\ref{eqn: memristor-cnn1}) deleted the noise from the image,    
       since {\bf most memristos} turn to the ```{\bf switch-on}''.  
   \newline 
   (e) The memoristor CNN (\ref{eqn: memristor-cnn1}) deleted the noise from the image. 
       At this point in time, {\bf most memristos} turn from the ``switch-on'' to the ``{\bf switch-off}'', {\bf again}.         
   \newline  
   (f) The memoristor CNN (\ref{eqn: memristor-cnn1}) can hold a binary output image, 
       even if {\bf almost all memristos} switched ``{\bf off}'', except for several cells 
      (which is marked by yellow circles and arrows in Figure \ref{fig:output-6}(f)).}
 \label{fig:output-5}
\end{figure}
\clearpage

%
\subsection{Erosion}
\label{sec: erosion}
%
Let us consider the \emph{erosion template} \cite{{Chua1998},{Roska}}
\begin{equation}
\begin{array}{l}
 A =
 \begin{array}{|c|c|c|}
  \hline
   ~0~   &  ~0~   &  ~0~   \\
  \hline
   ~0~   &  ~2~  & ~0~   \\  
  \hline 
   ~0~   &  ~0~   & ~0~  \\ 
  \hline
  \end{array} \ , \ \ \ 
 B = 
   \begin{array}{|c|c|c|}
  \hline
   ~0~ &  ~1~  &  ~0~ \\
  \hline
   ~1~ &  ~1~  &  ~1~   \\  
  \hline 
   ~0~ &  ~1~  &  ~0~    \\ 
  \hline
  \end{array} \ ,  \ \ \ 
 z =
  \begin{array}{|c|}
  \hline
    -4.5 \\
  \hline
  \end{array} \ . 
\end{array}
\label{eqn: template35} 
\end{equation}
The initial condition for $v_{ij}$ is  given by 
\begin{equation}
  v_{ij}(0)  = 0,   
\end{equation}
and the input $u_{kl}$ is equal to a given binary image.  
The boundary condition is given by 
\begin{equation}
  v_{k^{*}l^{*}}  = -1,  \  u_{k^{*}l^{*}}  = -1,
\end{equation}
where $k^{*}l^{*}$ denotes boundary cells. 
This template can be used to peel off all boundary pixels of binary image objects, as shown in Figure \ref{fig:text-1}. 

\begin{figure}[h]
 \centering
  \begin{tabular}{cc}
  \psfig{file=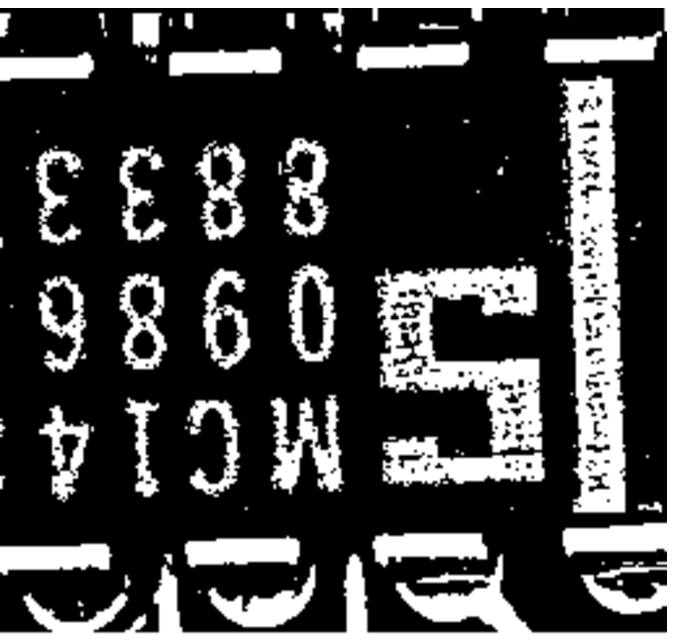,width=6cm} &
  \psfig{file=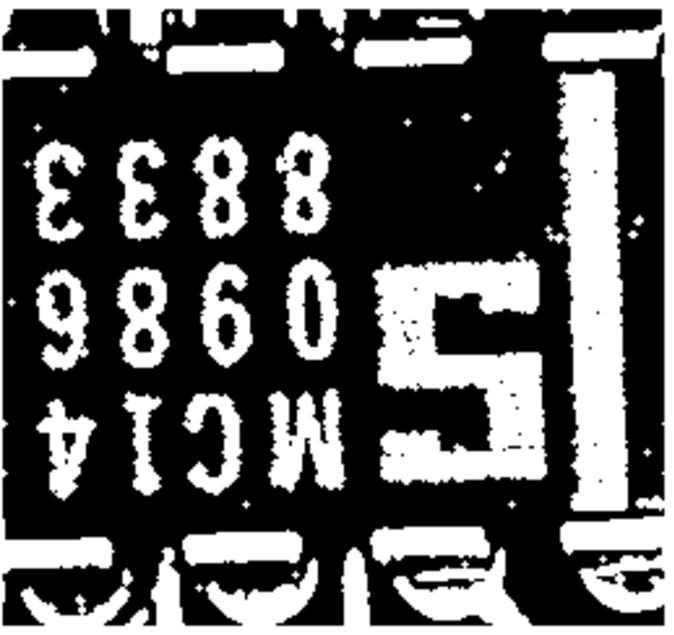,width=6cm} \vspace{1mm} \\
  (a) input image   & (b)   output image   \\ 
  \end{tabular}
  \caption{Input and output images for the erosion template (\ref{eqn: template35}).  
  It can be used to peel off all boundary pixels of binary image objects.   
  Hence, the text becomes clearly visible by this template.  Image size is $256 \times 256$. }
 \label{fig:text-1}
\end{figure}

Consider the memristor CNN circuit in Figure \ref{fig:cnn-memistor}.  
Suppose that the constitutive relation and memductance of the flux-controlled memristor are given by Eqs. (\ref{eqn: qm-hm}) and (\ref{eqn: wm}), respectively (see Figure \ref{fig:relation-30}).  
Thus, its terminal voltage $v_{m}$  satisfies 
\begin{equation}
\begin{array}{lll}
  |v_{m}| &=& \scalebox{0.95}{$\displaystyle    \left | \displaystyle \sum_{k, \, l \in N_{ij}, \ k \ne i, \ l \ne j}
      a_{k l} \ y_{kl} + \sum_{k, \, l \in N_{ij}} b_{k l} \ u_{kl} + z \right |  $} \vspace{2mm} \\
          &=&  \displaystyle   \left |  \sum_{k, \, l \in N_{ij}} b_{k l} \ u_{kl} + 4.5 \right |   \vspace{2mm} \\ 
          &=&  \displaystyle  \Bigl | u_{i-1, \, j}  +  u_{i, \, j-1} + u_{i, \, j} + u_{i, \, j+1}        \vspace{2mm} \\ 
           &&   ~~~+ u_{i+1, \, j} - 4.5  \Bigr |  \ge  0.5,
\label{eqn: abs-vm-100}
\end{array}
\end{equation}
where 
\begin{equation}
\left.
 \begin{array}{l}
   b_{i-1, \, j} =  b_{i, \, j-1} = b_{i, \, j} = b_{i, \, j+1} = b_{i+1, \, j}  =  1, \vspace{2mm} \\ 
   u_{kl} = \pm 1.
 \end{array}
\right \}
\end{equation}
For example, if we assume that the inputs $u_{kl}$ satisfy 
\begin{equation}
  u_{i-1, \, j} =  u_{i, \, j-1} = u_{i, \, j} = u_{i, \, j+1} = u_{i+1, \, j}  =  1,  
\label{eqn: u-1}
\end{equation}
then we obtain
\begin{equation}
\begin{array}{lll}
  v_{m} &=& u_{i-1, \, j}  +  u_{i, \, j-1} + u_{i, \, j} + u_{i, \, j+1} + u_{i+1, \, j} - 4.5  \vspace{2mm} \\
        &=& 5 - 4.5  = 0.5.
\end{array}
\end{equation}
In this case, we obtain from Eq. (\ref{eqn: phi-m-1})
\begin{equation}
  \varphi_{m}(t) = 0.5t, 
\end{equation}
where we assume $\varphi_{m}(0)=0$.  

Assume next that the constitutive relation and memductance of the flux-controlled memristor are given by Eqs. (\ref{eqn: qm-hm}) and (\ref{eqn: wm}), respectively.  
Then, we obtain 
\begin{equation}
  W (\varphi_{m}(t)) 
  =
    \begin{cases}
      0  &  0  \le t < 4,  \\
      1  &  4  \le t \le 20,  \\
      0  &  20 < t,   
    \end{cases}
\end{equation}
where the input $u_{kl}$ is given by Eq. (\ref{eqn: u-1}) and $\varphi_{m}(t) = 0.5t$.   
Note that other inputs may satisfy the equation $W (\varphi_{m}(t))=1$ more quickly than the above.  
That is, all memristors do \emph{not} switch ``on'' \emph{synchronously}, since their terminal flux are not always identical.  

Our computer simulations are shown in Figures \ref{fig:text-100} and \ref{fig:text-200}.  
The boundary pixels of the given binary image are peeled off,   
and the text becomes clearly visible.  
We conclude as follow: \\ \\
%
%
\begin{center}
\begin{minipage}{14cm}
\begin{shadebox}
Suppose that the memoristor CNN (\ref{eqn: memristor-cnn1}) has the following property:  
\begin{enumerate}
\renewcommand{\theenumi}{\alph{enumi}}
\item The feedback, control, and threshold parameters are given by Eq. (\ref{eqn: template32}) or (\ref{eqn: template35}). 
\item The constitutive relation and the memductance of the flux-controlled memristors are 
given by Eq. (\ref{eqn: qm-hm}) and (\ref{eqn: wm}), respectively.  
\end{enumerate}
Then, the memoristor CNN (\ref{eqn: memristor-cnn1}) can exhibit the following behavior:
\begin{enumerate}
\item The memoristor CNN (\ref{eqn: memristor-cnn1}) cannot respond to the input quickly, since the memristors cannot switch ``on'' rapidly.  
\item The memoristor CNN (\ref{eqn: memristor-cnn1}) can hold the output image even if almost all memristors switch ``off'' (refractory period).  
\end{enumerate}
\end{shadebox}
\end{minipage}
\end{center}
%
%

\clearpage
\begin{figure}[h]
 \centering
  \begin{tabular}{ccc}
   \psfig{file=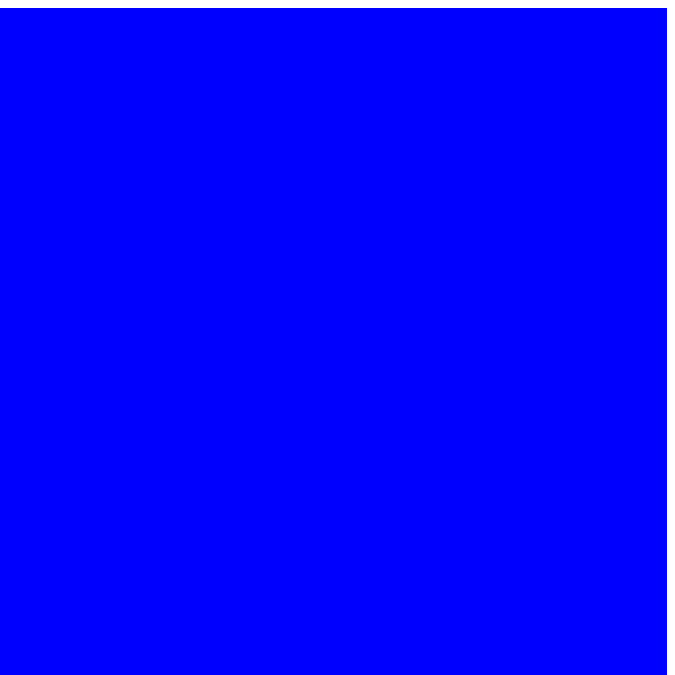,width=4cm} &
   \psfig{file=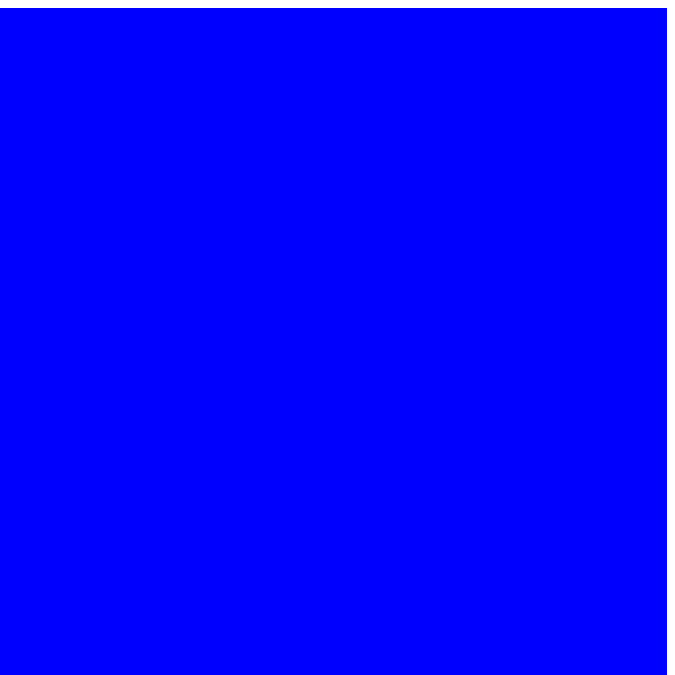,width=4cm} &
   \psfig{file=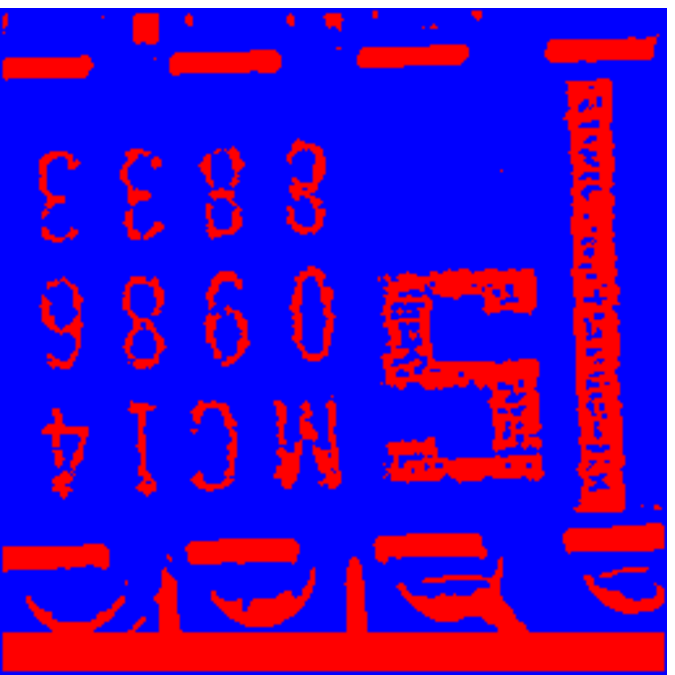,width=4cm} \vspace{2mm} \\
  (a) $t=0$  & (b) $t=0.1$ &  (c) $t=0.3$  \vspace{5mm} \\
  \psfig{file=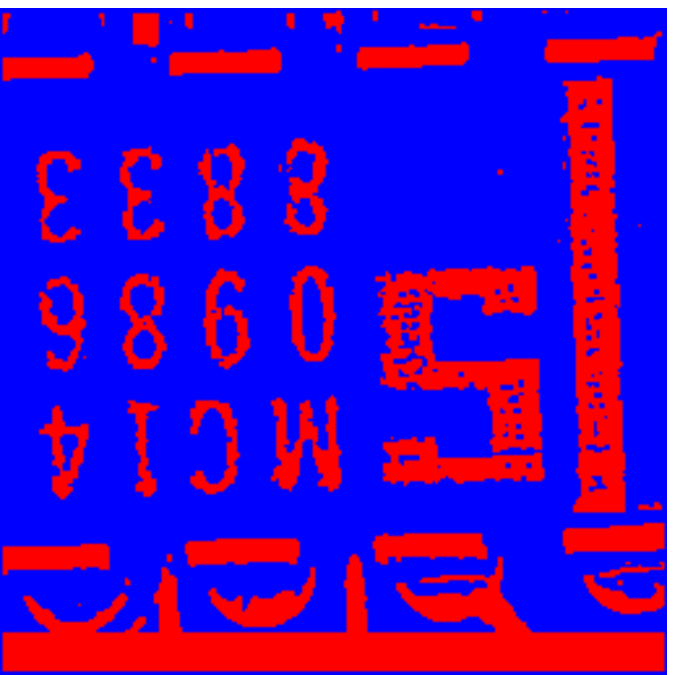,width=4cm} &  
  \psfig{file=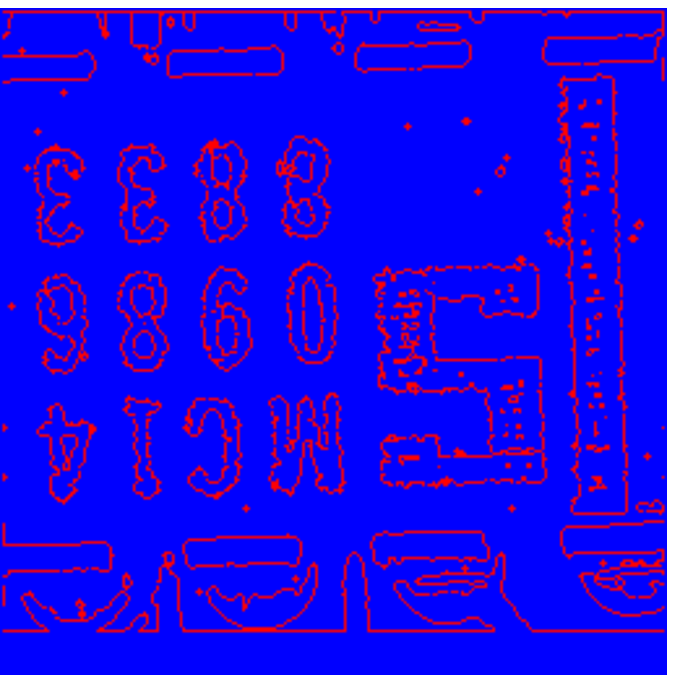,width=4cm} &
  \psfig{file=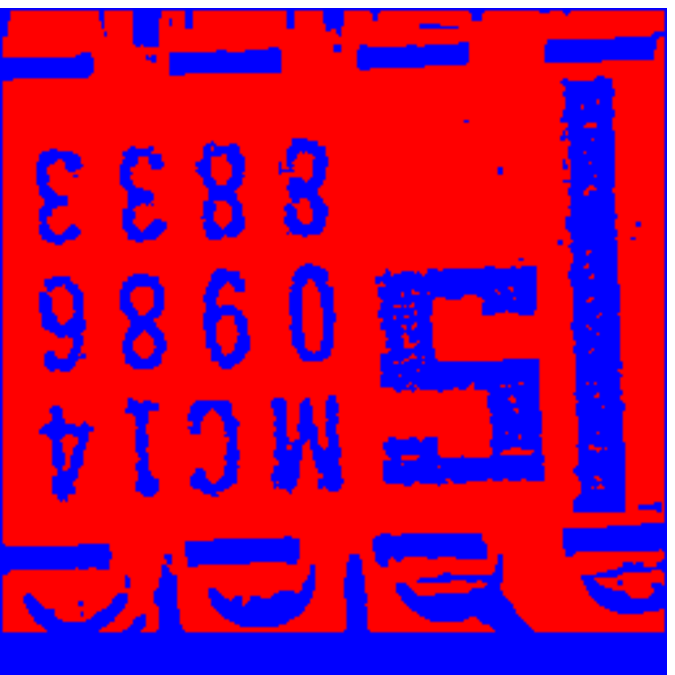,width=4cm} \vspace{2mm} \\  
  (d) $t=0.5$ & (e) $t=2$ &  (f) $t=4$    \vspace{5mm} \\  
  \psfig{file=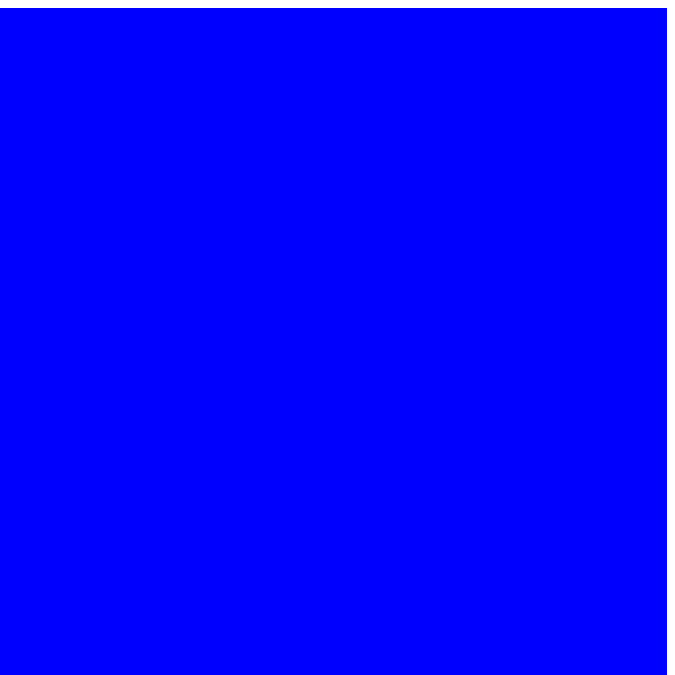,width=4cm} &
  \psfig{file=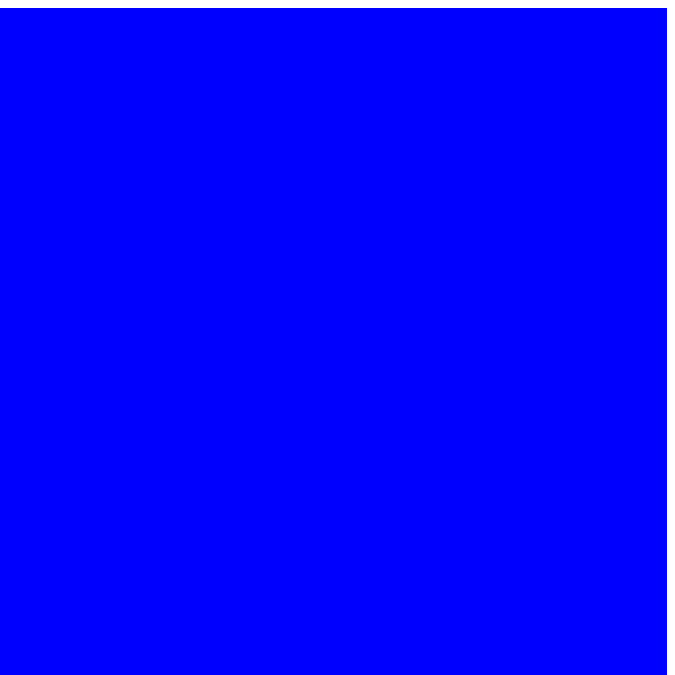,width=4cm} & \vspace{2mm} \\  
  (g) $t=20$ & (h) $t=50$ &  
  \end{tabular}
  \caption{``Switch-on'' and ``switch-off'' states of memristors.    
   Red point indicates the cell whose memristor switched``on'', 
   and the blue point indicates the cell whose memristor switched ``off''.   
   The memristor CNN cell size is $256 \times 256$, since the input image size is $256 \times 256$. 
   \newline
   (a) All memristors switch ``off''.  Thus, all cells are printed in blue. 
   \newline  
   (b) All memristors switch ``off'', since the memristor can not switch ``on'' rapidly.  
   \newline
   (c)-(d) Memristors are turning from the ``switch-off'' state (printed in blue) to the ``switch-on'' state (printed in red).
   \newline
   (e) ``Switch-on'' memristors (printed in red) are turning to the ``switch-off'' (printed in blue).      
   \newline
   (f) ``Switch-off'' memristors (printed in blue), which do \emph{not} have switched ``on'' \emph{yet}, 
   turn to the ``switch-on'' state (printed in red).  
   \newline
   (g)-(h) All memristos switch ``off'', again.  Thus, all cells are printed in blue. }
 \label{fig:text-100}
\end{figure}

\clearpage
\begin{figure}[h]
 \centering
  \begin{tabular}{ccc}
   \psfig{file=./figure/Text-2-t-0.eps,width=4cm} &
   \psfig{file=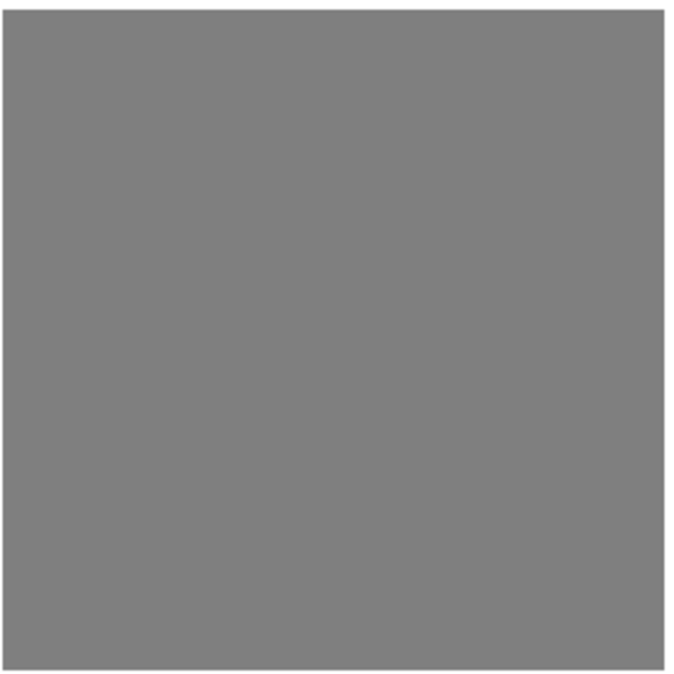,width=4cm} &
   \psfig{file=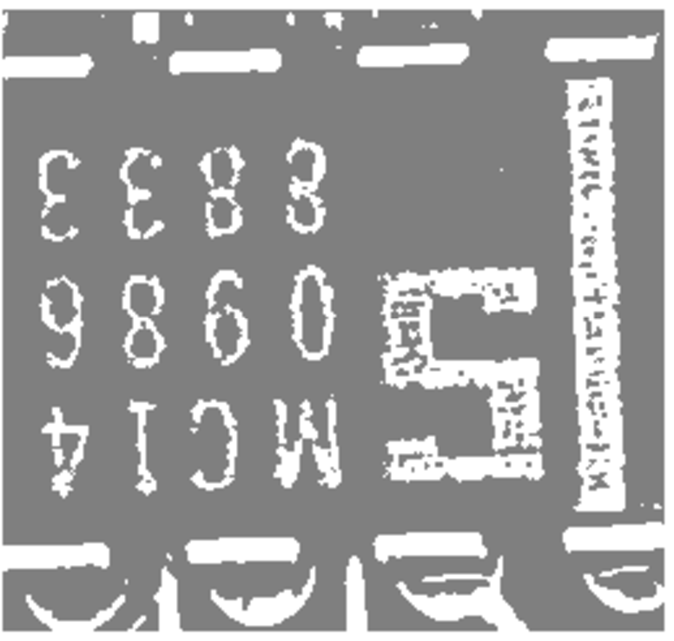,width=4cm} \vspace{2mm} \\
  (a) input binary image ($t=0$)  & (b) output image ($t=0.1$) &  (c) output image ($t=0.3$)  \vspace{5mm} \\
  \psfig{file=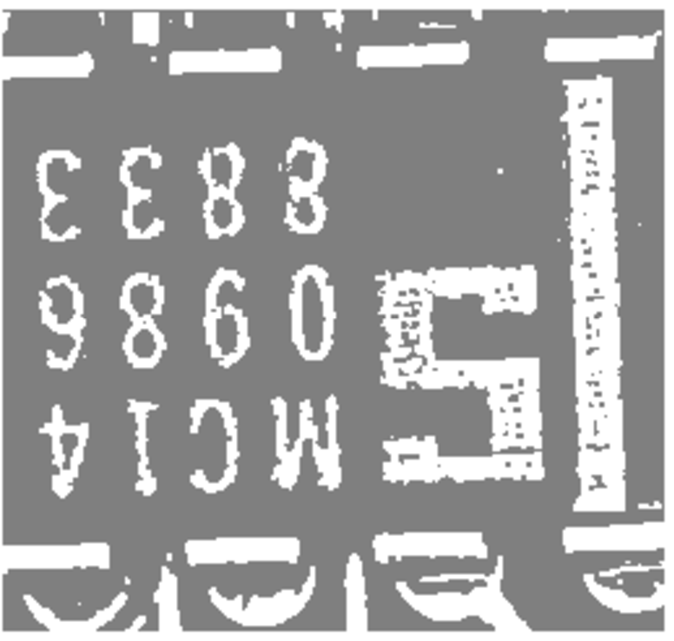,width=4cm} &  
  \psfig{file=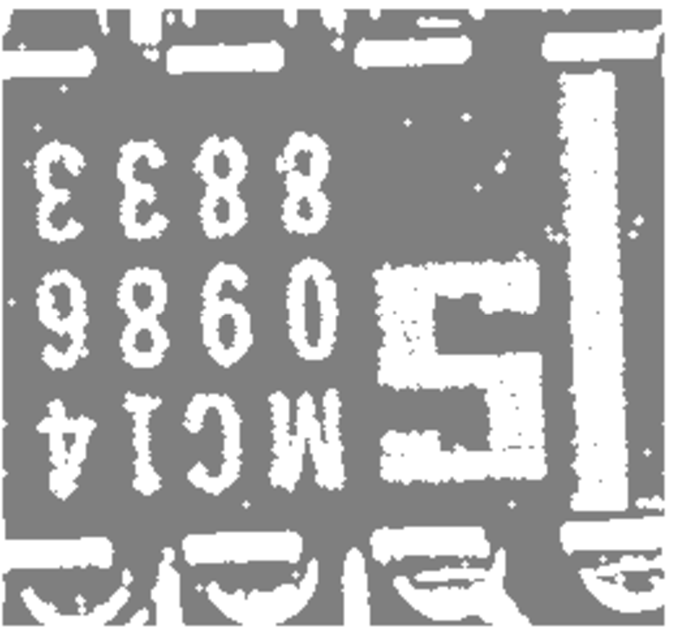,width=4cm} &
  \psfig{file=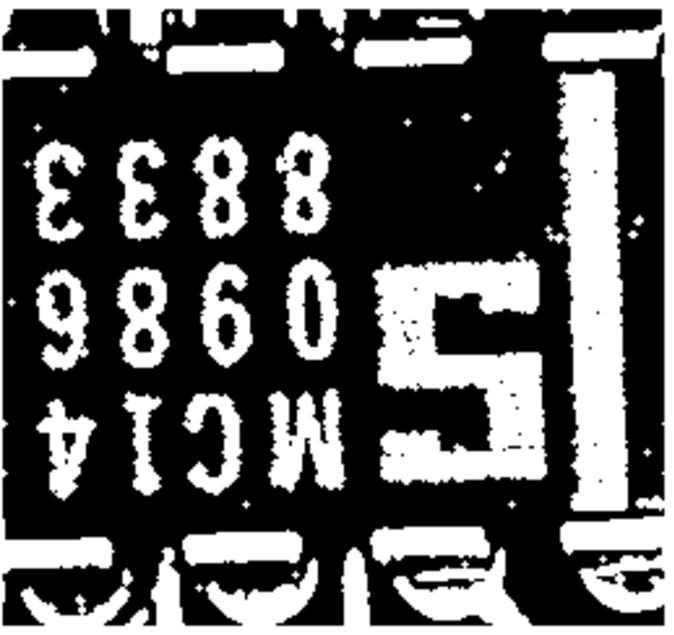,width=4cm} \vspace{2mm} \\  
  (d) output image ($t=0.5$) & (e) output image ($t=2$) &  (f) output image ($t=4$)   \vspace{5mm} \\
  \psfig{file=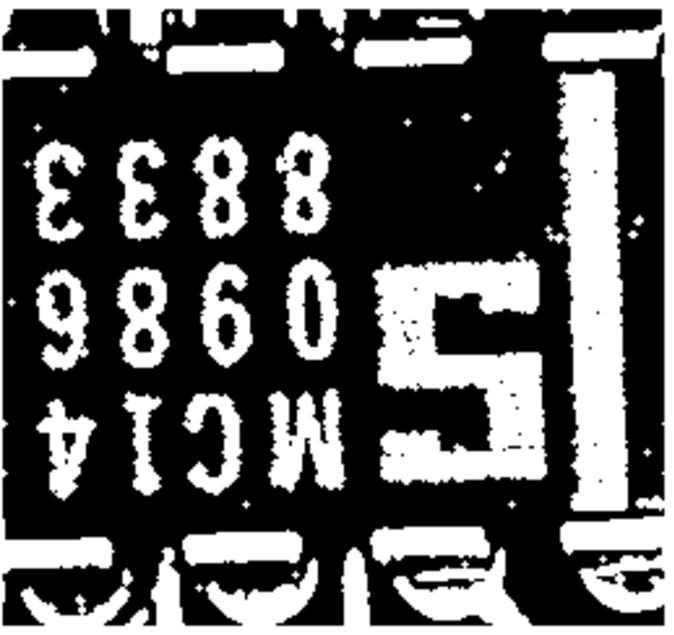,width=4cm} &
  \psfig{file=./figure/Text-2-t-500.eps,width=4cm} &  \vspace{2mm} \\
  (g) output image ($t=20$) & (h) output image ($t=50$) & 
  \end{tabular}
  \caption{Neuron-like behavior for the memoristor CNN (\ref{eqn: memristor-cnn1}) with the erosion template (\ref{eqn: template35}).   
   Observer that the memoristor CNN (\ref{eqn: memristor-cnn1}) cannot respond to the input quickly, 
   and it can hold the output image even if all memristors switch ``off''.  
   \newline
   (a) Binary input image for the memoristor CNN (\ref{eqn: memristor-cnn1}).  
       At this point, all memristors switch ``{\bf off}''.  The state $y_{ij}=0$ is shown in gray.   
   \newline  
   (b) The memoristor CNN (\ref{eqn: memristor-cnn1}) does \emph{not} peel off the boundary pixels, 
       since {\bf all memristors} still switch ``{\bf off}'', that is, the memristor can not switch ``on'' rapidly.   
   \newline
   (c)-(e) The boundary pixels are peeled off gradually,  
    since the memristors are turning to the ``{\bf switch-on}'' state from the ``switch-off'' state.      
   \newline
   (f) The color of the cells is changed to ``black and white'' from ``gray and white''.     
   \newline
   (g)-(h) The memoristor CNN (\ref{eqn: memristor-cnn1}) hold the output image,  
    even if all memristos switch ``{\bf off}'', again (see Figure \ref{fig:text-100}(g) and (h))}
 \label{fig:text-200}
\end{figure}
\clearpage

%
\section{Suspend and Resume Feature}
\label{sec: suspend}
%

The \emph{suspend} and \emph{resume feature} are useful when we want to save the current state, 
and continue work later from the same state.  
We show that the memristor CNN has the similar feature.

\begin{figure}[t]
 \begin{center} 
  \psfig{file=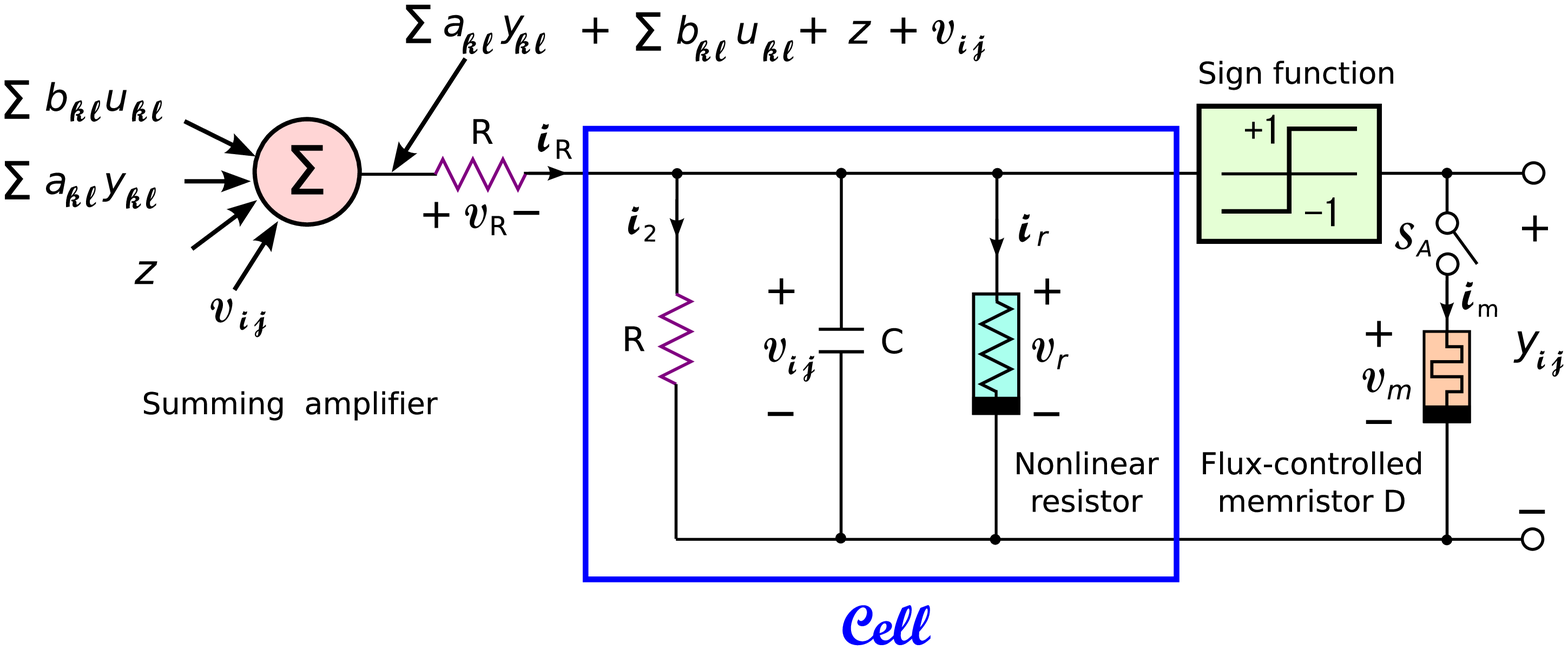,width=15cm}
  \caption{Circuit which can store the time average of the output state $y_{ij}$.    
   The switch $S_{A}$ is closed during the time averaging.   
   \newline   
   (1) Parameters: $R=1, \ C=1$.   
   \newline 
   (2) The $v-i$ characteristic of the nonlinear resistor (light blue) is given by 
   $i_{r} = f_{r}(v_{r}) \stackrel{\triangle}{=} - 0.5 \, a \, ( |v_{r}+ 1 | -  |v_{r} - 1 | )$, 
   where $a$ is a constant.
   \newline 
   (3) The terminal currents and voltages of the flux-controlled memristor $D$ (orange) satisfies 
   $i_{m} = W( \varphi_{m} ) \, v_{m} = \mathfrak{s}  [\, \varphi_{m} \,] \, v_{m} $.  
   The symbol $\mathfrak{s}  [\, z \,]$ denotes the \emph{unit step} function, 
   equal to $0$ for $z < 0$ and 1 for $z \ge 0$. 
   \newline 
   (4) The output voltage of the summing amplifier (pink) is given by  
   $\displaystyle \sum_{k, \, l \in N_{ij}, \  k \ne i, \ l \ne j} a_{kl} \, \operatorname {sgn} (v_{kl}) 
   + \sum_{k, \, l \in N_{ij}}b_{k l} \ u_{kl} + z + v_{ij}$.  
   \newline
   (5) The above output voltage contains the voltage $v_{ij}$ of the capacitor $C$ (the last term).      
   \newline
   (6) The above sum 
   $\displaystyle \sum_{k, \, l \in N_{ij}, \  k \ne i, \ l \ne j} a_{kl} \, \operatorname {sgn} (v_{kl}) $
   does \emph{not} contain the term 
   $ a_{ij} \, \operatorname {sgn} (v_{ij})$ \ ($k=i, \ l=j$).  
   \newline   
   (7) The output $y_{ij}$ and the state $v_{ij}$ of each cell are related via the sign function (green): 
    $y_{ij} = \operatorname {sgn} (v_{ij})$.  }
 \label{fig:store}
 \end{center}
\end{figure}

Let us consider the memristor CNN in Figure \ref{fig:store}.  
The parameters are given by 
\begin{equation}
   E=1, \, J= 0.5, \, R=1, \ C=1.  
\end{equation}
The characteristic of the nonlinear resistor is given by 
\begin{equation}
  i_{r} =  f(v_{r}) = - 0.5 \, a \, ( |v_{r}+ 1 | -  |v_{r} - 1 | ), 
\end{equation}
where $a$ is a constant.  
The constitutive relation and memductance of the flux-controlled memristor are given by   
\begin{equation}
 q_{m} = h(\varphi_{m}) \stackrel{\triangle}{=}  0.5(|\varphi_{m}| + \varphi_{m}), 
\label{eqn: qm-hm-40}
\end{equation}
and 
\begin{equation}
  W( \varphi_{m} ) = \frac{dh(\varphi_{m})}{d\varphi_{m}} = \mathfrak{s}  [\, \varphi_{m} \,],
\label{eqn: wm-40}
\end{equation}
respectively (see Figure \ref{fig:relation-40}), 
where $q_{m} $ and $\varphi_{m}$ denote the charge and the flux of the memristor, respectively, the symbol $\mathfrak{s}  [\, z \,]$ denotes the \emph{unit step} function, equal to $0$ for $z < 0$ and 1 for $z \ge 0$.

\begin{figure}[h]
 \centering
  \begin{tabular}{cc}
   \psfig{file=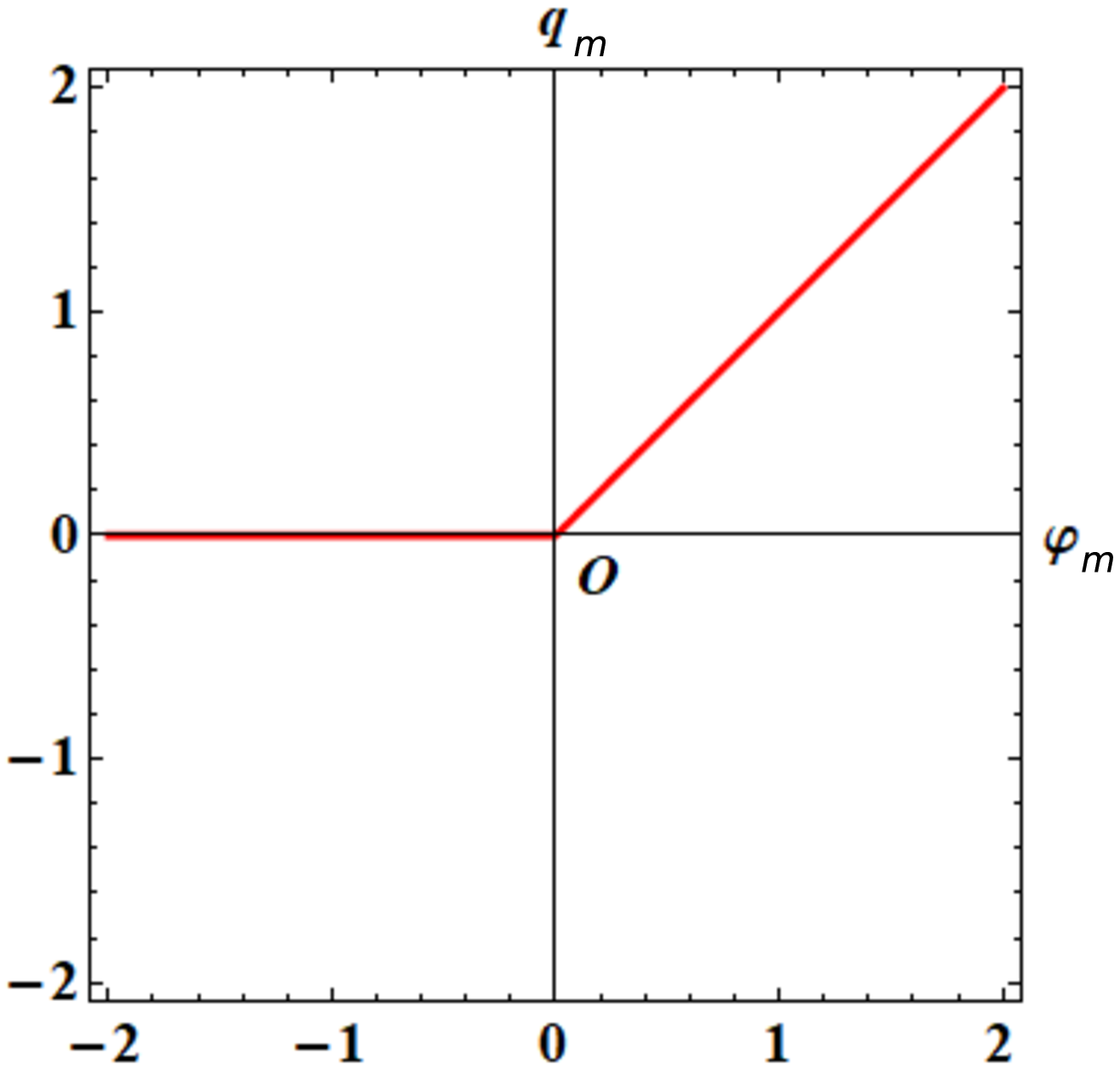, width=6.0cm} & 
   \psfig{file=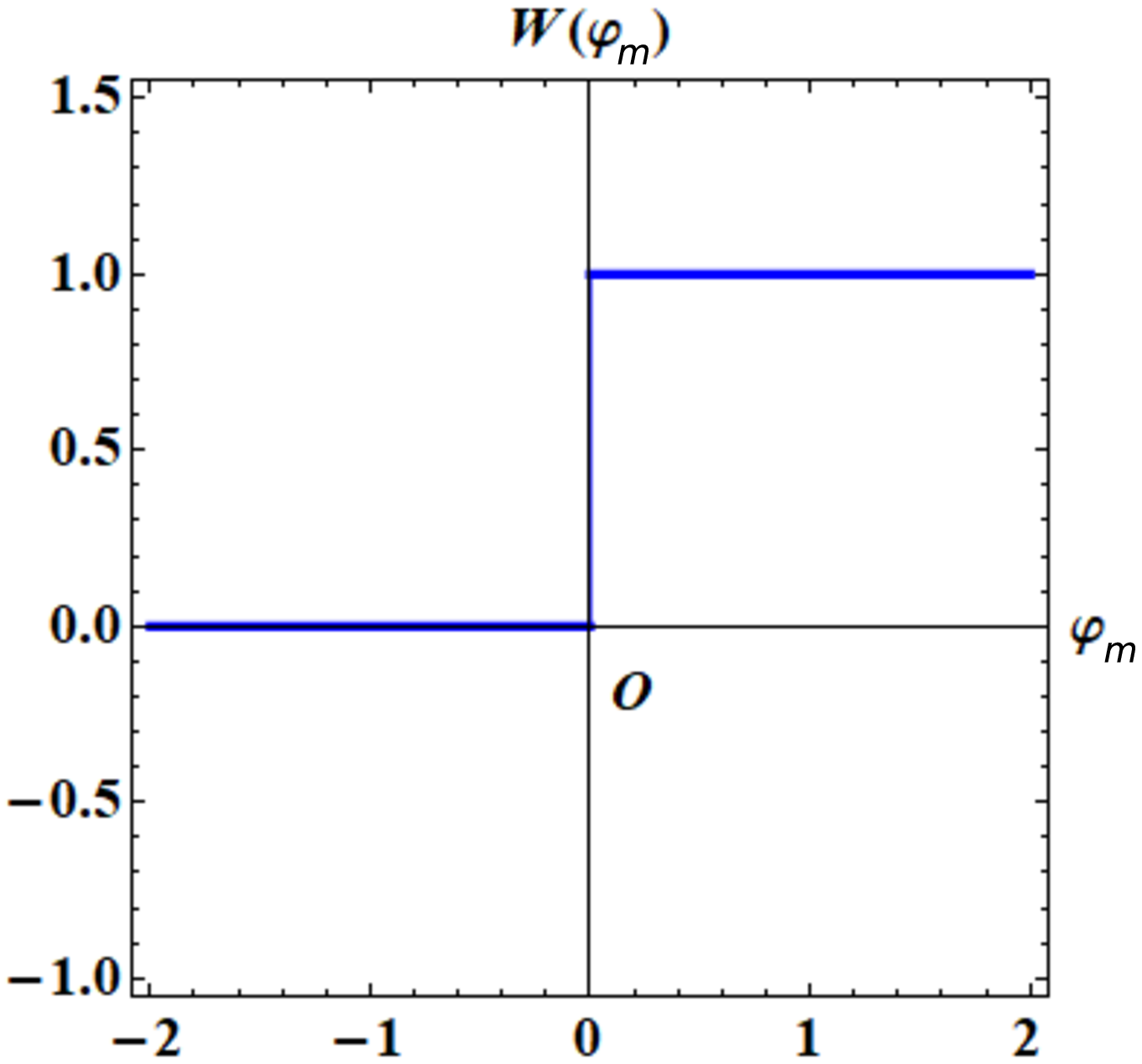, width=6.0cm} \vspace{1mm} \\
   (a) constitutive relation $q_{m} = h (\varphi_{m})$ & 
   (b) memductance  $W( \varphi_{m} )$ \\
  \end{tabular} 
  \caption{Constitutive relation and memductance of the flux-controlled memristor.  
   The memristor switches ``off'' and ``on'' depending on the value of the flux $\varphi_{m}$. 
   \newline 
   (a) The constitutive relation of the memristor, which is given by 
       $q_{m} = h(\varphi_{m}) \stackrel{\triangle}{=}  0.5(|\varphi_{m}| + \varphi_{m})$.   
   \newline
   (b) Memductance $W( \varphi )$ of the  memristor, which is defined by   
       $\displaystyle W( \varphi_{m} ) \stackrel{\triangle}{=} \frac{dh(\varphi_{m})}{d\varphi_{m}} 
        = \mathfrak{s}  [\, \varphi_{m} \,]$. 
       Here, the symbol $\mathfrak{s}  [\, z \,]$ denotes the \emph{unit step} function, 
       equal to $0$ for $z < 0$ and 1 for $z \ge 0$.   }
 \label{fig:relation-40}
\end{figure}
%
%

\begin{figure}[p]
 \begin{center} 
  \psfig{file=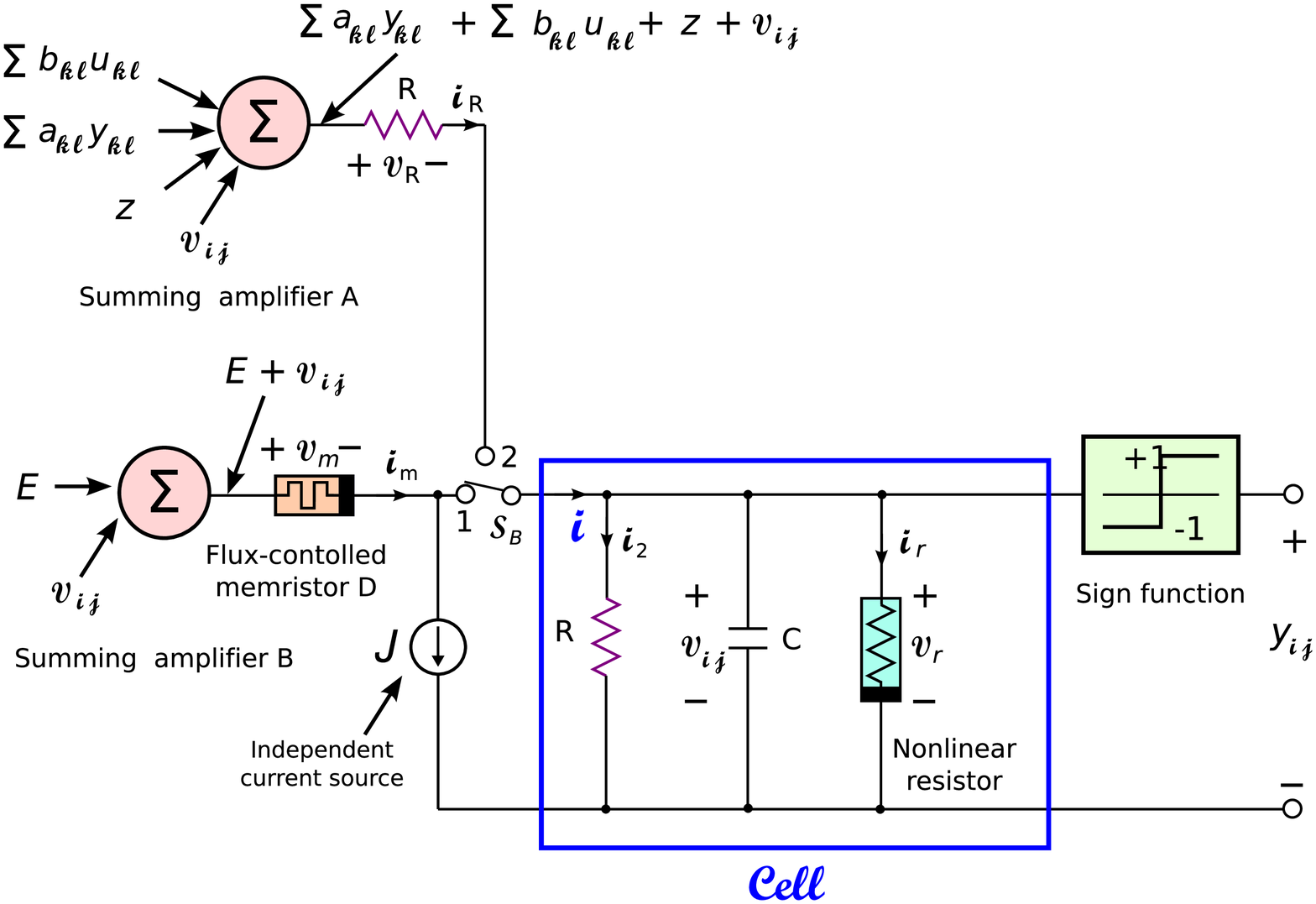,width=15cm}
  \caption{Circuit for recovering the previous average output, and for resuming the computation.  
   At first, the switch $S_{B}$ is set to the position $1$ in order to the recover the previous average output.   
   Then, the switch $S_{B}$ is set to the position $2$ in order to resume the computation.  
   \newline 
   (1) Parameters:  $E=1, \, J= 0.5, \, R=1, \ C=1$. 
   \newline 
   (2) The $v-i$ characteristic of the nonlinear resistor (light blue) is given by 
   $i_{r} = f_{r}(v_{r}) \stackrel{\triangle}{=} - 0.5 \, a \, ( |v_{r}+ 1 | -  |v_{r} - 1 | )$, 
   where $a$ is a constant.  
   \newline 
   (3) The terminal currents and voltages of the memristor $D$ satisfies 
   $i_{m} = W( \varphi_{m} ) \, v_{m} = \mathfrak{s}  [\, \varphi_{m} \,] \, v_{m} $. 
   \newline 
   (4) The output voltage of the summing amplifier $A$ is given by 
   $\displaystyle \sum_{k, \, l \in N_{ij}, \  k \ne i, \ l \ne j} a_{kl} \, \operatorname {sgn} (v_{kl}) 
   + \sum_{k, \, l \in N_{ij}}b_{k l} \ u_{kl} + z + v_{ij}$.  
   \newline  
   (5) The above output voltage contains the voltage $v_{ij}$ of the capacitor $C$ (the last term).     
   \newline   
   (6) The above sum 
   $\displaystyle \sum_{k, \, l \in N_{ij}, \  k \ne i, \ l \ne j} a_{kl} \, \operatorname {sgn} (v_{kl}) $
   does \emph{not} contain the term 
   $ a_{ij} \, \operatorname {sgn} (v_{ij})$, \ ($k=i, \ l=j$).
   \newline 
   (7) The output voltage of the summing amplifier $B$ is equal to $E+v_{ij}$. 
   \newline    
   (8) The output $y_{ij}$ and the state $v_{ij}$ of each cell are related via the sign function (green): 
    $y_{ij} = \operatorname {sgn} (v_{ij})$.  }
 \label{fig:recovery}
 \end{center}
\end{figure}

The memristor CNNs in Figures \ref{fig:store} and \ref{fig:recovery} work as follows:   
%
%
\begin{enumerate}
\item \underline{Normal procedure} \vspace*{2mm}\\
Turn ``off''  the switch $S_{A}$ in Figure \ref{fig:store}.   
In this case, the output $y_{ij}$ is \emph{not} connected to the memristor,   
and the memristor CNN circuit works normally.  
Te dynamics is given by Eq. (\ref{eqn: memristor-cnn1}).   \\ 

\item \underline{Stored procedure} \vspace*{2mm}\\
Turn ``on'' the switch $S_{A}$  in Figure \ref{fig:store} during the period from $t_{0}$ to $t_{0} + T$, where $0 < T \ll 1$.   
In this period, the output $y_{ij}$ is connected to the memristor.  
The stored flux is give by 
\begin{equation} 
  \varphi_{m} (T) = \int_{t_{0}}^{t_{0}+T} y_{ij}(t)(\tau) d\tau.    
\label{eqn: phi-40}
\end{equation} 
It is recast into the form 
\begin{equation}
  \frac{\varphi_{m} (T)}{T} = \frac{\displaystyle \int_{t_{0}}^{t_{0}+T} y_{ij}(t)(\tau) d\tau}{T}, 
\end{equation}
which is the \emph{time average} of the output $y_{ij}(t)$.  \\
Thus, $\varphi_{m} (T)$ is regarded as the time average of the output $y_{ij}(t)$ except for a scale factor of $T$.  \\

\item \underline{Suspend procedure} \vspace*{2mm}\\
Power off the memristor CNN in Figure \ref{fig:store} at $t = t_{0}+T$. 
Then the computing process is suspended.   
The ideal memristor can retrieve stored information even after power off, since it is a \emph{non-volatile} element. If the memristor is not ideal, for example, it has a parasitic capacitance as shown in Figure \ref{fig:para-c}, then after long time power off, the flux of the memristor may decay via a parasitic capacitance \cite{Itoh2016}.  We discuss its effect in Sec. \ref{sec: flux}. \\

\begin{figure}[t]
 \begin{center} 
  \psfig{file=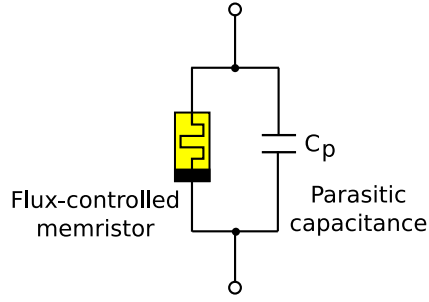,width=5.5cm}
  \caption{Parasitic capacitance $C_{p}$  of the flux-conrolled memristor.  }
 \label{fig:para-c}
 \end{center}
\end{figure}
%
%

\item \underline{Recovery procedure} \vspace*{2mm}\\
In order to continue the process from the previous state, 
we use the circuit in Figure \ref{fig:recovery}, where we use the memritor in Figure \ref{fig:store}.   

~~Let us first set the switch $S_{B}$ to the position $1$ at $t = t_{1}$. 
Then, the current $i$ through the cell is given by   
\begin{equation}
\begin{array}{lll}
  i(t) &=& W( \varphi_{m}(t) )E + J  \vspace{2mm} \\
  &=& \mathfrak{s}[\, \varphi_{m} (t)\,] - 0.5   \vspace{2mm} \\ 
  &=& \left \{ 
   \begin{array}{rlcc}
      0.5 & \ for \ & \ & \varphi_{m}(t) \ge 0,  \vspace{2mm} \\
     -0.5 & \ for \ & \ & \varphi_{m}(t) < 0,  
   \end{array}
   \right.  
\end{array}
\label{eqn: i-w}
\end{equation}
where $E=1, \, J= 0.5$,  
and $\varphi_{m}(t)$ is defined by 
\begin{equation}
\begin{array}{lll}
  \varphi_{m}(t) &=& \displaystyle  \int_{t_{1}}^{t} E  d\tau  + \varphi_{m} (T) \vspace{2mm} \\
                  &=& E \Delta t + \varphi_{m} (T) = \Delta t + \varphi_{m} (T) , 
\end{array}
\end{equation}
where $0 < \Delta t \stackrel{\triangle}{=} t - t_{1}$ and $E=1$. 

Assume that $0 < \Delta t< |\varphi_{m} (T)|$.     
Then we obtain 
\begin{equation}
  \operatorname {sgn} ( \varphi_{m}(t)) = \operatorname {sgn} ( \varphi_{m} (T)).   
\end{equation}
From Eq. (\ref{eqn: i-w}), we obtain 
\begin{equation}
\begin{array}{l}
  i(t) = 
   \left \{
   \begin{array}{rlcc}
      0.5 & \ for \ & \ & \varphi_{m}(T) \ge 0,  \vspace{2mm} \\
     -0.5 & \ for \ & \ & \varphi_{m}(T) < 0.    
   \end{array}
   \right.  
\end{array}
\label{eqn: i-t-50}
\end{equation}
%
%

\begin{figure}[ht]
 \centering
  \begin{tabular}{cc}
  \psfig{file=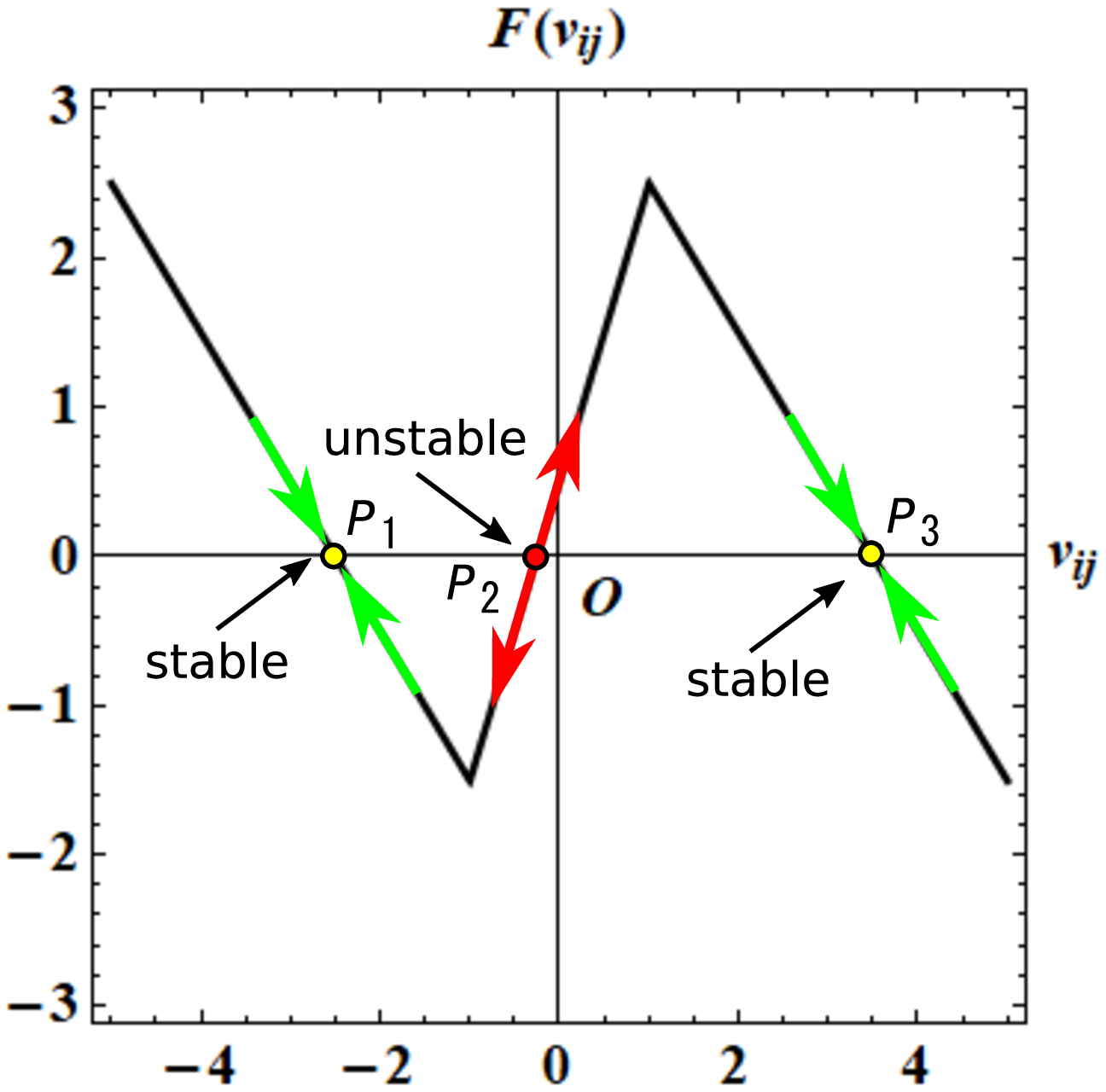,width=7.0cm} & 
  \psfig{file=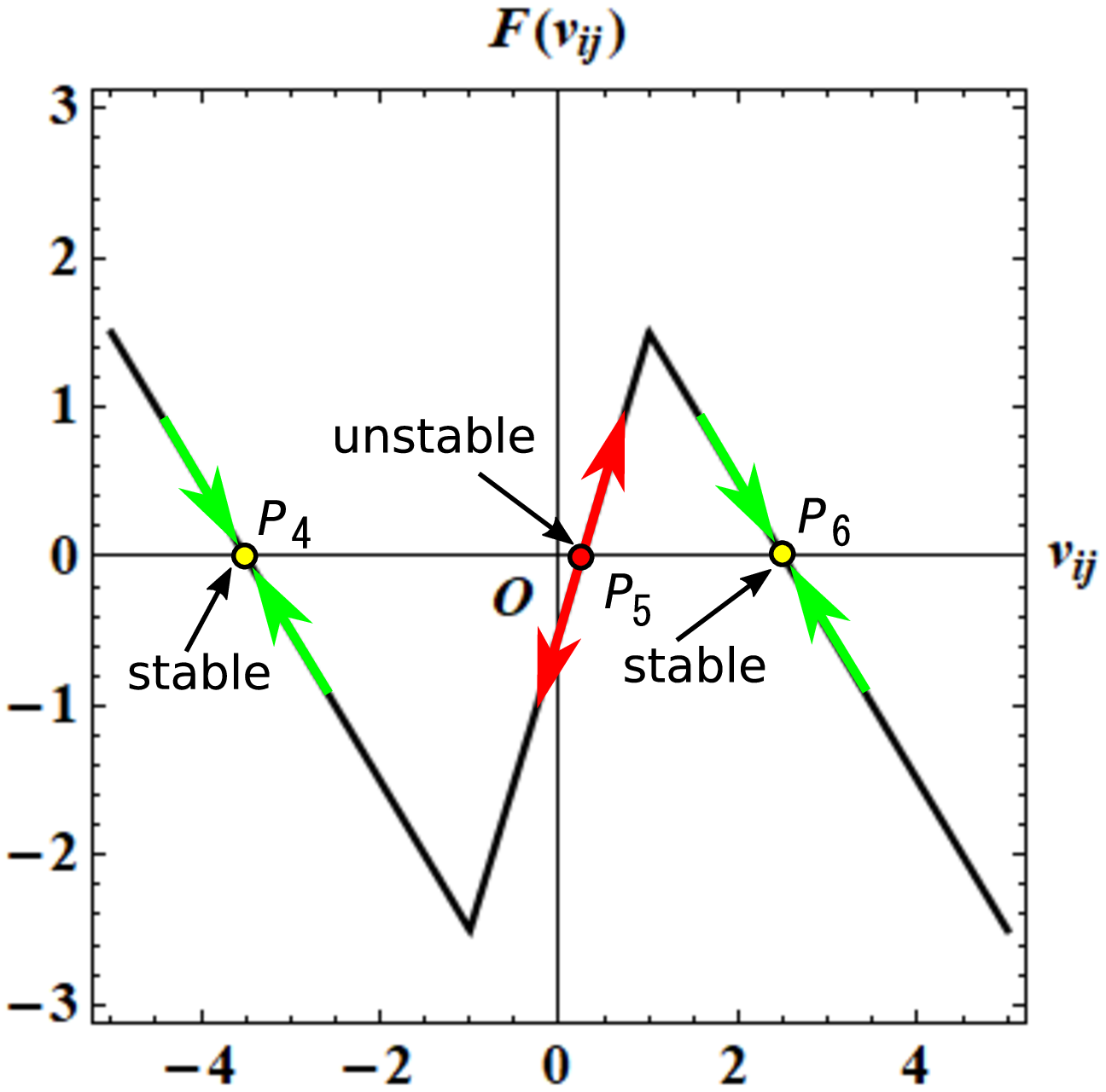,width=7.0cm} \vspace{1mm} \\
  (a) i= 0.5 & (b) i = -0.5  
  \end{tabular}   
  \caption{Driving-point plot of Eq. (\ref{eqn: isolated-cell-400}) with $a=3$.  
  Equation (\ref{eqn: isolated-cell-400}) has two stable equilibrium points (yellow), and one unstable equilibrium points (red). 
  If $i=0.5$, then $P_{1}=-2.5, \, P_{2}=-0.25, \, P_{3}=3.5$.  
  If $i=-0.5$, then $P_{4}=-3.5, \, P_{5}=0.25, \, P_{6}=2.5$.
  }
  \label{fig:stability-40}
\end{figure}

~~Consider next the dynamics of the cell, which is given by 
\begin{equation}
 \begin{array}{l}
 \displaystyle \frac{dv_{ij}}{dt} = F(v_{ij})  \vspace{2mm} \\
 ~\stackrel{\triangle}{=} - v_{ij} + 0.5 \, a \, ( |v_{ij} + 1 | -  |v_{ij}  - 1 | ) + i(t).
 \end{array}
\label{eqn: isolated-cell-400} 
\end{equation}
We show the driving-point plot of Eq. (\ref{eqn: isolated-cell-400}) in Figure \ref{fig:stability-40}. 
Let us consider the case where  $a=3 > 1$.  
Then, Eq. (\ref{eqn: isolated-cell-400}) has two stable equilibrium points, 
and one unstable equilibrium points as shown in Figure \ref{fig:stability-40}. 
Furthermore, if $i(t) = 0.5$, then the equilibrium points are given by 
\begin{equation}
  P_{1}=-2.5, \, P_{2}=-0.25, \, P_{3}=3.5,    
\end{equation}
where the points $P_{1}$ and $P_{3}$ are stable, and the point $P_{2}$ is unstable.  
If $i(t) = -0.5$, then the equilibrium points are given by 
\begin{equation}
  P_{4}=-3.5, \, P_{5}= 0.25, \, P_{6}=2.5, 
\end{equation}
where the points $P_{4}$ and $P_{6}$ are stable, and the point $P_{5}$ is unstable.  

~~The initial condition for Eq. (\ref{eqn: isolated-cell-400}) is given by $v_{ij} (t_{1}) = 0$, 
since the power of the system was turned off during the period of time from $t_{0}+T$ to $t_{1}$. 
From the driving-point plot in Figure \ref{fig:stability-40}, we obtain 
\begin{equation} 
   \left.
   \begin{array}{l}
     \text{~if~}  i(t) = 0.5 \text{~and~} v_{ij} (t_{1}) = 0,  \vspace{2mm} \\
     \text{~then~} v_{ij}(t) \to P_{3}=3.5  \text{~for~} t \to \infty.  \vspace{4mm} \\
     \text{~if~} i(t) = -0.5 \text{~and~} v_{ij} (t_{1}) = 0,    \vspace{2mm} \\
     \text{~then~} v_{ij}(t) \to P_{4}=-3.5  \text{~for~}  t \to \infty.  
   \end{array}
   \right \}
\label{eqn: yij-45}
\end{equation}
Here, the initial condition is given by $v_{ij} (t_{1}) = 0$.\footnote{Note that the power was turned off during the period of time from $t_{0}+T$ to $t_{1}$.  Thus, $v_{ij} (t_{1}) = 0$.} 
Since the solution $v_{ij}(t)$ of Eq. (\ref{eqn: isolated-cell-400}) for $t>t_{1}$ can \emph{not} move across the unstable equilibrium point,   
we obtain 
\begin{equation} 
   \left.
   \begin{array}{l}
     y_{ij}(t) = 1,  \text{~if~} i(t)= 0.5, \vspace{2mm} \\
     y_{ij}(t) = -1  \text{~if~} i(t)= -0.5,
   \end{array}
   \right \}
\label{eqn: yij-46}
\end{equation}
where $y_{ij}(t) = \operatorname {sgn} (v_{ij}(t))$. 
Furthermore, from Eqs. (\ref{eqn: i-t-50}) and (\ref{eqn: yij-46}), we obtain 
\begin{equation}
  y_{ij}(t) = \operatorname {sgn} (v_{ij}(t))
  = \left \{ 
   \begin{array}{rlcc}
      1 & \ if \ & \ & \varphi_{m}(T) \ge 0,  \vspace{2mm} \\
     -1 & \ if \ & \ & \varphi_{m}(T) < 0,   
   \end{array}
    \right.  
\label{eqn: yij-47}
\end{equation}
where $\varphi_{m}(T)$ denotes the stored flux during the period from $t_{0}$ to $t_{0} + T$, and it is the time average of the output $y_{ij}(t)$ except for a scale factor of $T$.
We have just recovered the previous average output from $\varphi_{m}(T)$.\footnote{We did not use $y_{ij}(t_{0}+T)$, but $\varphi_{m}(T)$, where the system is turned off at $t = t_{0}+T$. }   \\

\item \underline{Resume procedure} \vspace*{2mm}\\
Set the switch $S_{B}$ to the position $2$.  
Then, the image processing starts again from the previous average output state.  
That is, we can resume the computation.  \\
\end{enumerate}
%
%
We show next two examples of the suspend and resume feature.

%
%
\subsection{Hole-filling}
\label{sec: hole}
%
Let us consider the \emph{hole-filling template} \cite{{Chua1998},{Roska1997}}
\begin{equation}
 A =
 \begin{array}{|c|c|c|}
  \hline
   ~0~   &  ~1~   &  ~0~   \\
  \hline
   ~1~   &  ~3~   & ~1~   \\  
  \hline 
   ~0~   &  ~1~   &  ~0~  \\ 
  \hline
  \end{array} \ , \ \ \ 
 B = 
   \begin{array}{|c|c|c|}
  \hline
   ~0~ &  ~0~  &  ~0~ \\
  \hline
   ~0~ &  ~4~  &  ~0~   \\  
  \hline 
   ~0~ &  ~0~  &  ~0~    \\ 
  \hline
  \end{array} \ ,  \ \ \ 
 z =
  \begin{array}{|c|}
  \hline
    -1 \\
  \hline
  \end{array} \  . \vspace*{2mm}\\
\label{eqn: template40} 
\end{equation}
The initial condition for the state $v_{ij}$ is equal $1$, 
and the input $u_{kl}$ is equal to a given binary image.  
The boundary condition is given by 
\begin{equation}
  v_{k^{*}l^{*}}  = 0,  \  u_{k^{*}l^{*}}  = 0,
\end{equation}
where $k^{*}l^{*}$ denotes boundary cells. 
This template can fill the interior of all closed contours in a binary image as shown in Figure \ref{fig:output-7}.

\begin{figure}[h]
 \centering
  \begin{tabular}{cc}
  \psfig{file=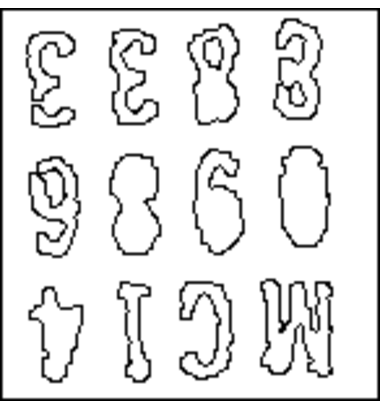,width=5cm} &
  \psfig{file=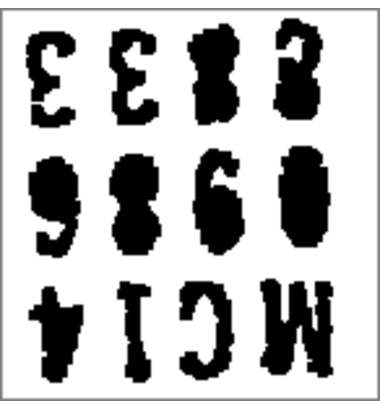,width=5cm} \vspace{1mm} \\
  (a) input image   & (b)   output image   \\ 
  \end{tabular}
  \caption{Input and output images for the hole-filling template (\ref{eqn: template40}).  
  This template can fill the interior of all closed contours in a binary input image (left).  
  Image size is $146 \times 151$.  }
 \label{fig:output-7}
\end{figure}

We show our detailed computer simulations in Figure \ref{fig:output-100}.   
Observe that even if we turn off the power of the memristor CNN during the computation, 
it can resume from the previous average output state.

\begin{figure}[p]
 \centering
  \begin{tabular}{ccc}
   \psfig{file=./figure/hole-filling-input.eps,width=4cm} &
   \psfig{file=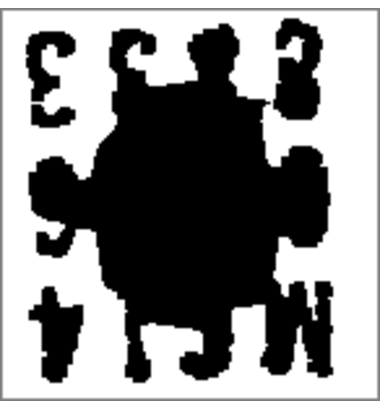,width=4cm} &
   \psfig{file=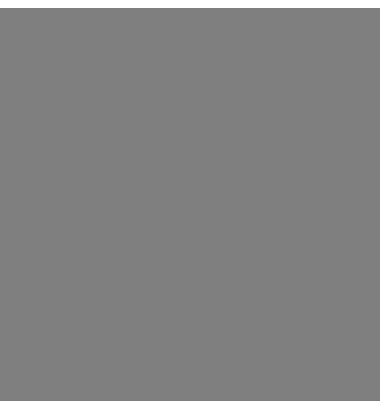,width=4cm} \vspace{2mm} \\
  (a) $t=0$  & (b) $t=69.8$  &  (c) $t = 70$  \\
    (input image) & (processing) & (power off) \vspace{5mm} \\
  \psfig{file=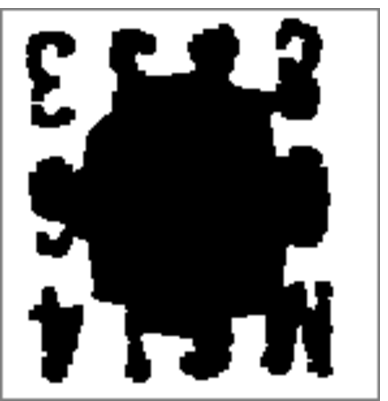,width=4cm} &  
  \psfig{file=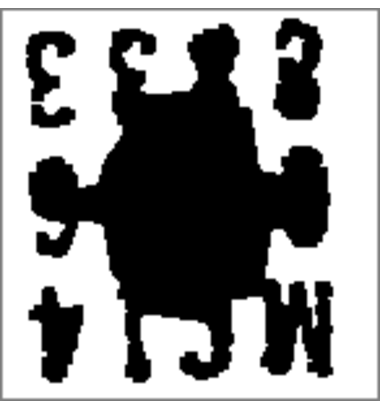,width=4cm} &
  \psfig{file=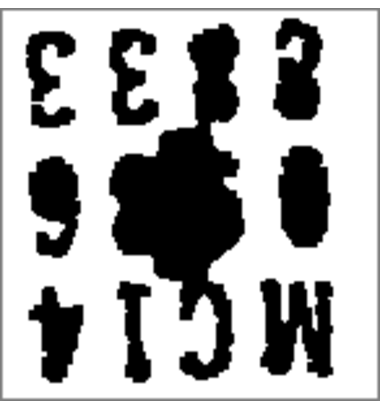,width=4cm} \vspace{2mm} \\  
  (d) $t=128$  & (e) $t=140$ &  (f) $t=160$  \\ 
   (recovered) &(resumed from $t=130$) &  (processing) \vspace{5mm} \\ 
  \psfig{file=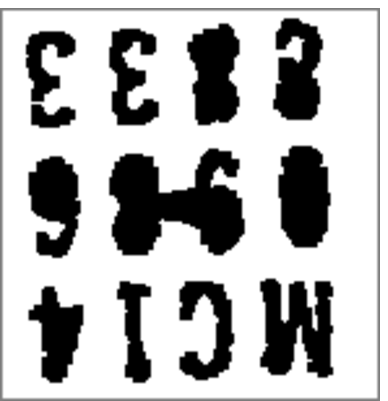,width=4cm}  &
  \psfig{file=./figure/hole-filling-1700.eps,width=4cm} &  \vspace{2mm} \\ 
  (g) $t=180$  & (h) $t=340$  &  \\
   (processing) & (processing) &  \vspace{5mm} \\
  \end{tabular}
  \caption{Suspend and resume feature for the hole-filling template (\ref{eqn: template40}).  
   \newline
   (a) Input image (image size is $146 \times 151$).  
   \newline  
   (b) Processing image at $t=69.8$.  
       The switch $S_{A}$ in Figure \ref{fig:store} is closed during the period  $t \in [60, \ 70)$.  
   \newline
   (c) Power off during the period $t \in [70, \ 120)$.  In this period, the cell output $y_{ij}=0$ (gray in pseudo-color).   
   \newline
   (d) Recovered image at $t=128$.  The previous \emph{average} output is recovered 
       by setting the switch $S_{B}$ in Figure \ref{fig:recovery} to the position $1$.    
       The recovering period is $t \in [120, \ 130)$.   
   \newline
   (e)-(h) Processed images at time $t=140, \, 160, \, 180, \, 340$, which are resumed from $t = 130$.  
        In this period, the switch $S_{B}$ in Figure \ref{fig:recovery} is set to the position $2$.  
        Observer that the memristor CNN can fill the interior of all closed contours in the binary input image, 
        even if we power off the circuit in the middle of image processing.                   
  }
 \label{fig:output-100}
\end{figure}

\clearpage
%
\subsection{Half-toning}
\label{sec: half}
%
Let us consider the \emph{half-toning template} \cite{{Chua1998},{Roska1997}}
\begin{equation}
\begin{array}{l}
 A =
 \begin{array}{|c|c|c|}
  \hline
   -0.07   &  -0.1~    &  -0.07   \\
  \hline
   -0.1~  &  1.15       & -0.1~   \\  
  \hline 
   -0.07   &  -0.1~    & -0.07  \\ 
  \hline
  \end{array} \ , \vspace{5mm} \\
 B = 
   \begin{array}{|c|c|c|}
  \hline
   ~0.07   &   ~~0.1~~~    &  ~0.07   \\
  \hline
  ~~0.1~~  &  0.32         &  ~~0.1~~   \\  
  \hline 
   ~0.07   &   ~~0.1~~~    &  ~0.07  \\ 
  \hline
  \end{array} \ ,  \vspace{5mm} \\
 z =
  \begin{array}{|c|}
  \hline
    0 \\
  \hline
  \end{array} \  .
\end{array} 
\label{eqn: template50} 
\end{equation}
The initial condition for the state $v_{ij}$ is equal to a given gray-scale image, 
and the input $u_{kl}$ is also equal to a given gray-scale image.  
The boundary condition is given by 
\begin{equation}
  v_{k^{*}l^{*}}  = 0,  \  u_{k^{*}l^{*}}  = 0,
\end{equation}
where $k^{*}l^{*}$ denotes boundary cells. 
This template can transform a gray-scale image into a ``half-tone'' binary image.  
The binary image preserves the main features of the gray-scale image as shown in Figure \ref{fig:output-200}.  

\begin{figure}[ht]
 \centering
  \begin{tabular}{cc}
  \psfig{file=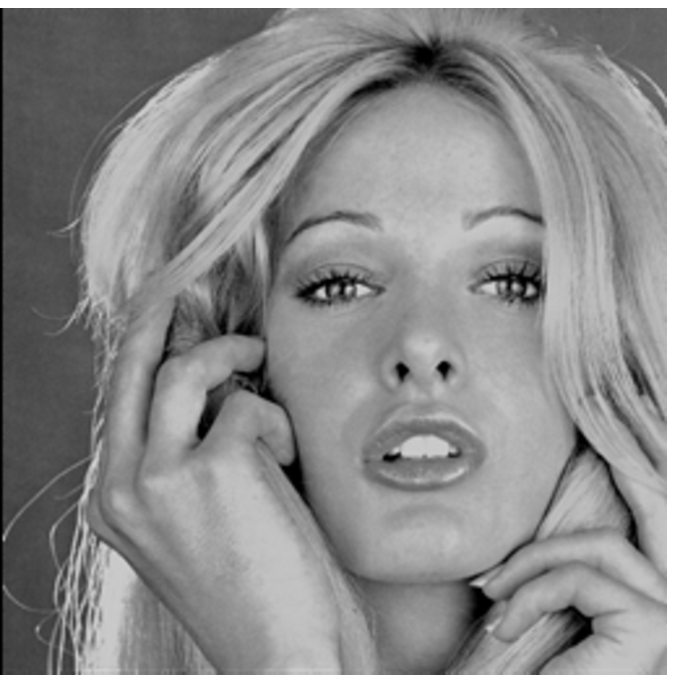,width=5cm} &
  \psfig{file=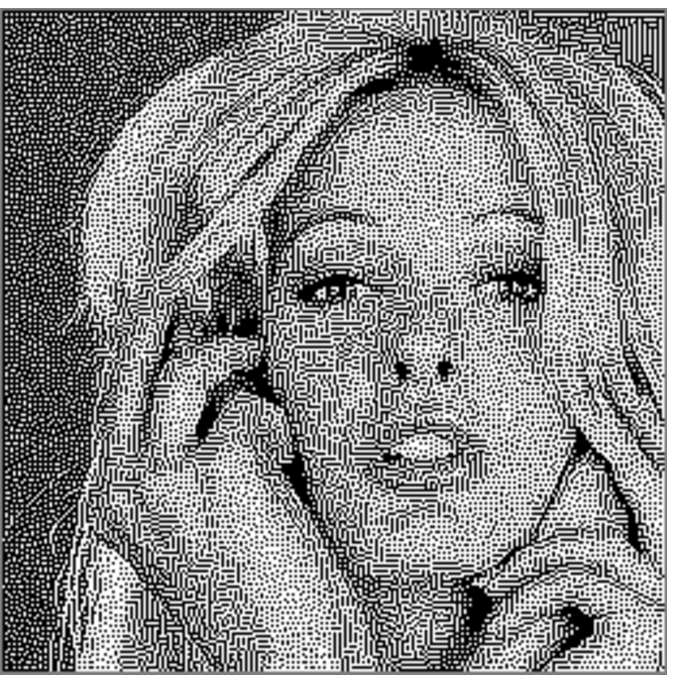,width=5cm} \vspace{1mm} \\
  (a) input image   & (b)   output image   \\ 
  \end{tabular}
  \caption{Input and output images for the half-toning template (\ref{eqn: template40}).  
  This template can transform a gray-scale input image (left) into a ``half-tone'' binary image (right).  
  The binary image preserves the main features of the gray-scale image. 
  Image size is $256 \times 256$. }
 \label{fig:output-200}
\end{figure}
%
%

\begin{figure}[ht]
 \centering
  \begin{tabular}{cc}
  \psfig{file=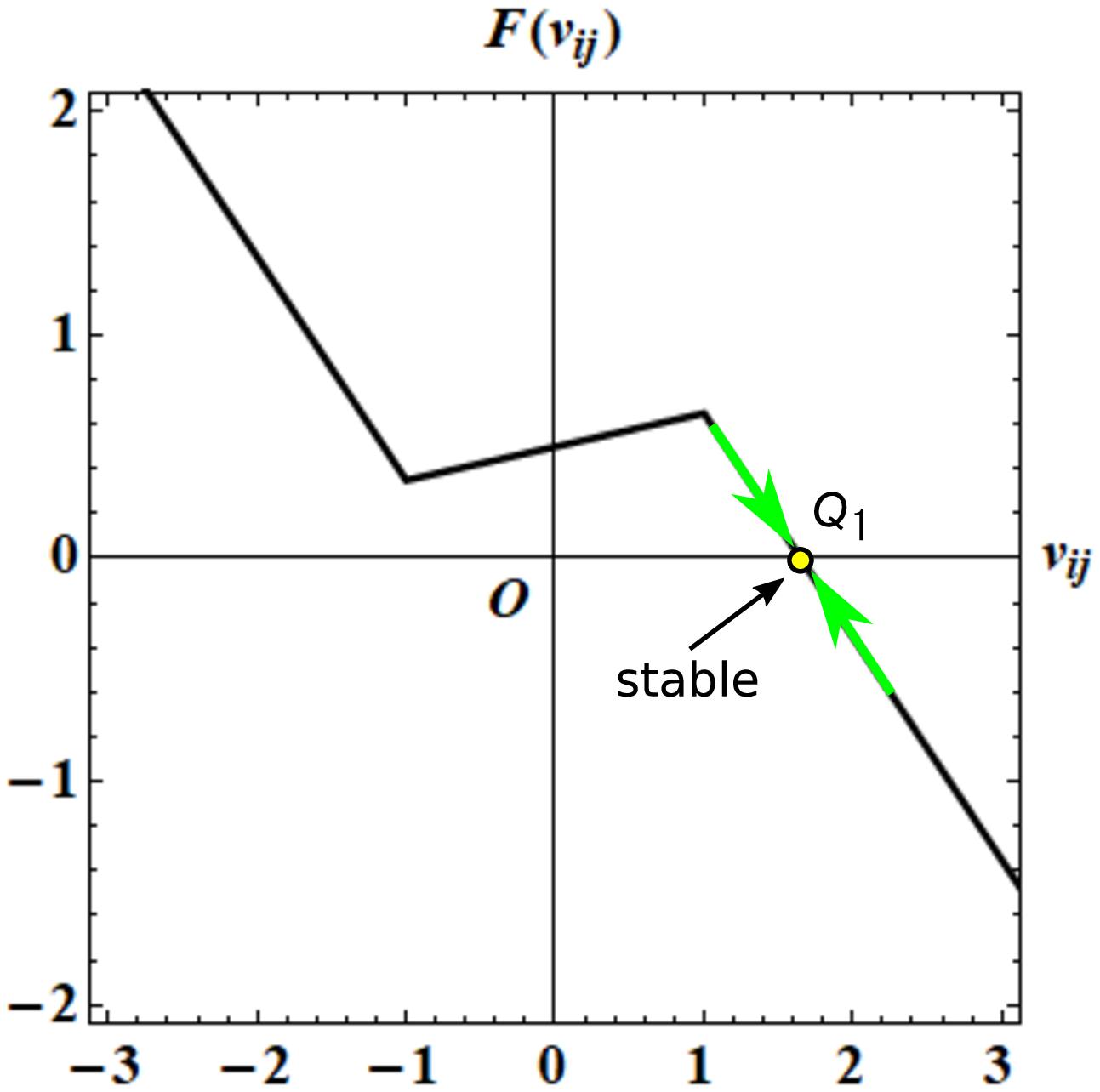,width=7.0cm} & 
  \psfig{file=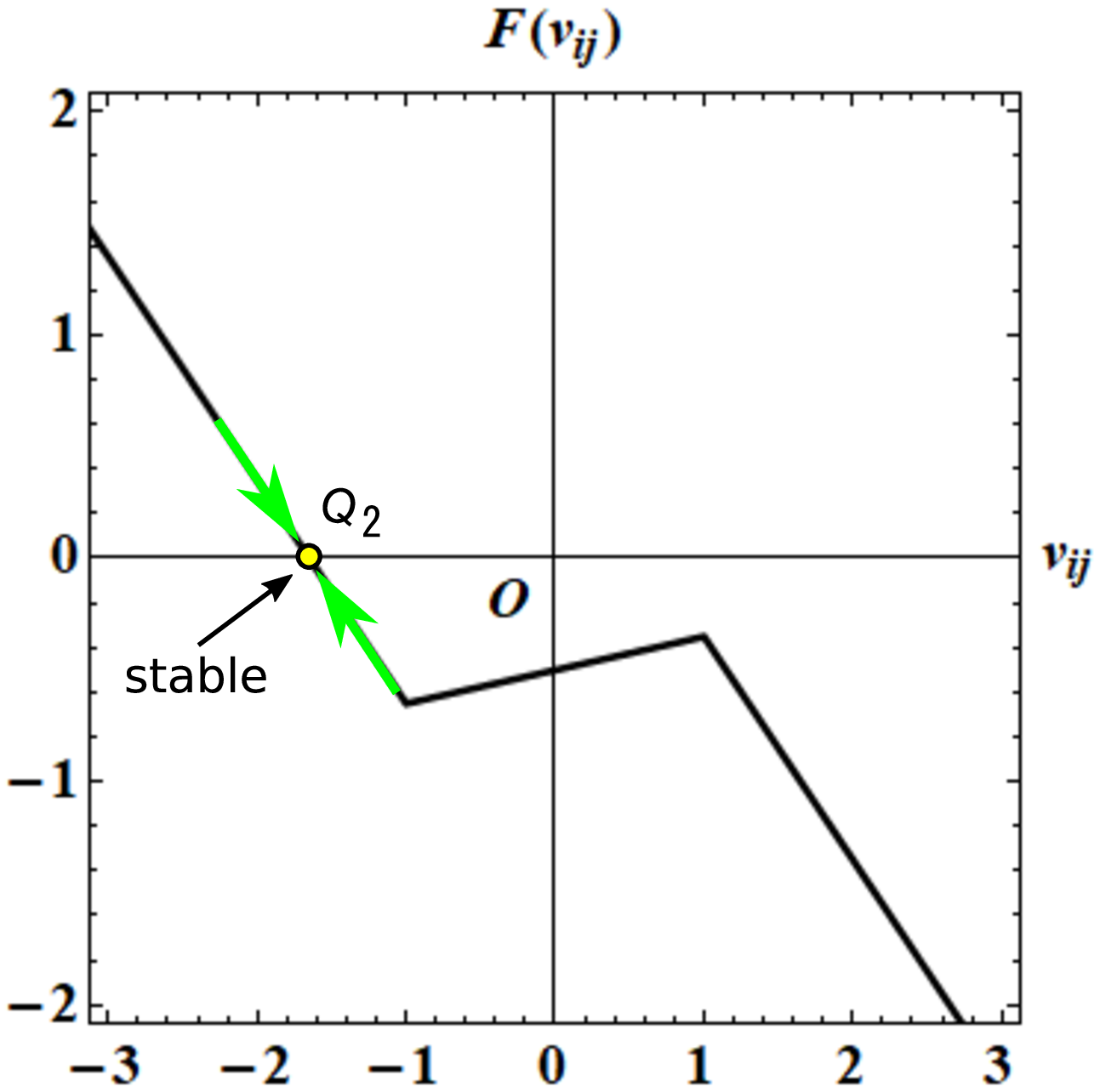,width=7.0cm} \vspace{1mm} \\
  (a) i= 0.5 & (b) i = -0.5  
  \end{tabular}   
  \caption{Driving-point plot of Eq. (\ref{eqn: isolated-cell-400}) with $a=1.15$. 
  Equation (\ref{eqn: isolated-cell-400}) has only one stable equilibrium points (yellow).   
  Furthermore, if $i=0.5$, then the stable equilibrium point is given by $Q_{1}=1.65$.  
  If $i=-0.5$, then the stable equilibrium point is given by $Q_{2}=-1.65$. 
  }
  \label{fig:stability-60}
\end{figure}

Since $a = a_{0, 0} = 1.15$, Eq. (\ref{eqn: isolated-cell-400}) has only one stable equilibrium points as shown in Figure \ref{fig:stability-60}.  
If $i(t) = 0.5$, then the equilibrium points are given by 
\begin{equation}
  Q_{1}=1.65.      
\end{equation}
If $i(t) = -0.5$, then the equilibrium points are given by 
\begin{equation}
  Q_{2}=-1.65.
\end{equation}
Thus, we obtain 
\begin{equation} 
   \left.
   \begin{array}{l}
     \text{~if~}  i(t) = 0.5 \text{~and~} v_{ij} (t_{1}) = 0,  \vspace{2mm} \\
     \text{~then~} v_{ij}(t) \to Q_{1}=1.65  \text{~for~} t \to \infty.  \vspace{4mm} \\
     \text{~if~} i(t) = -0.5 \text{~and~} v_{ij} (t_{1}) = 0,    \vspace{2mm} \\
     \text{~then~} v_{ij}(t) \to Q_{2}=-1.65   \text{~for~}  t \to \infty.  
   \end{array}
   \right \}
\label{eqn: yij-145}
\end{equation}
Here, the initial condition is given by $v_{ij} (t_{1}) = 0$.\footnote{Note that the power was turned off during the period of time from $t_{0}+T$ to $t_{1}$.  Thus, $v_{ij} (t_{1}) = 0$.}  
Since $y_{ij}(t) = \operatorname {sgn} (v_{ij}(t))$, we obtain 
\begin{equation} 
   \left.
   \begin{array}{l}
     y_{ij}(t) = 1,  \text{~if~} i(t)= 0.5, \vspace{2mm} \\
     y_{ij}(t) = -1  \text{~if~} i(t)= -0.5.  
   \end{array}
   \right \}
\label{eqn: yij-146}
\end{equation}
Furthermore, from Eqs. (\ref{eqn: i-t-50}) and (\ref{eqn: yij-46}), we obtain 
\begin{equation}
  y_{ij}(t) = \operatorname {sgn} (v_{ij}(t))
  = \left \{ 
   \begin{array}{rlcc}
      1 & \ if \ & \ & \varphi_{m}(T) \ge 0,  \vspace{2mm} \\
     -1 & \ if \ & \ & \varphi_{m}(T) < 0,   
   \end{array}
    \right.  
\label{eqn: yij-147}
\end{equation}
where $\varphi_{m}(T)$ denote the stored flux during the period from $t_{0}$ to $t_{0} + T$,  
and $\varphi_{m}(T)$ is regarded as the time average of the output $y_{ij}(t)$.
Thus, we obtained the same result as that for $a=3$ (see Eq. (\ref{eqn: yij-47})).

We show next our detailed computer simulations in Figure \ref{fig:output-300}.   
Observe that even if we turn off the power of the memristor CNN during the computation, 
it can resume from the previous average output state. 
Thus, we obtain the following result: 
%
%
\begin{center}
\begin{minipage}{12cm}
\begin{shadebox}
Assume $a_{00} > 1$.  
Then the memristor CNN has the suspend and resume feature, if we use the circuits in Figures \ref{fig:store} and \ref{fig:recovery}.  
\end{shadebox}
\end{minipage}
\end{center}
%
%

\begin{figure}[th]
 \centering
  \begin{tabular}{ccc}
   \psfig{file=./figure/WOMAN-2-input.eps,width=4cm} &
   \psfig{file=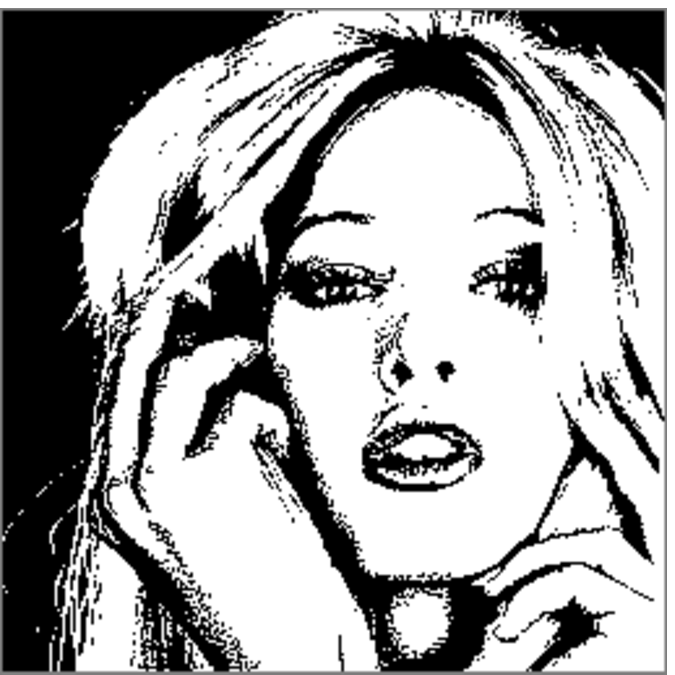,width=4cm} &
   \psfig{file=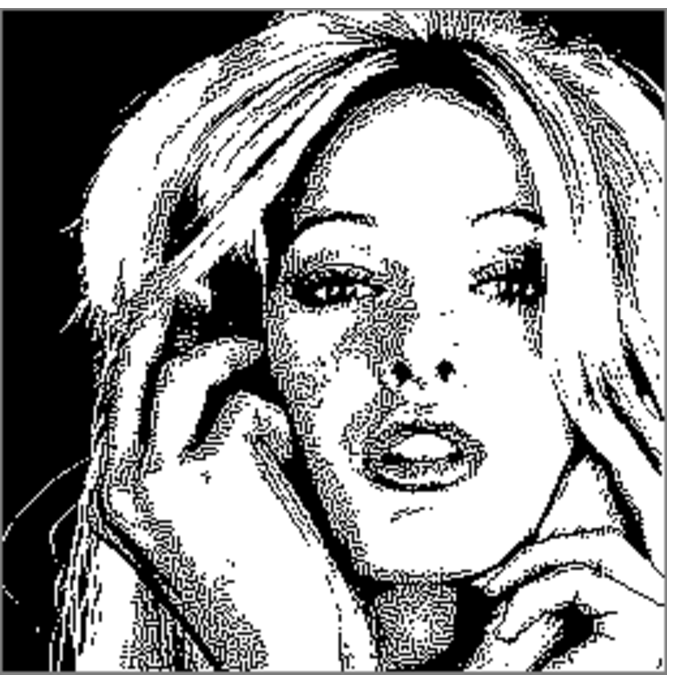,width=4cm} \vspace{2mm} \\
  (a) $t=0$  & (b) $t=0.1$  &  (c) $t = 0.34$  \\
    (input image) & (processing) & (processing)  \vspace{5mm} \\
  \psfig{file=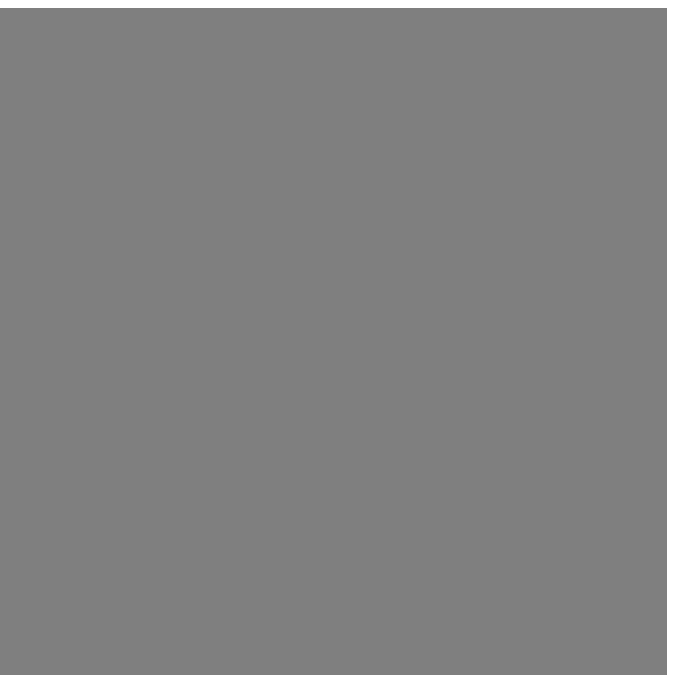,width=4cm}  &
  \psfig{file=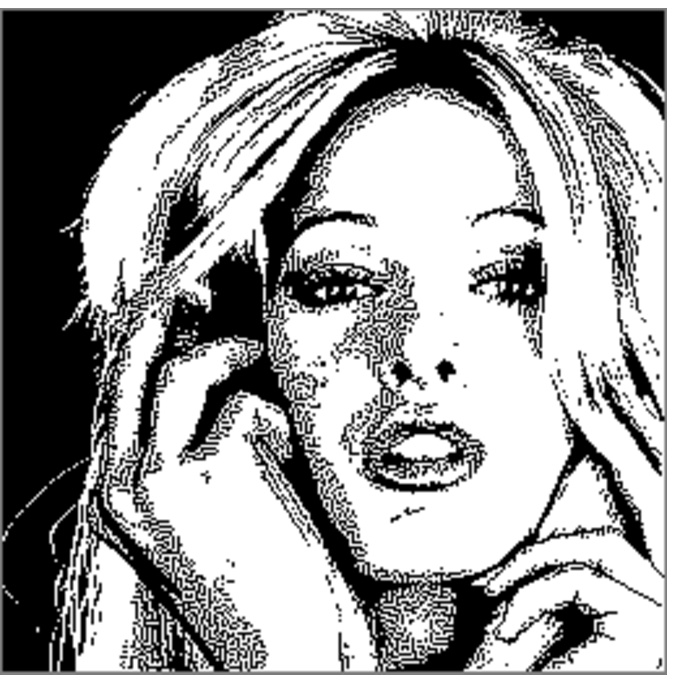,width=4cm} &  
  \psfig{file=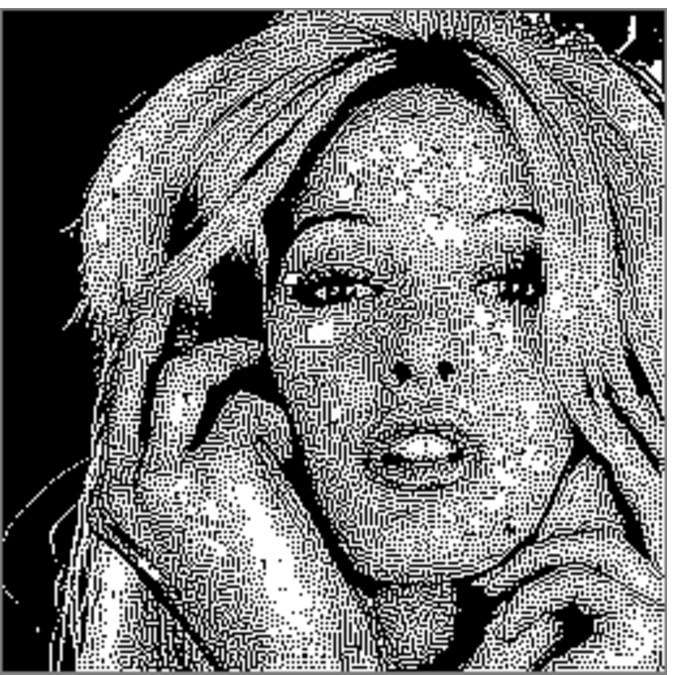,width=4cm}  \vspace{2mm} \\ 
  (d) $t=0.35$  & (e) $t=1.09$ &  (f) $t=1.2$  \\ 
   (power off) & (recovered) & (resumed from $t=1.1$)   \vspace{5mm} \\ 
  \psfig{file=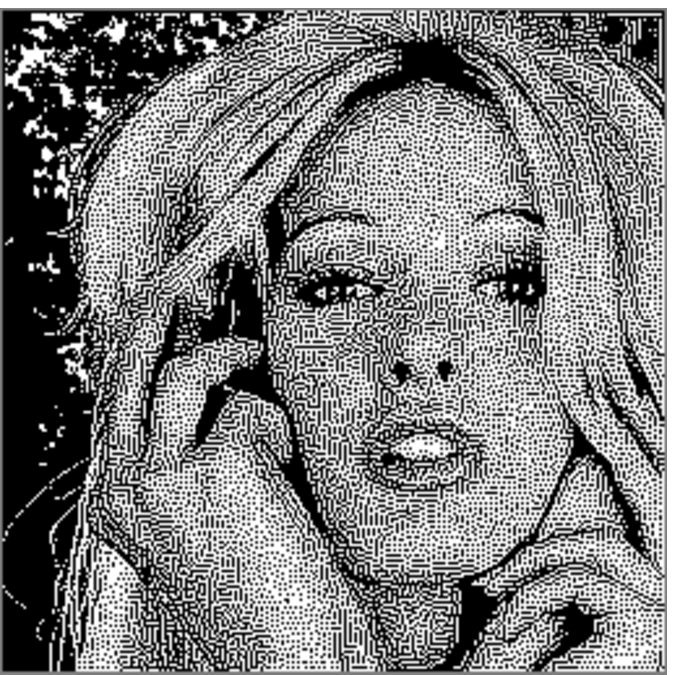,width=4cm} &
  \psfig{file=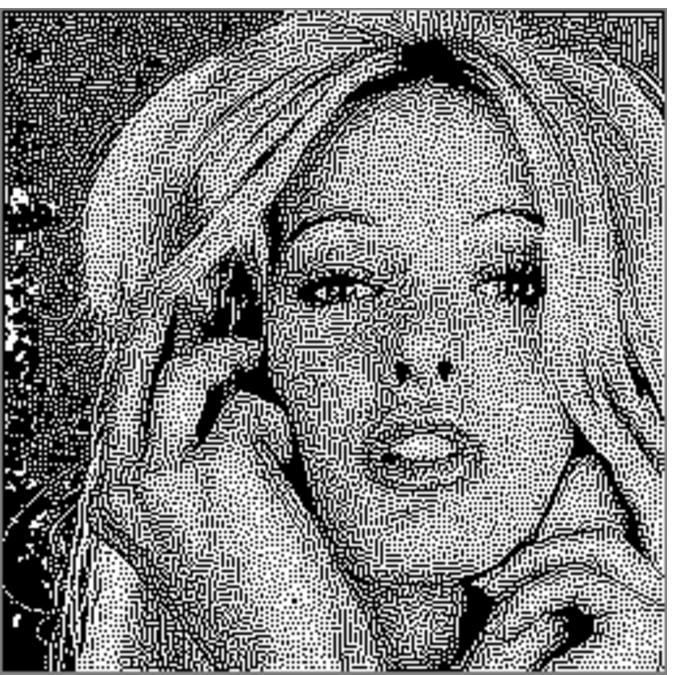,width=4cm} &
  \psfig{file=./figure/WOMAN-2-500.eps,width=4cm}  \vspace{2mm} \\ 
  (g) $t=1.25$   & (h) $t=1.3$ &  (i) $t=5$  \\
   (processing) & (processing) & (processing)  \vspace{5mm} \\
  \end{tabular}  
  \caption{Suspend and resume feature for the half-toning template (\ref{eqn: template50}).     
   \newline
   (a) Input image (image size is $256 \times 256$).
   \newline  
   (b) Processing image at $t=0.1$.  
   \newline  
   (c) Processing image at $t=0.34$. 
       Note that the switch $S_{A}$ in Figure \ref{fig:store} is closed during the period $t \in [0.35, \  0.6)$.      
   \newline
   (c) Power off during the period $t \in [0.35, \  1)$.  
       In this period, the cell output $y_{ij}=0$ (gray in pseudo-color).   
   \newline
   (d) Recovered image at $t=1.09$.  The previous average output is recovered 
       by setting the switch $S_{B}$ in Figure \ref{fig:recovery} to the position $1$.    
       The recovering period is $t \in [1, \ 1.1)$.   
   \newline
   (f)-(i) Processed images at time $t=1.2, \, 1.25,\, 1.3, \, 2, \, 5$, which are resumed from time $t = 1.1$.  
           In this period, the switch $S_{B}$ in Figure \ref{fig:recovery} is set to the position $2$.  
           Observer that the memristor CNN can transform the recovered image into the half-tone binary image, 
           even if we power off the circuit in the middle of image processing.                
  }
 \label{fig:output-300}
\end{figure}
\clearpage

%
\subsection{Long-term and short-term memories}
\label{sec: memory}
%
In our brain's system, a \emph{long-term memory} is a storage system for storing and retrieving information.  
A \emph{short-term memory} is the \emph{short-time} storage system that keeps something in mind before transferring it to a long-term memory. 

Consider the circuit in Figure  \ref{fig:store}. 
Thus, if the switch $S_{A}$ in Figure \ref{fig:store} is ``off'', then the output is hold only when the power is turned on.  
In this case, the circuit is regarded as a \emph{short-term memory circuit}.  
If the switch $S_{A}$ is ``on'', then the time average of the output is stored in the memristor, and we can retrieve it, even if the power is turned off.  
That is, the ideal memristor can retrieve stored information even after power off, since it is a \emph{non-volatile} element. 
In this case, the circuit becomes a \emph{long-term memory circuit}.   
Thus, we conclude as follow: 
%
%
\begin{center}
\begin{minipage}{12cm}
\begin{shadebox}
The memristor CNN has functions of the \emph{short-term} and \emph{long-term} memories, if we use the circuits in Figure \ref{fig:store}.
\end{shadebox}
\end{minipage}
\end{center}
%
%
%
\section{Effect of Flux Decay of Memristors}
\label{sec: flux}
%
A certain degree of decay of the flux or the charge is inevitable in physical devices.  
For example, the flux of the memristor may decay via a parasitic capacitance after long time power off \cite{Itoh2016}. 

Let us assume that the flux $\varphi_{m}$ decays to small value $\epsilon$ without change of sign.  
Then, the memductance $W (\varphi_{m})$ satisfies 
\begin{equation}
  W( \varphi_{m} ) \Bigr |_{\varphi_{m} = \epsilon} = W( \epsilon ) 
   =  \mathfrak{s}  [\, \epsilon \,] =     
   \left \{
   \begin{array}{rlcc}
      1 & \ if \ & \ &   \epsilon >0,  \vspace{2mm} \\
      0 & \ if \ & \ &   \epsilon <0,    
    \end{array}
   \right.
\end{equation}
where $ 0< | \epsilon | \ll 1$.
The current $i$ through the cell in Figure \ref{fig:recovery} is given by   
\begin{equation}
 \begin{array}{lll}
  i(t) &=& W( \epsilon )E + J  \vspace{2mm} \\
  &=& \mathfrak{s}[\, \epsilon \,] - 0.5   \vspace{2mm} \\ 
  &=& \left \{
   \begin{array}{rlcc}
      0.5 & \ if \ & \ &   \epsilon >0,  \vspace{2mm} \\
     -0.5 & \ if \ & \ &   \epsilon <0.    
    \end{array}
   \right.
 \end{array}
\label{eqn: i-w-2}
\end{equation}
where $E=1, \, J= 0.5$.  
From Eq. (\ref{eqn: yij-46}), we obtain 
\begin{equation}
 y_{ij}(t) = 
   \left \{
   \begin{array}{rlcc}
      1 & \ if \ & \ &   \epsilon >0,  \vspace{2mm} \\
     -1 & \ if \ & \ &  -\epsilon <0.    
    \end{array} 
    \right.
\end{equation}
Thus, the output of the cell does not change even if the flux decays to small value.   
Note that the output $y_{ij}(t)$ depends on only the sign of $\epsilon$, but it does not depend on its value.    
Thus, we can choose that $| \epsilon | = 0.001$ for the computer simulation.  

We show our computer simulations in Figures \ref{fig:output-2000} and \ref{fig:output-3000}, where we assume that the flux of the $50$ percent of memristors decays to small value without change of sign after power off.   
Observe that the memristor CNN defined by Figures \ref{fig:store} and \ref{fig:recovery} works well.  

The interesting thing is that if the flux of only the ``$1$ percent'' of memristors decay to small value \underline{with}  \emph{change of sign}, 
then the hole-filling template does \emph{not} work, as shown in Figure \ref{fig:output-4000}.  
Thus, the suspend and resume feature does not work if the change of sign occurs by memristor failures or parasitic elements.  
In this case, some error-correcting mechanisms should be used.  
However, the half-toning template works well even if the flux of ``all'' memristors decay to small value \underline{with} \emph{change of sign}, as shown in Figure \ref{fig:output-5000}.  
In this case, the output image for the half-toning template does not depend on the initial condition.  
Thus, we conclude as follow: 
%
%
\begin{center}
\begin{minipage}{12cm}
\begin{shadebox}
Assume $a_{00} > 1$.  
Then the memristor CNN defined by Figures \ref{fig:store} and \ref{fig:recovery} has the suspend and resume feature, even if the flux of the memristors decays to a small (but not too small) value, without change of sign, after power off.   
\end{shadebox}
\end{minipage}
\vspace{10mm}
\end{center}
%
%

\begin{figure}[ht]
 \centering
  \begin{tabular}{ccc}
  \psfig{file=./figure/hole-filling-input.eps,width=5cm} & 
  \psfig{file=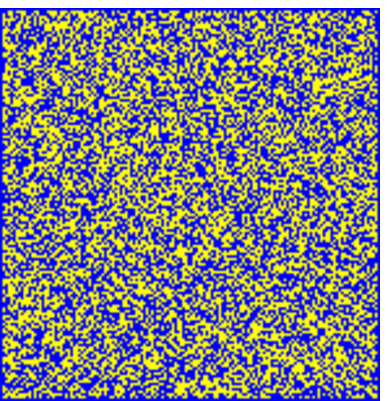,width=5cm} \vspace{1mm} \\
  (a) input image  &  (b) flux-decayed cell (yellow)\vspace{5mm} \\
  \psfig{file=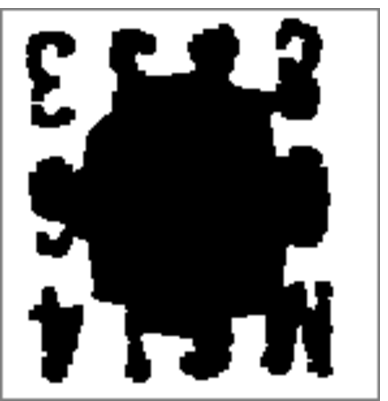,width=5cm} &
  \psfig{file=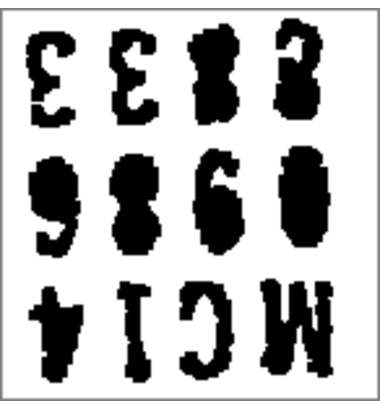,width=5cm} \vspace{1mm} \\
  (c) recovered image at $t=128$ & (d) output image at $t=340$ \\ 
  \end{tabular}
  \caption{Suspend and resume feature for the hole-filling template (\ref{eqn: template40}) 
   under the condition that the flux of the $50$ percent of memristors decays to small value \emph{without change of sign}.    
   \newline
   (a) Input binary image (image size is $146 \times 151$). 
   \newline 
   (b) After power off (during the period $t \in [70, \ 120)$), we assume that the flux of the $50$ percent of memristors decays to 
   small value \emph{without change of sign} by parasitic capacitances.  
   The flux-decayed cells are colored in yellow.  
   Other cells are colored in blue.     
   The cell size is $146 \times 151$, since the input image size is  $146 \times 151$. 
   \newline 
   (c) Recovered image at $t=128$. 
   \newline 
   (d) Output image at $t=340$.  
   Observe that the memristor CNN defined by Figures \ref{fig:store} and \ref{fig:recovery} works well 
   even if the flux of the $50$ percent of memristors decays to sufficiently small value.  
   }
 \label{fig:output-2000}
\end{figure}
%
%

\begin{figure}[ht]
 \centering
  \begin{tabular}{ccc}
  \psfig{file=./figure/WOMAN-2-input.eps,width=5cm} & 
  \psfig{file=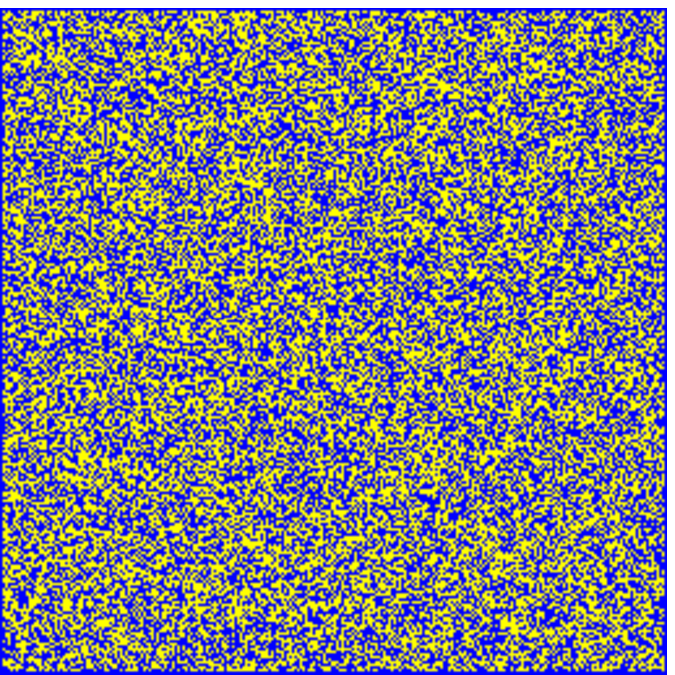,width=5cm} \vspace{1mm} \\
  (a) input image  & (b) flux-decayed cell (yellow) \vspace{5mm} \\
  \psfig{file=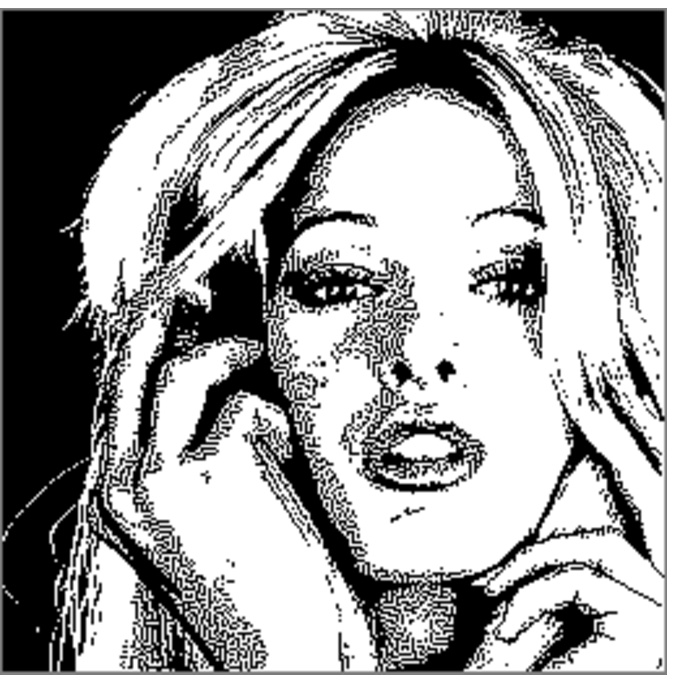,width=5cm} &
  \psfig{file=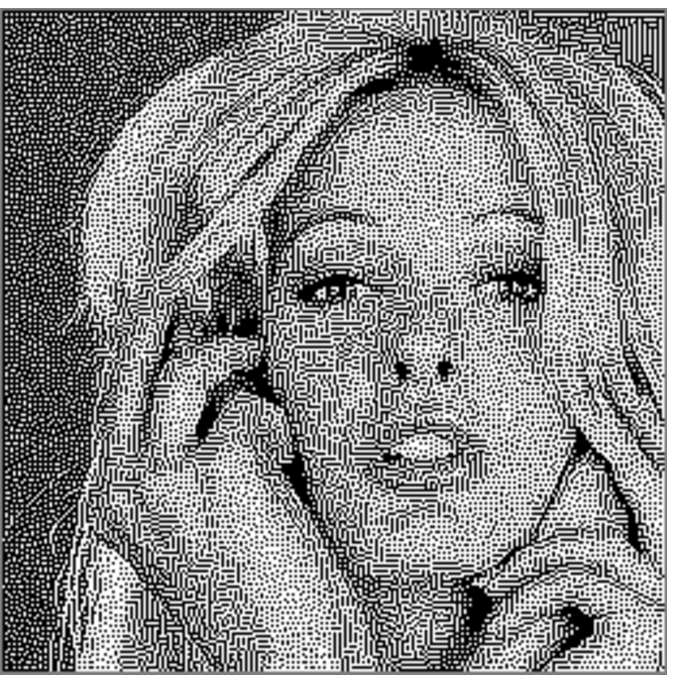,width=5cm} \vspace{1mm} \\
  (c) recovered image at $t=1.09$ & (d) output image at $t=5$ \\ 
  \end{tabular}
  \caption{Suspend and resume feature for for the half-toning template (\ref{eqn: template50})  
   under the condition that the flux of the $50$ percent of memristors decays to small value \emph{without change of sign}.   
   \newline 
   (a) Input gray-scale image (image size is $256 \times 256$). 
   \newline 
   (b) After power off at $t=0.35$, we assume that the flux of the $50$ percent of memristors decays to 
   small value \emph{without change of sign} by parasitic capacitances.  
   The flux-decayed cells are colored in yellow.  
   Other cells are colored in blue.   
   The cell size is $256 \times 256$, since the input image size is $256 \times 256$. 
   \newline 
   (c) Recovered image at $t=1.09$.  
   \newline 
   (d) Output image at $t=5$.  
   Observe that the memristor CNN defined by Figures \ref{fig:store} and \ref{fig:recovery} works well 
   even if the flux of the $50$ percent of memristors decays to sufficiently small value.    
   }
 \label{fig:output-3000}
\end{figure}
%
%

\begin{figure}[ht]
 \centering
  \begin{tabular}{ccc}
  \psfig{file=./figure/hole-filling-input.eps,width=5cm} & 
  \psfig{file=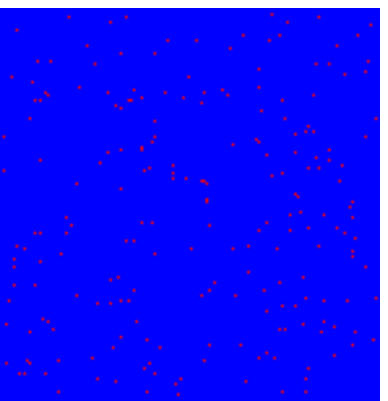,width=5cm} \vspace{1mm} \\
  (a) input image  &  (b) flux-decayed cell (red)\vspace{5mm} \\
  \psfig{file=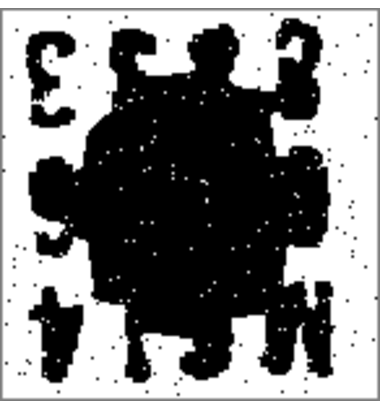,width=5cm} &
  \psfig{file=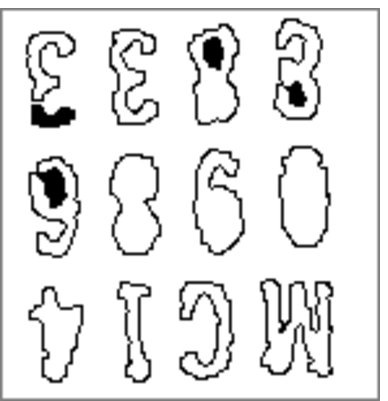,width=5cm} \vspace{1mm} \\
  (c) recovered image at $t=128$ & (d) output image at $t=340$ \\ 
  \end{tabular}
  \caption{Suspend and resume feature for the hole-filling template (\ref{eqn: template40}) 
   under the condition that the flux of the \emph{$1$ percent} of memristors decays to 
   small value \emph{with change of sign}.
   \newline
   (a) Input binary image (image size is $146 \times 151$). 
   \newline 
   (b) After power off at $t=70$, we assume that the flux of the \emph{$1$ percent} of memristors decays to 
   small value \emph{with change of sign}.  
   The flux-decayed cells are colored in red, and the other cells are colored in blue.   
   The cell size is $146 \times 151$, since the input image size is $146 \times 151$.   
   \newline 
   (c) The recovered image at $t=128$. 
   \newline 
   (d) Output image at $t=340$.  
    Observe that if the flux of the \emph{$1$ percent} of memristors decays to small value \emph{with change of sign}, 
    then the hole-filling template (\ref{eqn: template40}) does \emph{not} work well.    
   }
 \label{fig:output-4000}
\end{figure}

\clearpage
\begin{figure}[ht]
 \centering
  \begin{tabular}{ccc}
  \psfig{file=./figure/WOMAN-2-input.eps,width=5cm} & 
  \psfig{file=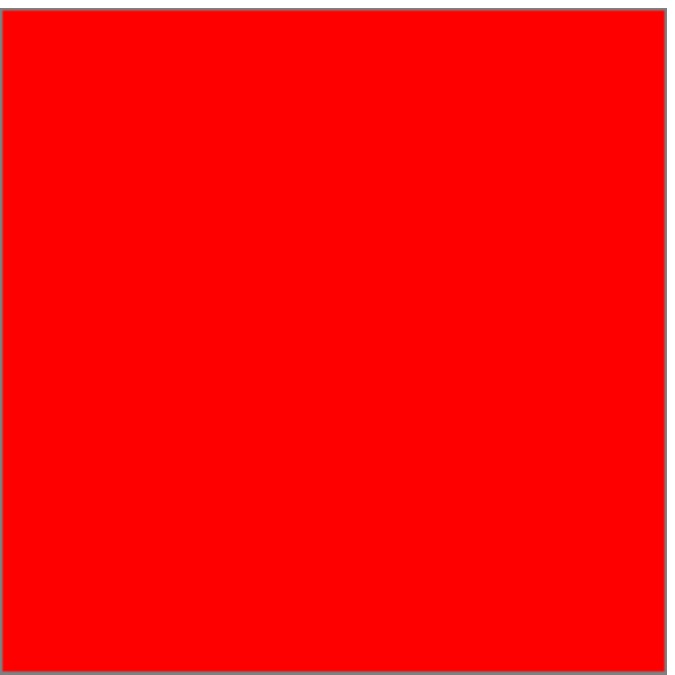,width=5cm} \vspace{1mm} \\
  (a) input image  & (b) flux-decayed cell (red) \vspace{5mm} \\
  \psfig{file=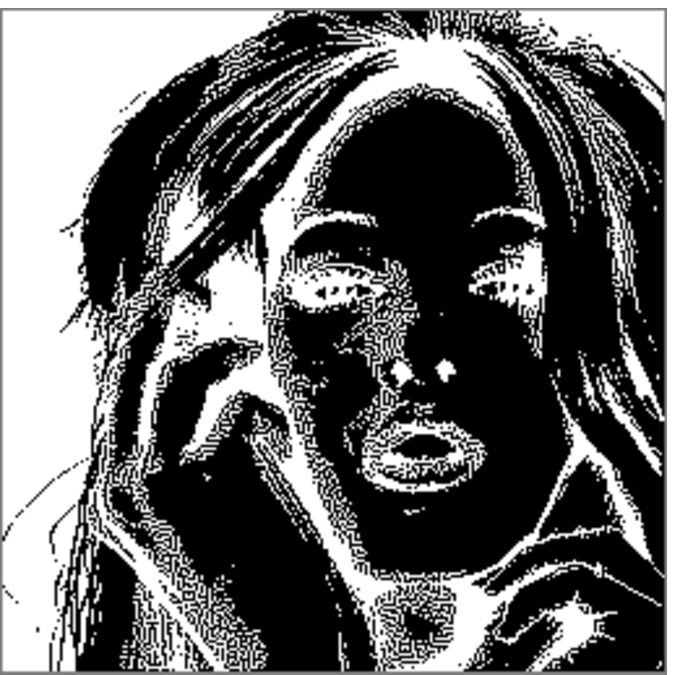,width=5cm} &
  \psfig{file=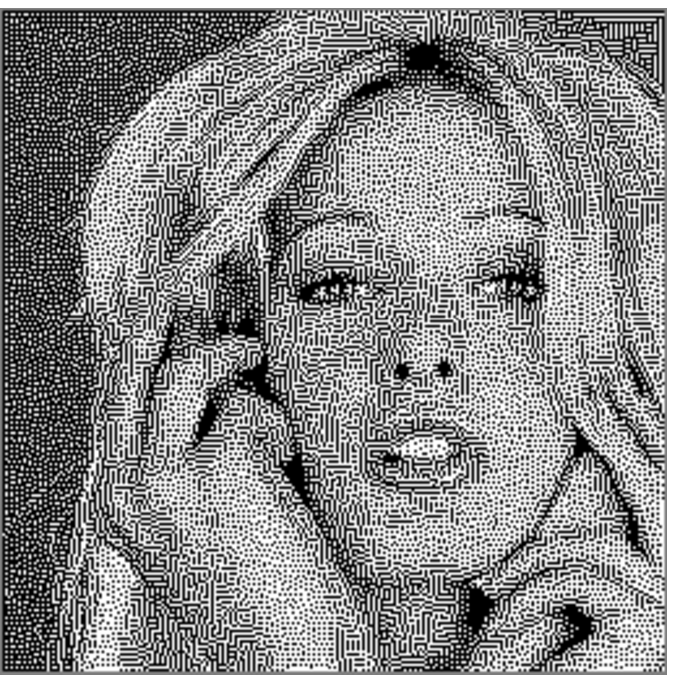,width=5cm} \vspace{1mm} \\
  (c) recovered image at $t=1.05$ & (d) output image at $t=5$ \\ 
  \end{tabular}
  \caption{Suspend and resume feature for for the half-toning template (\ref{eqn: template50}) 
   under the condition that the flux of \emph{all} memristors decays to small value \emph{with change of sign}.   
   \newline 
   (a) Input gray-scale image (image size is $256 \times 256$). 
   \newline 
   (b) After power off at $t=0.35$, we assume that the flux of \emph{``all''} memristors decays to 
   small value \emph{with change of sign}.  
   The flux-decayed cells are colored in red.   In this case, all cells are colored in red.   
   The cell size is  $256 \times 256$, since the input image size is $256 \times 256$.  
   \newline 
   (c) Recovered image at $t=1.09$.  
   \newline 
   (d) Output image at $t=5$.  
   Observe that  the half-toning template (\ref{eqn: template50}) works well,  
   even if the flux of \emph{all} memristors decays to sufficiently small value \emph{with change of sign}.    
   }
 \label{fig:output-5000}
\end{figure}

\newpage
%
\section{Reset of Memristors}
\label{sec: reset}
%

Before a new programming, we should reset the memristor to its zero state.  
A capacitor can simply be discharged by shorting its terminals (a wire has a negligibly small resistance).  
However, in order reset the flux of the memristor, we have to supply the \emph{reversed} input signal, which was used in the previous programming. 
If the flux-controlled memristor is \emph{ideal} and \emph{passive}, then we can reset the memristor to its zero state by connecting a small capacitor to the memristor in parallel \cite{Itoh2016}.  
 
%
%
%
%
\section{Two-dimensional Waves}
\label{sec: waves}
%
%
%
In this section, we show that the memristor CNN can exhibit interesting two-dimensional waves.  
Consider the memristor CNN circuit in Fig. \ref{fig:cnn-memistor-wave}.  
We connected the inductor $L$ in parallel with the capacitor $C$.   
In this case, the dynamics of the memristor CNN is given by  
\begin{center}
\begin{minipage}[]{12cm}
\begin{itembox}[l]{Dynamics of the memristor CNN}
\begin{equation}
 \begin{array}{lll}
  \displaystyle C \frac{dv_{ij}}{dt} 
   &=& \displaystyle i_{ij} - v_{ij} + 0.5 \, a \, \bigl ( \, |v_{ij}+1|-|v_{ij}-1| \, \bigr ) 
     \displaystyle + \ W (\varphi_{m})v_{m},  \vspace{2mm} \\
  \displaystyle L \frac{di_{ij}}{dt} 
   &=& \displaystyle - v_{ij} \vspace{2mm} \\
   && (i, \, j) \in \{ 1, \cdots , M \} \times \{ 1, \cdots , N \},
 \end{array}
\label{eqn: memristor-cnn-inductor} 
\end{equation}
\end{itembox}
\end{minipage}
\end{center}
where $L=C=1$.
The terminal currents and voltages of the flux-controlled memristor $D$ (yellow) satisfies  $i_{m} = W( \varphi_{m} ) \, v_{m}$.   
Here, the memductance $W( \varphi_{m} )$ is defined by 
\begin{equation} 
\left. 
 \begin{array}{cll}
   W( \varphi_{m} ) 
    &=&  \alpha \, \mathfrak{s}[\varphi_{m} - a] -  \beta \ \mathfrak{s}[\varphi_{m} - b] \vspace{2mm} \\
    &=& \left \{ 
      \begin{array}{clcc}
      \alpha & \ for \ & \ & -a \le \varphi_{m} < a,  \vspace{2mm} \\
      (\alpha - \beta)  & \ for \ & \ & \varphi_{m} < -b  \text{~and~}  b \le \varphi_{m},
     \end{array} \right.
  \end{array}
\right \}
\label{eqn: wave-W}
\end{equation} 
where  $\varphi_{m} (t) =  \int_{-\infty}^{t}v_{m}(\tau) d\tau$, $\alpha$, $\beta$, $a$, and $b$ are parameters, and the symbol $\mathfrak{s}  [\, z \,]$ denotes the \emph{unit step} function, equal to $0$ for $z < 0$ 
and 1 for $z \ge 0$, and $0 < a < b $.  

Let us consider the \emph{hole-filling template} \cite{{Chua1998},{Roska1997}} again 
\begin{equation}
 A =
 \begin{array}{|c|c|c|}
  \hline
   ~0~   &  ~1~   &  ~0~   \\
  \hline
   ~1~   &  ~3~   & ~1~   \\  
  \hline 
   ~0~   &  ~1~   &  ~0~  \\ 
  \hline
  \end{array} \ , \ \ \ 
 B = 
   \begin{array}{|c|c|c|}
  \hline
   ~0~ &  ~0~  &  ~0~ \\
  \hline
   ~0~ &  ~4~  &  ~0~   \\  
  \hline 
   ~0~ &  ~0~  &  ~0~    \\ 
  \hline
  \end{array} \ ,  \ \ \ 
 z =
  \begin{array}{|c|}
  \hline
    -1 \\
  \hline
  \end{array} \  . \vspace*{2mm}\\
\label{eqn: template-wave} 
\end{equation}
The initial condition for the state $v_{ij}$ and the input $u_{kl}$ are equal to a given binary image.  
The boundary condition is given by 
\begin{equation}
  v_{k^{*}l^{*}}  = 0,  \  u_{k^{*}l^{*}}  = 0,
\end{equation}
where $k^{*}l^{*}$ denotes boundary cells. 
If we adjust the parameters  $\alpha$, $\beta$, $a$, and $b$, then the memristor CNN (\ref{eqn: memristor-cnn-inductor})  can exhibit many interesting two-dimensional waves, as shown in Fig. \ref{fig:wave-1}.

\begin{figure}[htpb]
 \begin{center} 
  \psfig{file=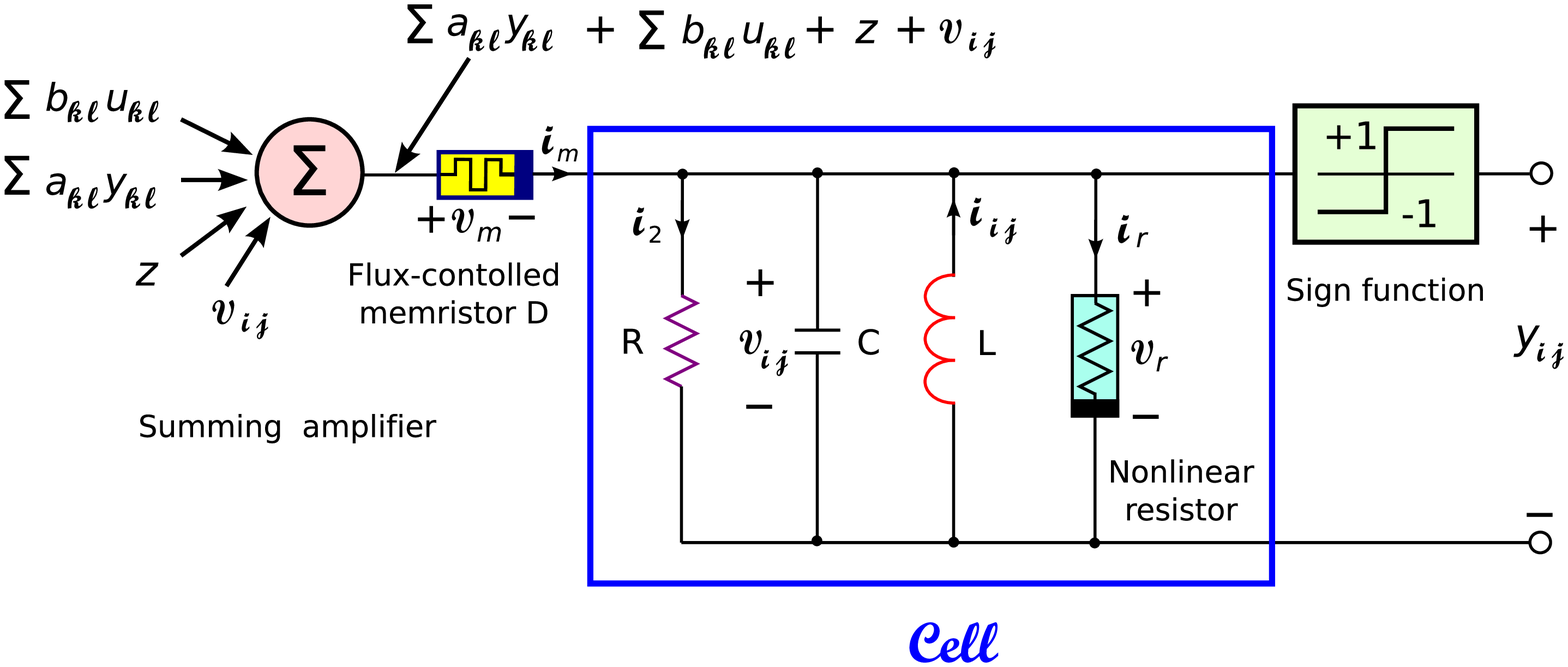,width=15cm}
  \caption{Memristor CNN circuit which can exhibit two-dimensional waves.  
  We connected the inductor $L$ (red) in parallel with the capacitor $C$.  
   \newline
   (1) Parameters: $R=1, \ C=1, \ L=1$.  
   \newline 
   (2) The $v-i$ characteristic of the nonlinear resistor (light blue) is given by 
   $i_{r} =  f_{r}(v_{r}) \stackrel{\triangle}{=} - 0.5 \, a \, ( |v_{r}+ 1 | -  |v_{r} - 1 | )$, 
   where $a$ is a constant.  
   \newline
   (3) The terminal currents and voltages of the flux-controlled memristor $D$ (yellow) satisfies 
   $i_{m} = W( \varphi_{m} ) \, v_{m}$, where $W( \varphi_{m} )$ is the memductance defined by Eq. (refl{eqn: wave-W}).  
   \newline
   (4) The output voltage of the summing amplifier (pink) is given by  
   $\displaystyle \sum_{k, \, l \in N_{ij}, \  k \ne i, \ l \ne j} a_{kl} \, \operatorname {sgn} (v_{kl}) 
   + \sum_{k, \, l \in N_{ij}}b_{k l} \ u_{kl} + z + v_{ij}$.  
   \newline
   (5) The above output voltage contains the voltage $v_{ij}$ of the capacitor $C$ (the last term).   
   \newline
   (6) The above sum 
   $\displaystyle \sum_{k, \, l \in N_{ij}, \  k \ne i, \ l \ne j} a_{kl} \, \operatorname {sgn} (v_{kl}) $ 
   does \emph{not} contain the term 
   $ a_{ij} \, \operatorname {sgn} (v_{ij})$ \ ($k=i, \ l=j$).   
   \newline  
   (7) The output $y_{ij}$ and the state $v_{ij}$ of each cell are related via the sign function (green): 
    $y_{ij} = \operatorname {sgn} (v_{ij})$.  }  
 \label{fig:cnn-memistor-wave}
 \end{center}
\end{figure}

\begin{figure}[p]
 \centering
  \begin{tabular}{ccc}
   \psfig{file=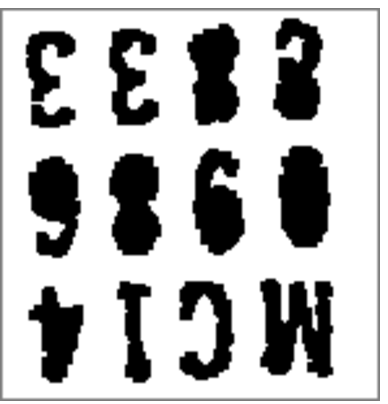,width=4cm} &
   \psfig{file=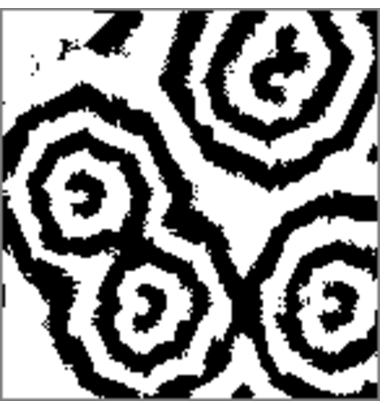,width=4cm} &
   \psfig{file=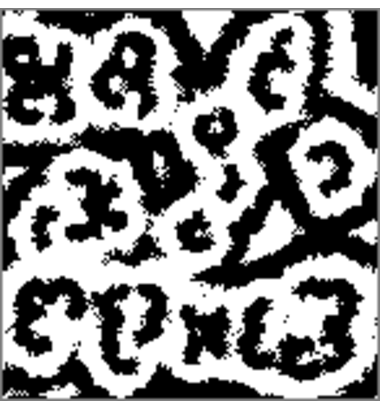,width=4cm} \vspace{2mm} \\
  (a) input image  & (b) $\alpha =1, \beta = 1, a=0.5, b=4000$  &  (c) $\alpha =1, \beta = 1, a=1, b=5000$ \vspace{2mm} \\
  \psfig{file=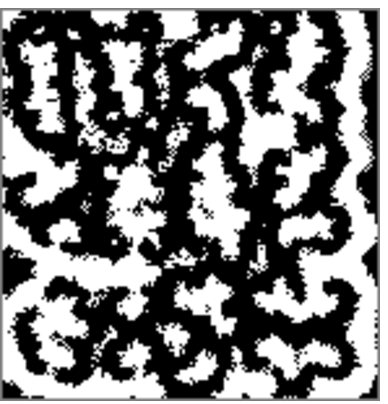,width=4cm} &  
  \psfig{file=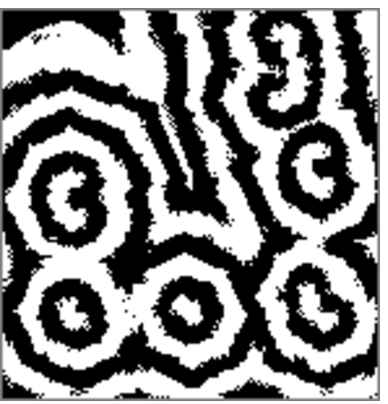,width=4cm} &
  \psfig{file=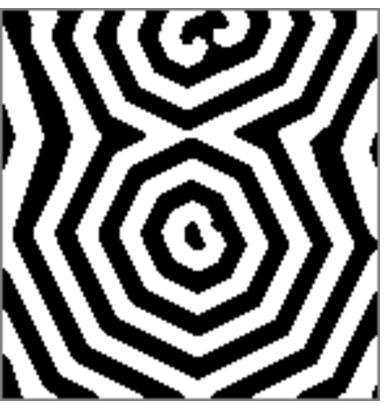,width=4cm} \vspace{2mm} \\  
  (d)  $\alpha =1, \beta = 1, a=2, b=5000$  & (e) $\alpha =1, \beta = 1, a=2.5, b=5000$ &  (f)  $\alpha =-1, \beta = 2, a=05, b=35$ \vspace{2mm} \\ 
  \psfig{file=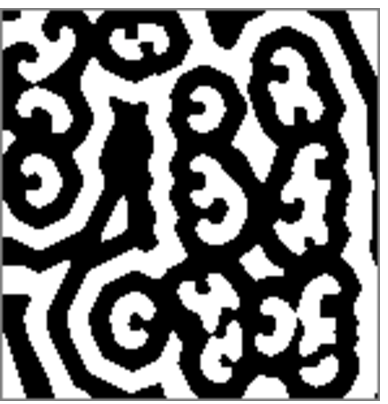,width=4cm}  &
  \psfig{file=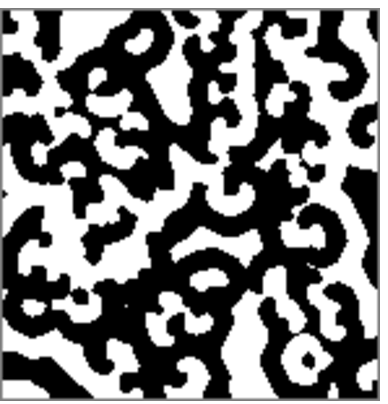,width=4cm} &  
  \psfig{file=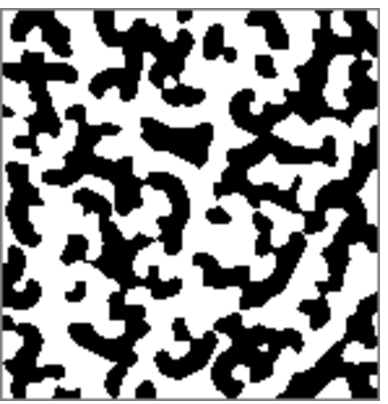,width=4cm} \vspace{2mm} \\ 
  (g) $\alpha =-1, \beta = 2, a=05, b=50$  & (h) $\alpha =-1, \beta = 2, a=05, b=2000$  & (i) $\alpha =-1, \beta = 2, a=05, b=3000$ \\
  \end{tabular}
  \caption{Two-dimensional waves, which is generated by the hole-filling template. (a) input image. (b)-(i) two-dimensional waves at $t=2000$.}
 \label{fig:wave-1}
\end{figure}
\clearpage
%
%
\section{Conclusion}
%
%
%
We have shown that the flux-controlled memristors cannot respond to the sinusoidal voltage source quickly.   
Furthermore, these memristors have the refractory period after switch ``on''.  
We have shown that the memristor-coupled two-cell CNN can exhibit chaotic behavior.  
We have also proposed the memristor CNN, which can hold the output image, even if even if all cells are disconnected and no signal is supplied to the cell after a certain point of time, by memristor's switching behavior.
We have next shown that even if we turn off the power of the memristor CNN during the computation, it can resume from the previous average output state.  That is, the memristor CNN has the suspend and resume feature.  
Furthermore, the memristor CNN has functions of the short-term and long-term memories.  
Finally, we have shown that the memristor CNN can exhibit the interesting two-dimensional waves, if an inductor is connected to each memristor CNN cell. 
In this paper, we used the Euler method for solving the differential equations.   
In order to get more accurate results, we may need high accuracy numerical methods, for example, the Runge-Kutta method.  
%
%

%
%

\end{document}